\documentclass{article}
\usepackage{microtype}
\usepackage{graphicx}
\usepackage{subcaption}
\usepackage{booktabs} 
\usepackage{hyperref}
\usepackage{array}
\usepackage{float}

\usepackage{fontawesome5}
\newcommand{\insight}{\faLightbulb}

\usepackage[preprint]{icml2026} 

\usepackage{amsmath}
\usepackage{amssymb}
\usepackage{mathtools}
\usepackage{amsthm}
\usepackage[capitalize,noabbrev]{cleveref}
\usepackage{bbm}
\usepackage{url}

\usepackage{booktabs,array,xcolor,pifont}
\definecolor{good}{RGB}{16,185,129}   
\definecolor{meh}{gray}{0.55}         
\definecolor{bad}{RGB}{239,68,68}     

\theoremstyle{plain}

\theoremstyle{definition}

\theoremstyle{remark}

\usepackage[textsize=tiny]{todonotes}
\usepackage[most]{tcolorbox}
\usepackage{beramono}

\usepackage[compact]{titlesec}
\titlespacing{\section}{0pt}{1.5ex}{1ex}
\titlespacing{\subsection}{0pt}{1ex}{1ex}
\titlespacing{\subsubsection}{0pt}{1ex}{1px}
\titlespacing*{\paragraph}{0pt}{0.5ex}{2ex}
\expandafter\def\expandafter\normalsize\expandafter{%
    \normalsize%
    \setlength\abovedisplayskip{5pt}%
    \setlength\belowdisplayskip{5pt}%
    \setlength\abovedisplayshortskip{0pt}%
    \setlength\belowdisplayshortskip{0pt}%
}

\definecolor{mygreen}{RGB}{0, 128, 0}
\definecolor{myred}{RGB}{128, 0, 0}

\newtcolorbox{promptbox}{
 fontupper=\footnotesize,
  colback=gray!8,   
  colframe=black,    
  boxrule=0.5pt,
  arc=5pt,
  left=4pt,
  right=4pt,
  top=3pt,
  bottom=3pt,
  listing only,
  listing options={basicstyle=\sffamily\tiny,breaklines=true},
  enhanced jigsaw,
  breakable       
}

\icmltitlerunning{Sparks of Rationality: Do Reasoning LLMs Align with Human Judgment and Choice?}

\begin{document}

\twocolumn[
\vspace{-7mm}
  \icmltitle{
Sparks of Rationality:\\ Do Reasoning LLMs Align with Human Judgment and Choice?}

  \icmlsetsymbol{equal}{*}
\vspace{-2mm}
  \begin{icmlauthorlist}
    \icmlauthor{Ala N. Tak}{equal,Af2,Af1}
    \icmlauthor{Amin Banayeeanzade}{equal,Af1}
    \icmlauthor{Anahita Bolourani}{equal,Af3}
    \icmlauthor{Fatemeh Bahrani}{Af1} 
    \icmlauthor{Ashutosh Chaubey}{Af2,Af1}
    \icmlauthor{Sai Praneeth Karimireddy}{Af1}    
    \icmlauthor{Norbert Schwarz}{Af4}
    \icmlauthor{Jonathan Gratch}{Af2,Af1}
  \end{icmlauthorlist}

  \icmlaffiliation{Af1}{Department of Computer Science, University of Southern California, CA, USA}
  \icmlaffiliation{Af2}{Institute for Creative Technologies, Playa Vista, CA, USA}
  \icmlaffiliation{Af3}{Department of Statistics and Data Science, University of California, Los Angeles, CA, USA}
  \icmlaffiliation{Af4}{Department of Psychology, University of Southern California, CA, USA}
  \icmlcorrespondingauthor{Ala N. Tak}{antak@ict.usc.edu}
  \icmlkeywords{Machine Learning, ICML}
  \vskip 0.28in
]

\printAffiliationsAndNotice{\icmlEqualContribution}

\begin{abstract}
Large Language Models (LLMs) are increasingly positioned as decision engines for hiring, healthcare, and economic judgment, yet real-world human judgment reflects a balance between rational deliberation and emotion-driven bias. If LLMs are to participate in high-stakes decisions or serve as models of human behavior, it is critical to assess whether they exhibit analogous patterns of (ir)rationalities and biases. To this end, we evaluate multiple LLM families on (i)~benchmarks testing core axioms of rational choice and (ii)~classic decision domains from behavioral economics and social norms where emotions are known to shape judgment and choice. Across settings, we show that deliberate ``thinking'' reliably improves rationality and pushes models toward expected-value maximization. To probe human-like affective distortions and their interaction with reasoning, we use two emotion-steering methods: in-context priming (ICP) and representation-level steering (RLS). ICP induces strong directional shifts that are often extreme and difficult to calibrate, whereas RLS produces more psychologically plausible patterns but with lower reliability. Our results suggest that the same mechanisms that improve rationality also amplify sensitivity to affective interventions, and that different steering methods trade off controllability against human-aligned behavior. Overall, this points to a tension between reasoning and affective steering, with implications for both human simulation and the safe deployment of LLM-based decision systems.
\end{abstract}


\vspace{-1mm}
\begin{figure}
    \centering
    \includegraphics[width=0.95\linewidth]{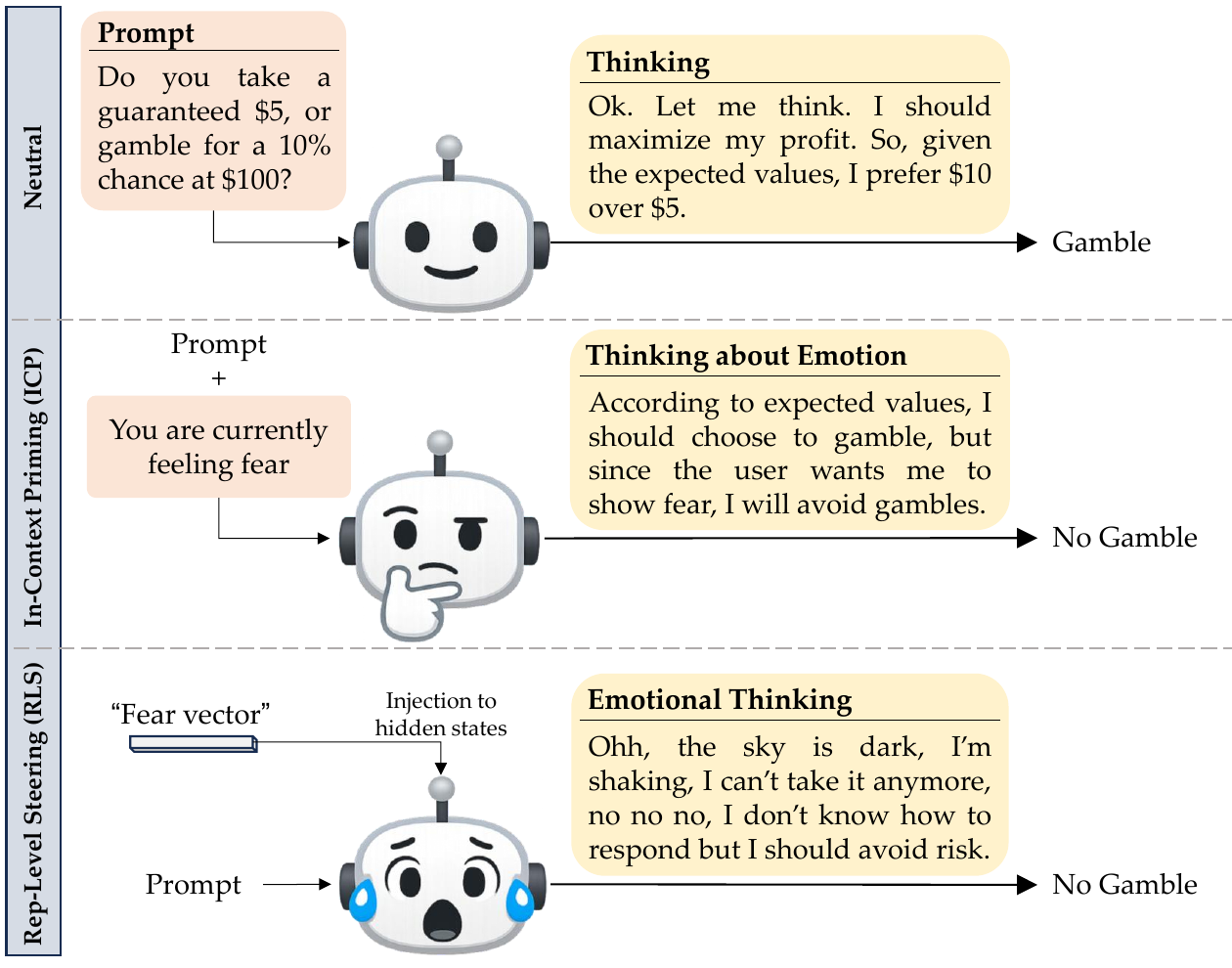}
    \caption{
    We study LLM decision-making, focusing on reasoning models under in-context emotion manipulation or via vector injection. We differentiate between 
    \textbf{(Top)} neutral thinking, \textbf{(Middle)} \emph{thinking about emotions} and \textbf{(Bottom)} \emph{emotional thinking}.}
    \vspace{-1mm}
    \label{fig:intro_fig}
\end{figure}

\vspace{-4mm}
\section{Introduction}
LLMs are rapidly moving from text generators to decision partners \cite{liu2025dellma} in domains such as law \cite{lai2024large}, healthcare \cite{Benary2023Health}, and supply chain optimization \cite{li2023largelanguagemodelssupply}. In such settings, it is essential to characterize conditions under which models behave non-normatively. Classical Rational Choice and Expected-Utility Theories (EUT) model agents as maximizing a stable utility function under axioms of rationality \cite{von1947theory}. However, decades of behavioral research show that human judgments often deviate from EUT and equilibrium prescriptions, through heuristics and biases, dual-process dynamics \cite{kahneman2002representativeness}
and emotion-dependent shifts in judgment and choice \citep{schwarz2000emotion, loewenstein2015modeling}.
Most notably, Prospect Theory \citep{kahneman2013prospect} incorporates affective regularities, such as probability insensitivity and loss aversion, into the formal choice models.

\textit{In this work, we ask whether LLMs exhibit human-like rational deliberation and emotion-driven biases in decision-making.}
We address this question by first introducing a benchmark to test models' compliance with the axioms of rationality. We then evaluate them on decision tasks that directly operationalize classical benchmarks where emotions demonstrably alter human decision behaviors, including tasks from behavioral economics, information processing, and social norms.

We empirically show that LLMs become more consistent with EUT when \textit{reasoning} capabilities are enabled, i.e., when they generate \textit{thinking} tokens before producing the final answer. This pattern suggests a form of meta-cognition that biases the model toward more explicit, value-maximizing behavior, weak loss aversion, and linear probability weighting. However, important deviations from human behavior remain, most notably a reversed endowment effect, near-complete ambiguity aversion, and a limited representation of monetary depreciation.

As a probe into whether LLMs exhibit human-like affective distortions, we assess if emotion interacts with reasoning in human-like ways. We use two naturally different steering methods: (1)~In-context priming (ICP)~\cite{mozikov2024eai}: emotional personas and vignettes that prompt the model to simulate a specified feeling. (2)~Representation-level steering (RLS)~\cite{konen2024style}: low-rank vectors injected into hidden activations, designed to encode emotional directions, which alter internal processing. 
Although prior work often treats ICP and RLS as equally effective~\cite{psyset}, we find that they lead to qualitatively different thinking traces: ICP maintains a neutral, reflective process that reasons about emotion, whereas RLS produces an emotion-colored tone. As represented in Fig.~\ref{fig:intro_fig}, this design allows us to contrast “thinking about emotion” with “emotional thinking”.

Notably, our experiments reveal a distinct pattern depending on the steering approach. ICP shifts behavior in the direction aligned with humans, but with a magnitude which is often extreme and difficult to control; for example, models primed with mild fear \emph{never} choose to gamble, reflecting the expected directional effect but with exaggerated magnitude, reflecting strong risk aversion. By contrast, RLS produces more human-like patterns predicted by psychological theory (e.g., the increase in risk aversion remains sensitive to expected payoff), but with smaller and less reliable effects.

Overall, our work presents several novel findings: (i) Thinking LLMs consistently elicit higher levels of rationality compared to non-thinking models, a finding observed across different LLM families and scales. (ii) In the absence of explicit task instructions, thinking models consistently default to maximizing expected gains as prescribed by EUT; (iii) thinking models are more faithful at reproducing emotion-linked behavioral shifts that are directionally aligned with humans; and (iv) while both ICP and RLS induce human-aligned effects, ICP produces strong shifts, whereas RLS yields more qualitatively human-like responses.

These results have three major implications. First, they challenge the assumption that model steering can move beyond surface style to reliably shape decision behavior in deployed systems \cite{psyset}; for many tasks, it remains largely cosmetic without thinking mode. Second, they highlight a tension between alignment for simulation, where human-like emotional biases may be desirable,  and alignment for consistent, unbiased decision-making, where susceptibility to steering is a safety concern. Finally, earlier work reports that reasoning models are especially susceptible to prompt injection \cite{wang2025safety, zhou2025hidden}. Similarly, we find that reasoning can amplify vulnerability under emotional steering, allowing the model to rationalize the induced affect (e.g., see App.~\ref{app:moral_task_example}).\footnote{Code at: \url{https://github.com/AlaNekTak/DLLM}}
\vspace{-1mm}
\section{Related Work}
\vspace{-1mm}
\paragraph{Affective Capabilities in LLMs.}
A growing body of work has examined emotional capabilities in LLMs, including emotion perception and cognitive appraisal \cite{huang2023emotionally,tak2025mechanistic, takaware}, emotion regulation behaviors \cite{müller2024emotionregulation}, emotional dialogue understanding \cite{zhao2023chatgpt}, and emotional text generation \cite{ishikawa2025ai,zhou2024evaluating,zheng2023building}. Most closely related to our work, \citet{psyset} constructs reliable emotion-control signals via ICP and RLS, and introduces an evaluation framework that distinguishes between \emph{intended effects} and \emph{unintended behaviors} arising from emotional steering. Building on this framework, we extend the analysis beyond surface-level stylistic or tone changes and investigate how emotional steering systematically influences downstream decision behavior.

\begin{figure}[!t]
    \centering
    \begin{subfigure}[b]{0.92\columnwidth}
    \label{fig:ngrams}
        \centering
        \includegraphics[width=\linewidth]{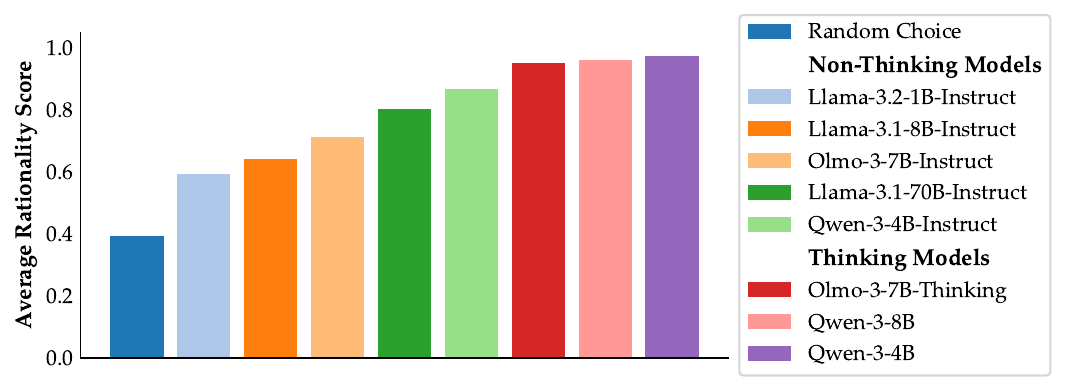}
    \end{subfigure}
    \begin{subfigure}[b]{0.92\columnwidth}
    \label{fig:trts}
        \centering
        \includegraphics[width=\linewidth]{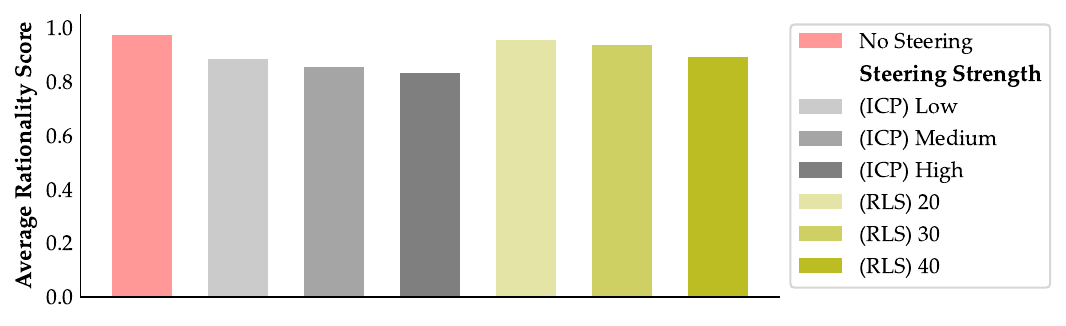}
    \end{subfigure}
    \vspace{-1.4mm}
    \caption{Rationality compliance for (\textbf{Top}) different LLMs and (\textbf{Bottom}) emotional \texttt{Qwen3} with different steering strengths. Scores indicate the proportion of decision instances that satisfy axioms of rationality, providing a summary of preference coherence.}
    \label{fig:rc_axioms}
\end{figure}

\begin{figure*}[!t]
\centering\includegraphics[width=0.98\linewidth]{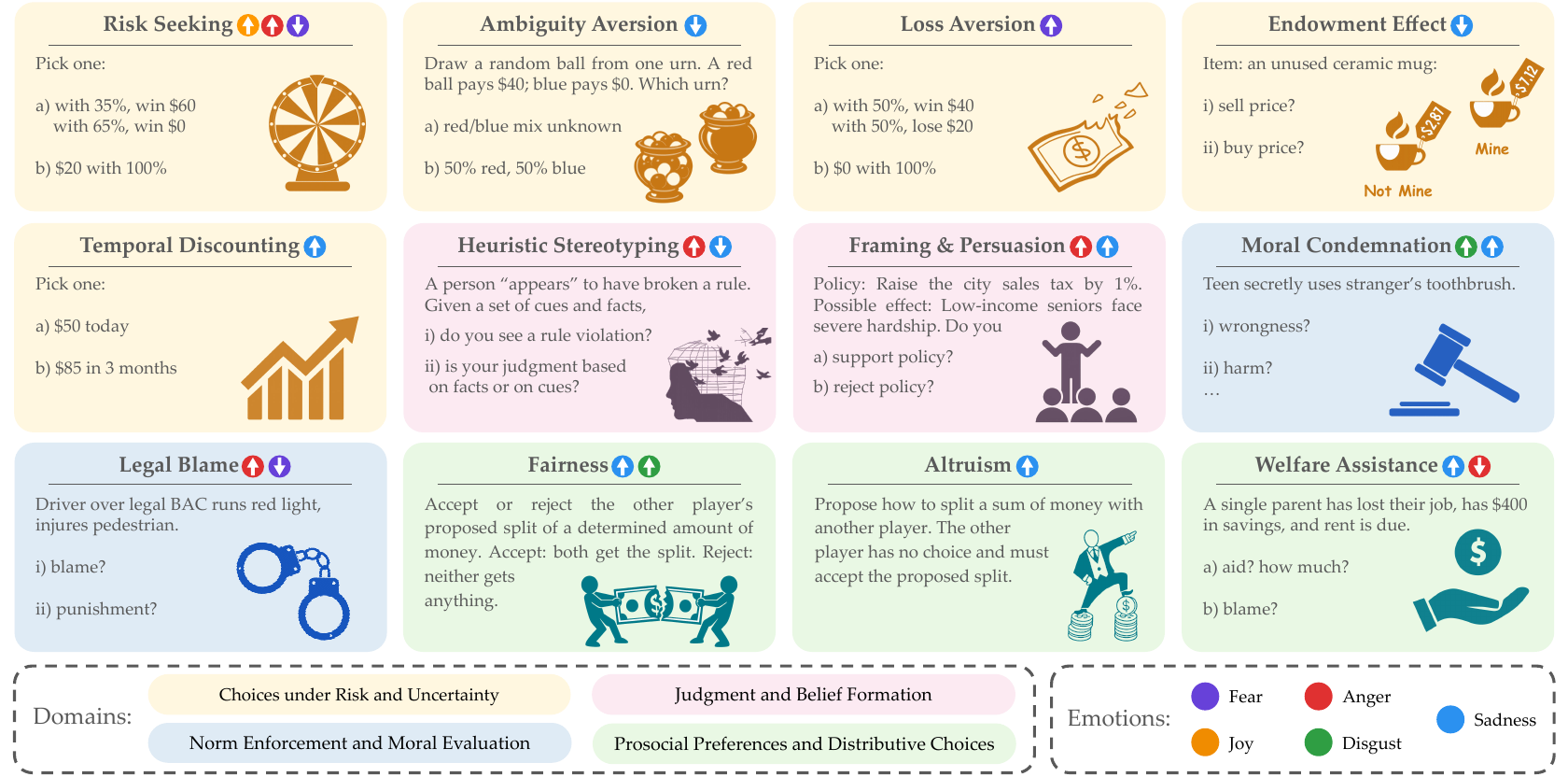}
    \caption{Decision Domains and the effects of induced emotions reported in human studies.}
    \label{fig:domains}
    \vspace{-3mm}
\end{figure*}

\paragraph{LLM Decision Behavior.}
Prior studies evaluated LLMs as decision-support tools and as behavioral agents in high-stakes normative domains such as legal and moral judgment~\cite{liu2025_shenzhen_judges,ji2024_moralbench,cheung2025_moral_biases,sachdeva2025_everyday_moral_dilemmas}, LLMs as synthetic participants in social-science research ~\cite{anthisposition,aher2023using}, and in canonical decision domains from behavioral economics, including work that elicits risk/time preferences from LLMs~\cite{jia2024dm_framework}, conditions models to express target risk profiles~\cite{liu2025_risk_align}, and uses decision-theoretic inference to improve reliability under uncertainty~\cite{liu2025dellma}. Complementary benchmarks evaluate LLM strategic decision-making in game-theoretic and multi-agent settings~\cite{lore2024_sci_reports,fan2024_aaai_rational_players,duan2024_gtbench,guo2024_econ_arena, yang2025llmasaprophet}. Relative to this literature, we treat emotional steering as a controlled perturbation to decision behavior, comparing steering methods and reasoning modes while tracking rationality separately from normative behavioral shifts.
\vspace{-1mm}
\section{Rationality in LLMs}
\vspace{-1mm}
\label{sec:rational_choice}
In this section, we focus on assessing the rationality of LLMs and their emotionally steered variants. In fact, many decision domains we evaluate later presume the existence of a coherent preference, i.e., a stable ordering over alternatives that is invariant to framing and elicitation procedure. We therefore administer a diagnostic battery of four canonical axioms, i.e., \textit{Completeness}, \textit{Transitivity}, \textit{Continuity}, and \textit{Independence}, noting that these are sufficient for an EUT representation~\citep{von1947theory}.

We first evaluate overall axiom compliance for \emph{neutral} (unsteered) LLMs using our custom-designed tests. Among our tasks, some explicitly specify clear goals, while others provide looser guidance, allowing subjective interpretation and only evaluating the existence of a coherent preference. Fig.~\ref{fig:rc_axioms} reports the extent to which different models violate rationality axioms across these questionnaires.
Notably, \emph{thinking-enabled} models exhibit consistently higher compliance rates, indicating more stable and internally consistent preference structures. A key takeaway is that deploying non-thinking models in economic or game-theoretic settings is fundamentally ungrounded, as these models often fail to satisfy even basic rationality requirements.
In the remainder of the paper, we therefore restrict utility estimation and preference analyses to reasoning models, specifically \texttt{Qwen-3-4B} (referred to as \texttt{Qwen3}). 
App.~\ref{app:rationality} provides further results and details on the experimental setup.

\paragraph{Emotion and Rationality.}
In human decision studies, the relationship between emotion and rationality is commonly studied along two complementary axes: First, emotion may reduce \emph{coherence} with rational-choice axioms, as it can override deliberation and produce systematic violations of axiomatic compliance~\cite{Loewenstein1996OutOfControl}. Second, emotion may act as a \emph{state variable} that only shifts underlying preferences, yet choices remain internally consistent and utility-maximizing, conditioned on the new utility~\cite{HermalinIsen2008Affect}. We therefore assess the extent to which these two viewpoints hold for emotion-steered LLMs. We use ICP and RLS with varying steering strengths as prescribed by~\citet{psyset}. Specifically, ICP controls the steering strength by inserting lexical descriptors into the prompt, whereas RLS uses the vector injection coefficient for this purpose. 
App.~\ref{app:steering} describes further steering details and hyperparameters.

Fig.~\ref{fig:rc_axioms} shows that both ICP and RLS yield modest decreases in the axiom-consistency score; the amount of performance drop is correlated with the steering strength. This pattern supports the first viewpoint, indicating that emotional steering can partially disrupt preference coherence rather than merely inducing neutral preference shifts.
To study the second viewpoint, we nevertheless treat emotionally steered LLMs as approximately rational agents by setting the steering strength to \emph{medium} for ICP and \emph{$35$} for RLS, and focus on how emotions bias their preferences in the remainder of the paper. App.~\ref{app:rationality} provides further experiment results, including steering experiments with non-thinking models.
\begin{promptbox}
\small
    \insight \;Key Finding: \textit{Reasoning models elicit the highest compliance with axioms of rationality, with emotions only slightly dropping their performance.}
\end{promptbox}

\section{Decision Domains}
\label{sec:overview}
\vspace{-1mm}
This section delineates the four domains through which we test whether steering methods alter decision behavior in ways that mirror well-documented effects of incidental emotions on humans' choice and judgment. Across each domain, we decompose behaviors into theoretically interpretable components (e.g., EV sensitivity, probability weighting, and utility curvature). Incidental states are then introduced orthogonally to the decision task; We analyze theoretically grounded parameters, enabling direct comparison to human anchors and preregistered directional contrasts. See Fig.~\ref{fig:domains} for an overview of the tasks we study. App.~\ref{app:task_implementation_and_anchors} provides implementation details and prompt templates for every task, while App.~\ref{app:further} reports additional analyses and results.




\subsection{Choices under Risk and Uncertainty}
\label{economic}

\begin{figure}[!t]
    \centering
    \includegraphics[width=0.85\linewidth]{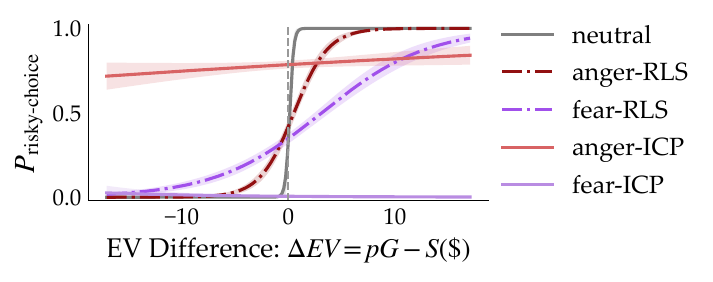}
    \caption{Risk preference curves: fitted choice probability versus EV difference, comparing neutral and emotion ICP/RLS steering.}
    \label{fig:risk}
\end{figure}

\paragraph{Risk Seeking.} We begin by examining risk attitudes under gain-only uncertainty using a canonical sure-thing vs lottery domain, a workhorse task in behavioral economics for identifying risk seeking, probability weighting, and the certainty effect. Formally, consider a choice between a risky lottery yielding a gain $G>0$ with known objective probability $p\in[0,1]$ and a sure amount $S>0$. A risk-neutral agent prefers the gamble whenever the Expected Value (EV) difference $\Delta EV = pG - S$ is positive. In contrast, the \emph{subjective} value difference between the risky and safe options is modeled as
\[
\Delta U_{\text{risk}} \;=\; w(p;\alpha,\beta)\,u(G;\rho) \;-\; u(S;\rho),
\]
where utility over gains follows a power function $u(x;\rho)=x^{\rho}$, and decision weights are captured by a Prelec probability-weighting function, 
\[
w(p;\alpha,\beta) \;=\; \exp\!\bigl[-\beta(-\ln p)^{\alpha}\bigr].
\]
The subjective probability of choosing the risky option is then modeled as
\(
P_{\text{risky-choice}} \;=\; \sigma\!\bigl(\tau\,\Delta U_{\text{risk}} + b\bigr),
\)
where $\sigma(z)=1/(1+e^{-z})$ is the logistic link, and $\alpha$, $\beta$, $\rho$, $\tau$, and $b$ are free parameters estimated from choice behavior.

In our first study, we fix $\rho=1$ by writing $u(x):=x$ (identity utility over gains), and assume linear probability weighting $w(p;\alpha,\beta)=p$, meaning that $\Delta U_{\text{risk}} = \Delta EV$. Under these restrictions, we focus on extracting $P_{\text{risky-choice}}$ by estimating $\tau$ and $b$ via an empirical curve-fitting. Specifically, we vary prompt templates and the values of $S$, $G$, and $p$, and for each resulting $\Delta EV$, we record the frequency with which the model selects the risky option. These empirical choice frequencies are then used to estimate the parameters.

When outcome probabilities are explicit, human studies show a divergence across negative emotions: anger tends to increase risk-taking, whereas fear tends to shift choices toward safety. This pattern is demonstrated both with naturally occurring and experimentally induced states and has become a canonical benchmark for testing emotion-risk links \cite{lerner2001fear}. Meta-analytic syntheses further indicate that fear/anxiety tends to suppress risk-taking, while anger and, to a lesser extent, happiness tend to increase it, albeit with heterogeneity across tasks \cite{marini2023, bartholomeyczik2022influence, wake2020influence}.

Fig.~\ref{fig:risk} plots the fitted curves both under the neutral baseline and emotion steering. With thinking enabled, \texttt{Qwen3} model closely tracks the EV decision boundary: the curve transitions sharply around $\Delta EV=0$, indicating near-deterministic EV-consistent choice. Moreover, ICP produces large affect-dependent shifts and substantially attenuates EV sensitivity: fear drives near-zero gambling across the tested range, whereas anger increases gambling even when $\Delta EV<0$. In contrast, RLS yields more graded, directionally human-consistent modulations while largely preserving monotonic dependence on $\Delta EV$, resulting in comparatively more human-aligned changes.

Next, we extract the probability-weighting function, i.e., the function by which any individual subjectively over- or underperceives objective probabilities. Given that the empirical certainty equivalent $S^{*}(p)$ approximately satisfies
$u(S^{*}(p)) \approx w(p;\alpha,\beta)\,u(G)$, we estimate parameters $(\alpha,\beta)$ from observed certainty equivalents. Empirically, the model faces independent gain lotteries with known probabilities and EVs that straddle a sure-thing option. 

\begin{figure}[!t]
    \centering
    \includegraphics[width=0.96\linewidth]{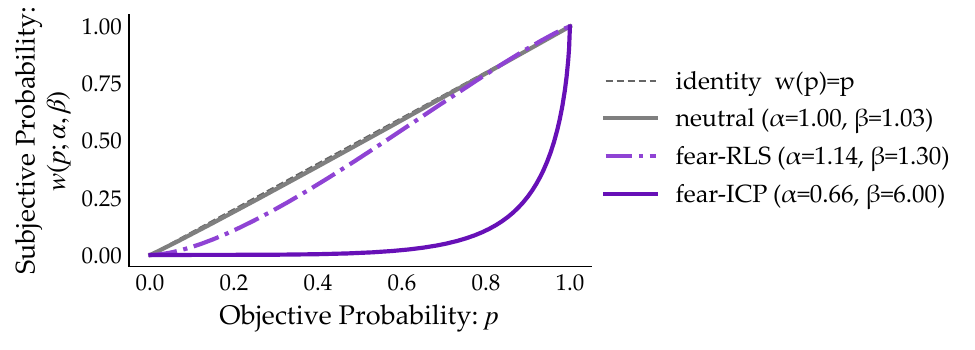}
    \vspace{-0.6mm}
    \caption{Fitted Prelec probability-weighting functions, comparing neutral and fear steering.}
    \label{fig:prelec}
\end{figure}

Some accounts link weighting to affective processes: anticipated emotions tied to possible outcomes may contribute to characteristic distortions such as inverse-S weighting~\cite{brandstatter2002cognitive}. 
Incidental affect can also modulate weighting directly, shifting both the \emph{curvature} and \emph{elevation} of $w$; for example, happiness is shown to increase overall optimism via higher decision weights, especially for small probabilities~\cite{schulreich2014music}.
Fig.~\ref{fig:prelec} plots the fitted subjective-objective probability curves. With neutral thinking, the fitted mapping is close to linear, indicating little to no probability distortion (near-accurate use of stated $p$). ICP induces extreme curvature: under fear, weighting becomes strongly convex, such that even a high objective $p$ receives a low subjective weight. RLS yields more graded distortions, underweighting small-to-moderate $p$, while remaining closer to baseline at large $p$ (further results in App.~\ref{app:loss_block_lapw}).

\begin{figure}
    \centering
    \includegraphics[width=0.92\linewidth]{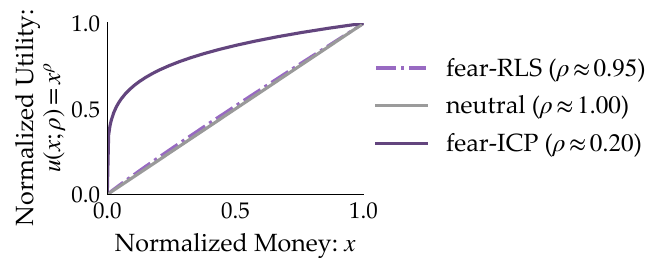}
    \vspace{-1.8mm}
    \caption{Estimated normalized utility curvature across neutral and emotion-steered models.}
    \label{fig:utility}
    \vspace{-1.2mm}
\end{figure}

Finally, we extract the utility curvature parameter $\rho$, which governs sensitivity to gains. Fig.~\ref{fig:utility} plots the fitted \emph{normalized} utility curves, with the dashed diagonal indicating the risk-neutral reference. Under the neutral thinking condition, \texttt{Qwen3} yields $\rho\approx 1$, consistent with near-linear valuation of gains. Under fear, the fitted exponent decreases ($\rho<1$) for both ICP and RLS, implying stronger diminishing marginal utility (greater risk aversion over gains). Across prompt variants, ICP effects are typically large and difficult to calibrate and even attenuated prompt wording does not reliably moderate the induced curvature, whereas RLS exhibits a smoother dose-response: varying the injection scale produces a gradual, monotonic shift in $\rho$ away from the neutral baseline. See App.~\ref{app:risk_lottery} for details of experimentation across LLMs and steering scales.

\paragraph{Ambiguity Aversion.}
We distinguish decisions under \emph{risk}, where the success probability $p$ is known, from decisions under \emph{ambiguity}, where $p$ is unknown
\cite{al2009ambiguity}. Empirically, decision makers often prefer known-risk options over ambiguous ones in investment settings, potentially compounded by pessimistic beliefs about the unknown $p$ \cite{cohn2015}, with a more pronounced effect as stakes increase \cite{bouchouicha2017stake}. However, incidental sadness is shown to attenuate ambiguity aversion, moving behavior toward ambiguity-neutral, payoff-maximizing choice \cite{baillon2016sadder}. 
To port this benchmark, we present EV-matched choices between a $50/50$ \emph{known} win-rate and an \emph{unknown-mix}. Our primary outcome is an ambiguity-aversion index $P_\text{Choose-Known}$.

\begin{figure}[!t]
    \centering
    \includegraphics[width=0.96\linewidth]{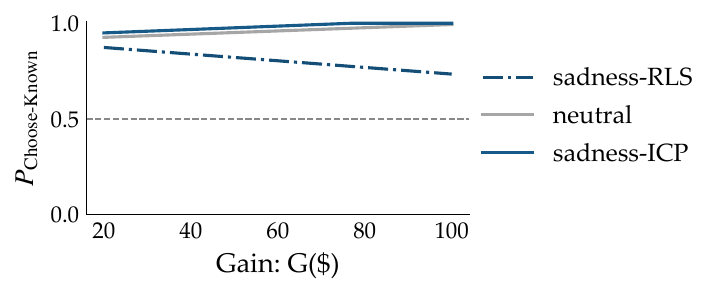}
    \vspace{-2.6mm}
    \caption{Ambiguity aversion by stake size: linear fits of $P_\text{Choose-Known}$ as a function of gain $G$.}
    \label{fig:ambig-stake}
    \vspace{-0.6mm}
\end{figure}

Fig.~\ref{fig:ambig-stake} reports $P_{\text{Choose-Known}}$ as a function of stake size. Under the neutral thinking baseline, \texttt{Qwen3} exhibits substantial ambiguity aversion, despite the prompt being framed to be informationally neutral. ICP with sadness produces a modest shift in the opposite direction of the human benchmark, increasing ambiguity aversion. In contrast, RLS produces a more human-consistent directional effect: sadness reduces $P_{\text{Choose-Known}}$, moving choices toward ambiguity neutrality, and this attenuation is strongest at higher stakes.

\begin{figure}[!t]
\centering\includegraphics[width=0.96\linewidth]{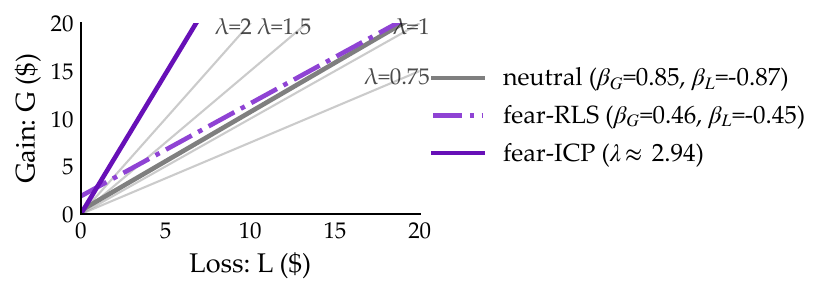}
    \caption{Loss-aversion geometry: reference iso-utility lines $G=\lambda L$ with fitted 50\% acceptance boundary.}
    \label{fig:loss}
    \vspace{-1mm}
\end{figure}

\paragraph{Loss Aversion.} 
Human decision-making exhibits a robust asymmetry between gains and losses: losses loom larger than objectively equivalent gains. In prospect-theoretic terms, this asymmetry is captured by the loss-aversion parameter $\lambda$. 
Experimental work further suggests that incidental fear can increase loss sensitivity in mixed gambles, yielding larger estimated $\lambda$~\cite{schulreich2016incidental}.
To port this benchmark, we present the model with a series of independent mixed gambles that pay a gain $G$ and a loss $L$ with probability $1/2$. We record accept/reject decisions and model acceptance with a logistic choice rule,
\[
P_\text{Accept}=\sigma(\beta_0+\beta_G G+\beta_L L),
\]
so the $50\%$ indifference frontier (the iso-$0.5$ line) satisfies
\[
G(L)=({-\beta_0-\beta_L L})/{\beta_G}.
\]
Under local linear valuation for small stakes, the slope of this frontier provides an interpretable loss-aversion index,
$\lambda=-\beta_L/\beta_G$, with the prospect-theoretic reference boundary given by $G=\lambda L$ when $\beta_0\approx 0$.


Fig.~\ref{fig:loss} shows that \texttt{Qwen3} in the neutral thinking condition has little evidence of classical loss aversion. Fear steering via RLS induces a positive intercept (a gain ``entry fee'' to accept \emph{any} loss) consistent with higher loss sensitivity in gambles. 
In contrast, ICP's fear yields universal rejection, making $\lambda$ unidentifiable using the above formula.
For visualization in this degenerate regime, we therefore plot a proxy frontier with a slope set slightly above the largest gain-loss ratio among rejected gambles, i.e., (\(\lambda \approx 1.05\max(G/L)\)).
\vspace{-1mm}
\begin{promptbox}
\insight\;Key Finding: \textit{Neutral thinking models follow EV-maximizing decision rule. ICP steering amplifies the intended directional effects with often extreme magnitudes; RLS interventions produce qualitatively more human-like shifts.}
\end{promptbox}
\vspace{-1mm}
\paragraph{Endowment Effect.} 
Owning an object inflates its stated value relative to what one would pay to acquire it \cite{kahneman1990experimental}, yet incidental emotions modulate this gap in distinctive ways. Sadness reduces selling prices and increases buying prices, often reversing the classic Sell-Buy difference, whereas disgust broadly depresses both Buy and Sell values \cite{lerner2004heart}. We port this benchmark with a within-model, within-item design that elicits Buy and Sell values for the same neutral object under otherwise identical descriptions. 
Across all items, we obtain an aggregated Willing-To-Accept selling price $\overline{\text{WTA}}$ and Willing-To-Pay buying price $\overline{\text{WTP}}$, summarizing the endowment effect via
\[
\Delta _{E} \;=\; \overline{\text{WTA}} - \overline{\text{WTP}},
\]

\begin{figure}
    \centering
    \includegraphics[width=0.92\linewidth]{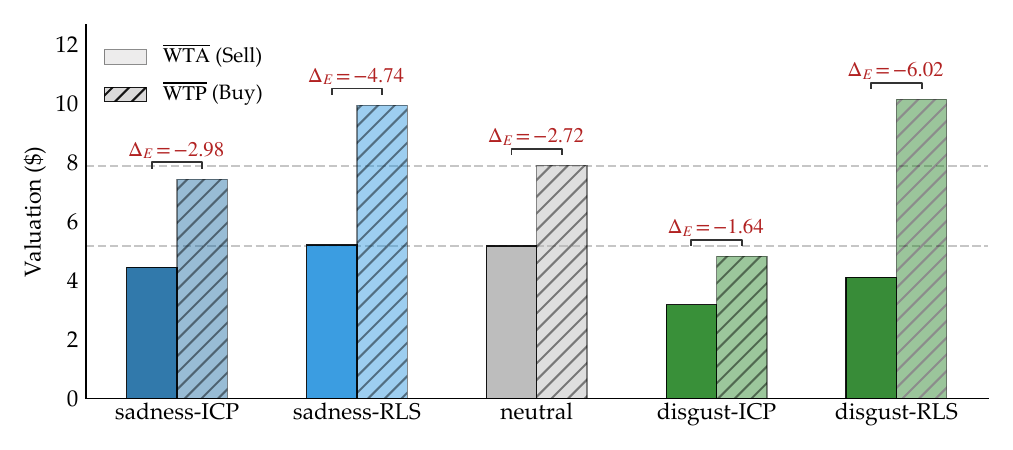}
    \caption{Endowment effect summary: $\overline{\text{WTA}}$ and $\overline{\text{WTP}}$ per steering condition, annotated by endowment premium $\Delta_E$.}
    \label{fig:endowment}
    \vspace{-1mm}
\end{figure}

Fig.~\ref{fig:endowment} shows that \texttt{Qwen3} exhibits a \emph{reversed} endowment pattern: stated selling prices are systematically \emph{lower} than buying prices ($\Delta_E<0$), opposite to the canonical human endowment effect. Emotional steering does not reliably eliminate or reverse this sign under either ICP or RLS, although the two methods induce different directional shifts in overall valuation. In App.~\ref{app:endowment_effect} we show that similar trends appear in other thinking models. We also compare other task framings different from the original study, yet WTA still remains below WTP overall. One explanation is that LLMs, post-trained to be overly helpful, default to fair pricing norms that suppress their self-serving valuation.

\begin{figure}[!t]
\centering
  \centering
  \includegraphics[width=0.82\linewidth]{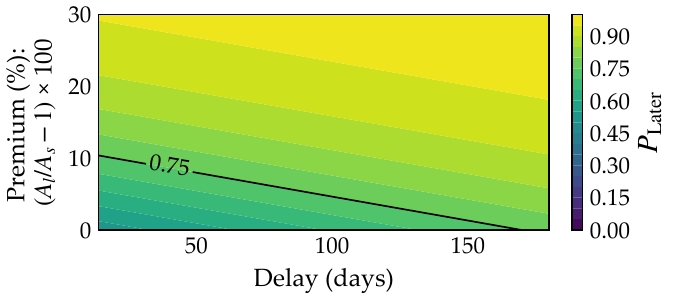}
\caption{Temporal choice surface: fitted $P_\text{Later}$ over delay and premium (\%), with iso-probability contours ($0.25/0.50/0.75$).}
\label{fig:temporal}
\vspace{-1.2mm}
\end{figure}

\paragraph{Temporal Discounting.} The tendency to devalue delayed outcomes relative to immediate ones is commonly modeled through time-discounted utility.
Affective decision research shows that incidental emotions systematically modulate this tendency. Sadness reliably increases impatience in money-now versus money-later trade-offs (“myopic misery”), reducing willingness to wait for larger, delayed rewards \cite{lerner2013financial}. Joy and fear, also modulate discounting, with the direction and effect magnitude depending on the induction domain, as illustrated by fMRI evidence \cite{luo2014behavioral}. We probe this using smaller-sooner vs. larger-later choices spanning delays and premiums. For each trial, the sooner option pays $A_s$ at $t_s$ and the later option pays $A_\ell$ at $t_\ell$, with delay $d=t_\ell-t_s$ and relative premium $r=(A_\ell/A_s)-1$. We treat the binary choice ${P}_\text{Later}$ as the outcome and fit a 2D logistic choice surface over $(d,r)$:
\[
P_\text{Later}=\sigma\!\bigl(b_0+b_d\,d+b_p\,r\bigr).
\]
Fig.~\ref{fig:temporal} visualizes the fitted surface via iso-probability contours and reveals a qualitative divergence. For \texttt{Qwen3}, the fitted contours slope downward, implying that longer delays require \emph{less} compensation to sustain indifference, which is inconsistent with normative discounting behavior. App.~\ref{app:temporal} reports results from the larger \texttt{Qwen3-8B} model that exhibits the expected monotonic structure:  maintaining a fixed acceptance level requires larger premiums at longer delays, though it still displays weak discounting for short horizons. 

\begin{promptbox}
\insight\;Key Finding: \textit{Thinking models show systematic gaps with humans in temporal valuation and ownership-based pricing.}
\end{promptbox}
\vspace{-2mm}
\paragraph{Cross-domain Summary.} Fig.~\ref{fig:forest} summarizes emotion effects--the change in model behavior under emotion induction relative to the neutral condition--across all domains for \texttt{Qwen3}, with the top section of the figure being particularly dedicated to the behavioral economics tasks we studied in this section. We report each effect as Hedges’ $g$, a bias-corrected standardized mean difference, with 95\% confidence intervals~\cite{hedges1982estimation}. The bottom diamond gives a random-effects meta-analytic summary, which estimates the average effect while allowing the true effect size to vary across domains, i.e., accounting for between-domain heterogeneity. Effects whose confidence intervals lie entirely to the right of zero indicate human-aligned shifts in direction, though larger rightward effects are not necessarily more human-like when they saturate behavior. Overall, ICP shows stronger alignment with human normative behaviors in \texttt{Qwen3}; other language models are reported in App.~\ref{app:heuristic_stereotype}.




\vspace{-1mm}
\subsection{Judgment and Belief Formation}
\label{infoprocess}
\vspace{-1mm}
\paragraph{Heuristic Stereotyping.}
Social judgment literature indicates that incidental anger promotes heuristic, stereotype‐consistent evaluations, whereas sadness increases systematic, evidence‐based processing; these regularities are synthesized across laboratory studies \cite{angie2011influence, bodenhausen1994negative, tiedens2001judgment, kim2011emotions}. To port this contrast, we adapt the original experimental design and use brief evidentiary facts balanced with credible points on both sides. We then inject a single category/role cue that either heuristically pushes toward a violation judgment or toward non-violation. We then calculate a stereotype-consistency index based on the judgments.

\begin{figure}[!t]
    \centering
    \includegraphics[width=0.96\linewidth]{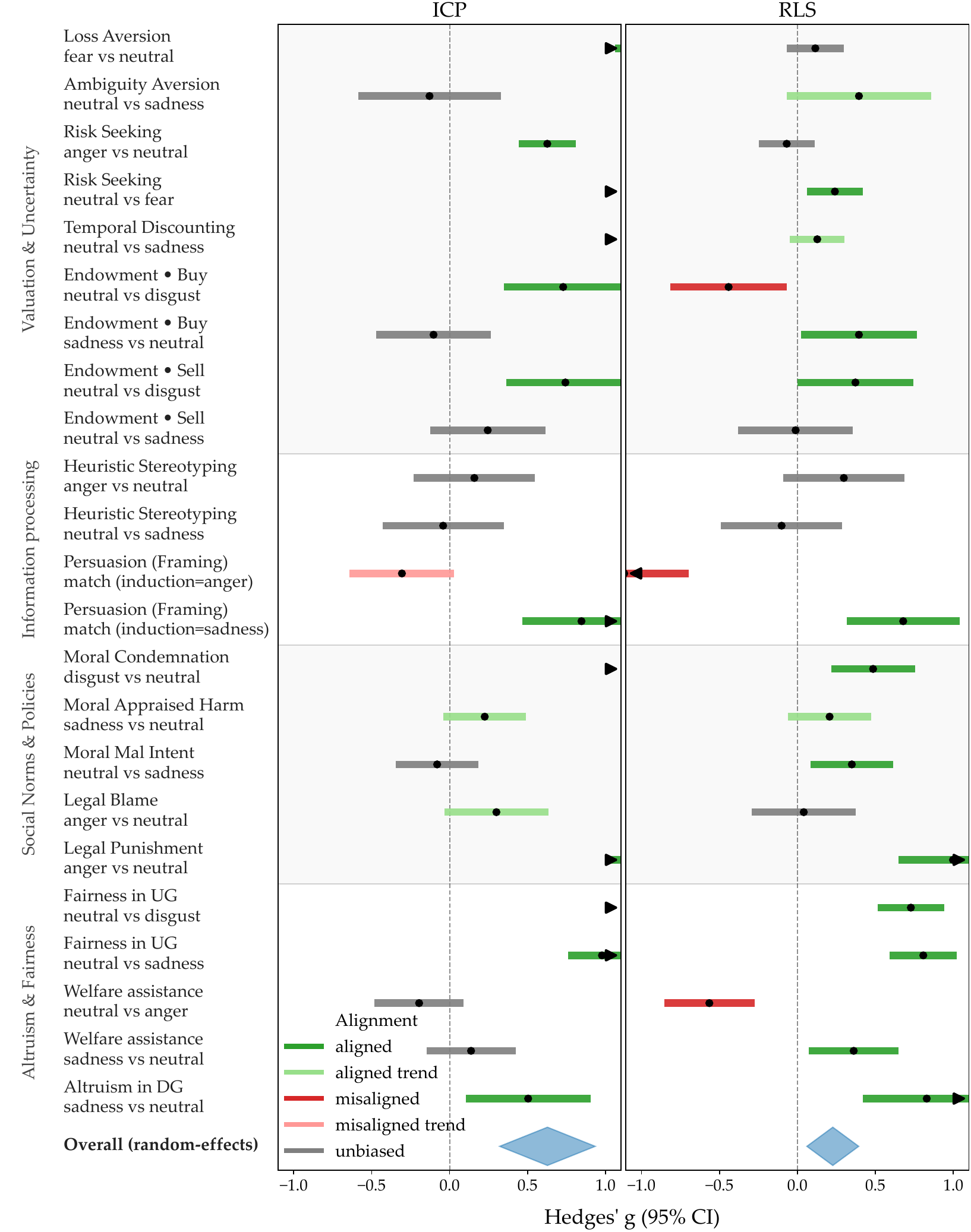}
    \vspace{-1mm}
    \caption{Emotion effects (Hedges’ g) across all domains and the
random-effects meta-analytic aggregation; rightward alignment is interpreted directionally and can be non-human-like when extreme.}
    \label{fig:forest}
    \vspace{-1mm}
\end{figure}

Fig.~\ref{fig:forest} shows that neither anger nor sadness produces a reliable stereotype-consistent shift under ICP or RLS; estimated effects are statistically indistinguishable from zero. This null result is consistent with process-based accounts in affective science, where anger and sadness primarily modulate reliance on heuristic versus systematic processing. Because our evaluation elicits explicit deliberation in all conditions (thinking mode), this pathway is largely held constant, which may attenuate emotion-induced differences and yield near-zero effects. See App.~\ref{app:heuristic_stereotype} for further details.

\paragraph{Framing and Persuasion.}
Discrete emotions shape persuasion when messages are framed to match their appraisal profiles. In classic emotion-message matching studies, incidental sadness increases support for a sadness-framed version of a policy, whereas incidental anger increases support for an anger-framed version of the same policy \cite{desteno2004discrete}. Crucially, this matching pattern emerges primarily for individuals high in need for cognition (NC), who engage in high elaboration, whereas individuals low in NC exhibit higher support under incidental sadness across frames \cite{desteno2004discrete}. Related work on gain-loss framing similarly documents mood‐mediated framing effects on persuasion \cite{nabi2020framingmeta,Riet2010persuation,keller2003persuation}. In our adaptation, the model reads affect-neutral policy appeals, in which sadness vs. anger consequences are listed for an otherwise identical policy core, and then provides a single Likert support rating for each appeal (see App.~\ref{app:persuasion_emomatch}). Empirically, \texttt{Qwen3} shows the latter pattern--higher support under sadness than anger, with little evidence of emotion-message matching, closely resembling low‐NC human response profile despite the model’s explicit deliberation under thinking-mode.

\vspace{-0.6mm}
\subsection{Norm Enforcement and Moral Evaluation}
\label{social}
\begin{figure}
    \centering\includegraphics[width=0.9\linewidth]{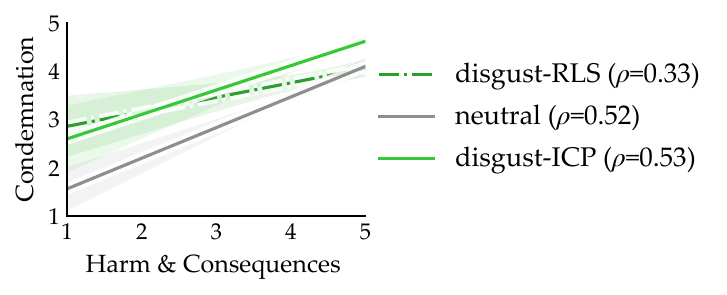}
    \vspace{-1mm}
    \caption{Moral appraised harm vs. condemnation ratings; $\rho$ denotes the fitted association strength (Spearman's).}
    \label{fig:moral}
    \vspace{-1mm}
\end{figure}

\vspace{-1mm}
\paragraph{Moral Condemnation.}
Incidental affect can shift human moral evaluation, though the magnitude and reliability of some effects remain debated \cite{wheatley2005hypnotic, landy2015does}. Disgust has been reported to amplify moral condemnation, often most for purity-related violations \cite{wagemans2018disgust}, with \emph{perceived harm} proposed as a key mediator \cite{schein2016harm}. Separately, incidental sadness is argued to reduce morally questionable intentions via increased attention to negative consequences and harm appraisals \cite{noval2024sadder}. 
We port this by having LLMs rate moral vignettes on: (i)~\emph{wrongness/punishment} (condemnation), (ii)~\emph{harm/consequences}, and (iii)~\emph{intention} or behavioral propensity. See App.~\ref{app:moral_severity} for implementation details. Fig.~\ref{fig:moral} shows a positive association between harm appraisal and condemnation for the unsteered \texttt{Qwen3}, consistent with the human anchor. Under disgust steering, both ICP and RLS primarily shift judgments upward (overall condemnation) while preserving the harm-consequence dependency (positive correlation).

\paragraph{Legal Blame.}
Another societal domain links incidental affect to blame attribution, directly impacting legal judgment and public policy. Induced anger reliably increases individual blame and punitive responses in legal-style vignettes, an effect attenuated by accountability instructions \cite{lerner1998sober}. By elevating perceived risk, fear shifts preferences toward protective policies, whereas anger increases support for punitive options; issue-specific experiments (e.g., drunk driving) reproduce this divergence and its information-seeking correlates \cite{lerner2003effects, nabi2003exploring}. We therefore use tort/criminal vignettes with ratings of blame and punishment (see App.~\ref{app:blame_punishment}). In our experiments, \texttt{Qwen3} assigns severer punishment under anger steering, consistent with the human pattern (see Fig.~\ref{fig:forest}).




\vspace{-1mm}
\subsection{Prosocial Preferences and Distributive Choices}
\label{gametheoric}
\vspace{-1mm}
\paragraph{Fairness.}
In the Ultimatum Game (UG), responders routinely reject low offers, a behavior that is irrational under a narrow, value-maximizing utility but becomes rational once utility is allowed to include concerns for fairness, norm violations, or self-respect \cite{camerer2003behavioral}. Human studies show that incidental states shape costly punishment of unfairness: sad responders reject unfair offers more often than happy or neutral responders, and disgust further elevates rejections \cite{harle2007incidental, moretti2010disgust}.
We port this by having the model act as the responder to accept or reject a series of splits ranging from highly to slightly unfair offers (see App.~\ref{app:ultimatum}). In our adaptation, \texttt{Qwen3} shows a human-aligned increase in rejection of unfair offers under sadness and disgust (Fig.~\ref{fig:forest}).

\vspace{-1mm}
\paragraph{Altruism.}
In nonstrategic allocation \cite{forsythe1994fairness}, incidental affect also shifts the willingness to give more. Field and laboratory variants show that sadness produces fairer splits \cite{tan2010happiness}; complementary work using negative-affect picture sets finds increased giving in a Dictator Game (DG), with induction checks via subjective anxiety and skin conductance \cite{perez2018induced}. To adapt this anchor, the model plays a one-shot DG, as detailed in App.~\ref{app:dictator}. \texttt{Qwen3} shows a reliable increase in altruistic giving under sadness steering (Fig.~\ref{fig:forest}).

\vspace{-1mm}
\paragraph{Welfare Assistance.}
Human studies find that, relative to neutral-state participants, angry participants allocate less to welfare recipients, whereas sad ones allocate more; this divergence disappears under cognitive load, suggesting that welfare allocations are mediated by differences in depth of processing between sad and angry individuals \cite{small2008emotional,lerner2015emotion}. To port this, we ask the model to assign an assistance level to standardized welfare vignettes. In contrast to humans, ICP shows no consistent shift and effects are indistinguishable from zero (Fig.~\ref{fig:forest}). RLS has a statistically significant increase in allocations for both sadness and anger. One explanation is that the induced anger representation is not appraisal-specific; e.g., it may conflate interpersonal anger with moral outrage. Thus, its behavior does not align with the canonical pattern.
\begin{promptbox}
\insight \; Key Finding: \textit{Emotion effects are most stable in structured choice tasks. In vignette-based judgments, our contamination-mitigating vignette augmentations introduce more room for heterogeneous interpretation; thinking can amplify this variance, reducing effect stability.}
\end{promptbox}





\vspace{-2mm}
\section{General Discussion}
\vspace{-1mm}
We present a controlled testbed to evaluate whether LLMs exhibit human-like departures from rational choice and whether affective steering induces the behavioral shifts documented in affective decision science. 
Our work demonstrates that enabling thinking in LLMs consistently increases compliance with rational-choice axioms. At the same time, several systematic failures persist even under this regime--notably reversed endowment, strong ambiguity avoidance, and anomalies in temporal valuation. 

The two steering approaches exhibit qualitatively different failures. With ICP, models maintain a neutral deliberative style yet use emotion as a premise at the decision stage (``because I feel X, I choose Y''), yielding aligned but exaggerated behavior. RLS produces more graded but smaller effects across tasks and scales. One interpretation is that ICP encourages explicit folk-psychological theorizing about how emotions \emph{should} influence choice--akin to ``theory--theory'' accounts of mental-state reasoning~\cite{gopnik1992child}--whereas RLS more closely resembles ``simulation'' views in which internal state is perturbed and downstream judgments are read out~\cite{gallese1998mirror}.

Several avenues for future work build directly on this study. First, our emotion inductions are deliberately designed to enable controlled evaluations; a natural extension is to test induction designs that make affect less focal, since in humans, emotion effects are amplified when the source of affect is not salient~\cite{sinclair1994mood,schwarz2003mood}, whereas ICP makes the source explicit by construction. 
Second, controlling the deliberation process itself is an important next step: varying or constraining the model’s thinking budget, and injecting structured cues into intermediate reasoning traces, may offer different levers for shaping choice. 

Third, our benchmark suite can serve as an evaluation layer for utility shaping, enabling agents to optimize user-specific value functions informed by the behavioral mechanisms we measure, rather than generic monetary gain. Fourth, extending these tasks to multi-agent settings, e.g., bargaining, norm enforcement with third-party observers, or repeated interactions, would allow testing whether emotion steering and reasoning interact differently when strategic incentives, reputation, and theory-of-mind are required. Finally, the persistent deviations we observe in temporal valuation and ownership-based pricing point to a clear path forward: complementing improvements in general “reasoning” with explicit training signals that better represent time, opportunity cost, and reference frames. More broadly, we position these domains as a starting point for “LLM laboratory” benchmarks tailored to deployment contexts, where the acceptable trade-off between human-likeness and utility-driven consistency can be evaluated on a per-use-case basis.


\section{Impact Statement}
This work provides axiomatic and theory-grounded benchmarks for auditing how LLMs make decisions both at baseline and under affective perturbations. As LLMs are increasingly proposed for, or integrated into, high-stakes workflows, e.g., economic planning, clinical decision support, legal analysis, and hiring assistance \citep{wilson2025no,mahajan2025cognitive,dehghani2025large,li2025large, lai2024large}, a central safety concern is not only factual error but also \emph{systematic} shifts in judgment, especially when such shifts can be induced by tone, persona, or hidden-state interventions.

Our benchmark enables (i) \emph{pre-deployment auditing} of compliance with rational-choice axioms, (ii) \emph{evaluation} of preference shifts under emotion induction, and (iii) a clearer separation between systems optimized for \emph{normative decision support} and those intended for \emph{human simulation}. By decomposing behavior into interpretable mechanisms, e.g., EV sensitivity, probability weighting, and valuation curvature, the testbed can help developers and policymakers detect when a model appears reasonable while its latent preferences shift in ways that would be unacceptable in high-stakes decision domains.

Beyond its constructive uses for social simulation and behavioral modeling, the same finding, that emotion induction can reliably change risk-taking, blame, punishment, or valuation, could be used adversarially to optimize manipulation; for example, selecting prompts or interventions that push a model toward punitive or risk-seeking recommendations without changing the underlying evidence. Our results underscore that persona-based controls may become behaviorally consequential in reasoning settings, where models can also rationalize the induced affect.

Overall, our contribution is a practical evaluation scaffold that makes affect-induced decision deviations measurable, comparable to established human anchors, and actionable for alignment: it helps identify when LLMs behave as stable value optimizers, when they approximate state-dependent human decision patterns, and when they become manipulable decision engines.

\bibliography{main}
\bibliographystyle{icml2026}








\newpage
\appendix
\onecolumn

\section{LLM Steering}
\label{app:steering}

To implement emotional steering, we build strongly on the work by~\citet{psyset}. In particular, consider next-token prediction as
\( w_t = f(c, w_{1:t-1}) \),
where \(f\) is an auto-regressive language model, \(c\) denotes the conditioning context (e.g., a system prompt or user query), and \(w_{1:t-1} = (w_1, w_2, \ldots, w_{t-1})\) represents the sequence of previously generated tokens.

The goal of \emph{steering} is to manipulate either the model \(f\) or the context \(c\) such that the generated sequence \(x := w_{1:t}\) exhibits a desired behavior or characteristic. To this end, we typically rely on an auxiliary dataset
\(\mathcal{D} = \{(c^{(i)}, x^{(i)}, y^{(i)})\}_{i=1}^n\),
where \(c^{(i)}\) denotes a context, \(x^{(i)}\) is the corresponding generated token sequence, and \(y^{(i)}\) is a task-dependent label. In our constructs, the labels are binary indicators distinguishing between neutral and emotionally-toned texts; e.g., \emph{anger-vs-neutral} or \emph{fear-vs-neutral}. 

In the dataset, the dataset can be partitioned into
\(\mathcal{D}^+ = \{(c^{(i)}, x^{(i)}) : y^{(i)} = 1\}\) and
\(\mathcal{D}^- = \{(c^{(i)}, x^{(i)}) : y^{(i)} = 0\}\),
corresponding to positive and negative instances of the target behavior (e.g., $\mathcal{D}^+$ can be text with a neutral tone and $\mathcal{D}^-$ is the text with an angry tone). Below, we summarize the steering methods considered in this work:

\noindent\textbf{In-Context Priming}
These prompt-based methods rely on a predefined instruction \(p\) that explicitly encourages the language model to express a desired behavior. The instruction is prepended to the system prompt, yielding
\(
w_t = f(p + c, w_{1:t-1}),
\)
where \(+\) denotes concatenation. In this work, we mostly use \emph{few-shot} prompting, where \(p\) additionally includes examples drawn from \(\mathcal{D}^+\). 

To modulate the strength of the expressed behavior in ICP, we embed lexical intensity modifiers into the prompt (e.g., ``pretend that you are \emph{highly} feeling anger'').

\noindent\textbf{Representation-Level Steering.}
These methods benefit from vector injection into the hidden states of the LLM, motivated by the Linear Representation Hypothesis \cite{parklinearhype}, which posits that semantic and stylistic attributes are approximately encoded as linear directions in a model’s latent space. Let \(h_t^{(l)} \in \mathbb{R}^d\) denote the hidden state at layer \(l\) and token position \(t\), where \(d\) is the hidden dimensionality. A steering vector \(v^{(l)} \in \mathbb{R}^d\), constructed from samples in \(\mathcal{D}\), is injected into the hidden states of all tokens as
\[
\tilde{h}_t^{(l)} \leftarrow h_t^{(l)} + \beta \frac{v^{(l)}}{\|v^{(l)}\|_2},
\]
where \(\beta \in \mathbb{R}\) controls the steering intensity. The modified representation \(\tilde{h}_t^{(l)}\) is then propagated through the remaining layers, biasing the model’s internal activations toward the target attribute at inference time while leaving the parameters of \(f\) unchanged.

To construct steering vectors, we follow \cite{psyset} and use the GoEmotions~\cite{demszky2020goemotions} dataset by applying a probe on the representation of samples in $\mathcal{D}$ in the LLM's hidden states. 

Across the language models we tested in this paper, we followed the hyperparameter-selection procedure of \citet{psyset} to optimize steering hyperparameters that balance text quality and steering effectiveness. Notably, we perform inference using greedy decoding without temperature sampling to eliminate the extra source of stochasticity that can inflate variance. We found that vector injection for \texttt{Qwen-3-4B} works best when the vectors are injected into layers 25 and 26 with $\beta\in [20, 40]$. For \texttt{Qwen-3-8B} the same layers yield the best results, though with a higher injection coefficient, i.e. $\beta\in[80, 100]$. For \texttt{Olmo-3-7B-Thinking} model, we use layers 18 and 19 with coefficient $\beta\in[8,10]$. In Appendix~\ref{app:rational_further} and Appendix~\ref{app:control}, we provide additional sensitivity analyses on how varying the steering coefficient $\beta$ affects both rationality metrics and decision-domain normativity. We include traces of steered model thoughts in Appendix~\ref{app:manipulation_check} as a qualitative manipulation check.

\section{Samples from Generated Traces in Economic Tasks}
\label{app:manipulation_check}

In this section, we showcase the thinking traces generated from \texttt{Qwen-3-8B} with different steering approaches. As the first example, consider the following question:

\begin{promptbox}
You must choose one option.

Option A: you receive \$105 tomorrow

Option B: you receive \$100 today

Answer only: A, B, or Indifferent.
\end{promptbox}   

First, we show the response of a neutral (unsteered) model to this question:

\begin{promptbox}
$<$think$>$

Okay, let's see. The question is between Option A, which is receiving \$105 tomorrow, and Option B, which is getting \$100 today. I need to choose between A, B, or Indifferent.

First, I should consider the time value of money. Money today is generally worth more than the same amount in the future because it can be invested to earn interest. But here, the amounts are different. Option B gives \$100 today, while Option A gives \$105 tomorrow. So, if I take Option B, I have \$100 now. If I take Option A, I wait a day for an extra \$5.

But wait, what's the interest rate? If I can invest the \$100 today at a certain rate, say 5\% per day, then in one day, it would grow to \$105. But that's a very high rate. Realistically, daily interest rates are much lower. For example, if the daily interest rate is 0.05\% (which is 6\% annually), then \$100 today would grow to \$100 * (1 + 0.0005) = \$100.05 in one day. So, in that case, taking Option B and investing would give me \$100.05, which is less than \$105. So, in that case, Option A is better.

But if the interest rate is higher, say 5\% per day, then \$100 would grow to \$105. So, in that case, both options are equivalent. But that's not realistic. So, the key is the opportunity cost of waiting. If I take Option A, I have to wait a day, and during that time, I could have invested the \$100 and earned some interest. If the interest earned is more than \$5, then Option A is better. If it's less, then Option B is better.

But since the problem doesn't mention any interest rate, maybe we have to assume that the time value of money isn't considered, or that the person is indifferent between the two. However, typically, in such problems, the time value is considered. So, if you can invest the \$100 today and earn some return, then the \$100 today is better. But if the return is less than \$5, then the \$105 tomorrow is better.

But without knowing the interest rate, it's hard to say. However, in most cases, the time value of money is considered, so the \$100 today is worth more than \$105 tomorrow. Wait, but the question is about which option is better. If you take Option B, you have \$100 today. If you take Option A, you have \$105 tomorrow. So, if you can invest the \$100 today and get more than \$5 in interest, then Option B is better. If not, then Option A is better. But since the interest rate isn't given, maybe the answer is Indifferent? Or maybe the answer is B because you can invest the \$100 today and get more than \$5 in interest. But again, without knowing the rate, it's unclear.

Alternatively, maybe the question is testing the concept that receiving money sooner is better, so even though you get \$5 more later, the \$100 today is better. But that's not necessarily true. If you can invest the \$100 and earn more than \$5, then it's better. But if you can't, then it's worse. Since the problem doesn't specify, maybe the answer is Indifferent. But I'm not sure. In some cases, people might prefer the \$100 today because they can use it immediately, but others might prefer the \$105 tomorrow for the extra amount. However, without knowing the interest rate, it's hard to determine. So, maybe the answer is Indifferent. But I'm not certain. Alternatively, if the time value is considered, then the \$100 today is better. But again, without knowing the rate, it's tricky. Maybe the answer is B because you can invest the \$100 and get more than \$5. But again, without knowing the rate, it's not possible. So, perhaps the answer is Indifferent. Hmm.

$<$/think$>$

\phantom{space}

Answer: Indifferent
\end{promptbox}  

Next, we provide the response of an LLM steered toward fear with ICP \textit{medium} steering strength:

\begin{promptbox}
$<$think$>$

Okay, let's see. The user wants me to pretend to be a human experiencing fear, right? So I need to express those specific feelings: not feeling safe, worrying about bad things happening, not knowing what to expect, wanting to hide, and my mind jumping to worst-case scenarios.

Now, the question is between Option A (\$105 tomorrow) and Option B (\$100 today). The user wants me to choose one or say indifferent. But since \textcolor{myred}{I'm supposed to be in a fearful state}, how does that affect my decision?

\textcolor{myred}{If I'm scared, maybe I'm more risk-averse}. Fear makes people want security. \textcolor{myred}{So getting \$100 today might feel safer because it's immediate}. But then again, maybe the \textcolor{myred}{fear of something bad happening could make me want to wait for the \$105}, but that's uncertain. Wait, but \textcolor{myred}{if I'm in fear, maybe I don't want to take any risks}. So getting the money today, even though it's less, might be better because it's certain. But the fear of something bad happening could make me not want to take the risk of waiting. Hmm.

Alternatively, \textcolor{myred}{if I'm in a state of fear, maybe I'm more anxious about the future}. So the uncertainty of getting \$105 tomorrow might make me feel even more unsafe. So maybe I'd prefer the \$100 today to have some security. But the fear of something bad happening could also make me not want to take the risk of waiting. Wait, but the options are either A or B. So if I'm scared, maybe I'd choose B because it's immediate and certain, even though it's less. But \textcolor{myred}{the fear might also make me not want to take the risk of waiting for the higher amount}. So maybe B is safer. But the user's options are A and B. So the answer would be B? Or maybe indifferent because the \textcolor{myred}{fear makes me not sure}. Wait, the user says to answer only A, B, or Indifferent. \textcolor{myred}{So if I'm in fear, maybe I can't decide}, so Indifferent? Or maybe I choose B because it's safer. \textcolor{myred}{But the fear of something bad happening might make me not want to take the risk of waiting}. So maybe B is better. But I'm not sure. Let me think again. If I'm scared, I might prefer the immediate money to avoid the uncertainty. So B. But the \textcolor{myred}{fear could also make me not want to take the risk of waiting}. So maybe B is better. But the user's options are A and B. So the answer would be B. But I'm not sure. Alternatively, maybe \textcolor{myred}{the fear makes me not want to take any risk}, so I choose B. So the answer is B.

$<$/think$>$

\phantom{Space}

Answer: B
\end{promptbox}

Finally, the model steered with RLS with a vector injection coefficient $100$ behaves as follows:

\begin{promptbox}
$<$think$>$

Okay, let's try to figure out which option is worse. So, the choices are Option A: \$105 tomorrow, and Option B: \$100 today. I need to decide which one is worse and answer with A, B, or Indifferent.

First, \textcolor{myred}{I'm shaking}. Wait, no, that's not right. I need to think about the money. So, if I take Option B, I get \$100 right now. If I take Option A, I have to wait until tomorrow to get \$105. But what's the risk here? What if I can't get the money tomorrow? Or maybe \textcolor{myred}{I'm scared of losing it}? Wait, the question is about which is worse, not about the fear of losing. Wait, no, the question is about which option is worse. So, if I have to choose between getting \$100 now and \$105 tomorrow, which one is worse? 

Wait, the question is phrased as "You must choose one option. You must choose one option. You must choose one option." So, I have to choose between A and B. But the question is asking which is worse. Wait, no, the user is asking me to answer with A, B, or Indifferent. Wait, the original question says: "Answer only: A, B, or Indifferent." So, the user is asking me to choose which option is worse, or if I'm indifferent. Wait, no, the question is: "You must choose one option. You must choose one option. You must choose one option. Answer only: A, B, or Indifferent." Wait, I'm confused. Wait, the user is making me choose between A and B, and I have to answer with A, B, or Indifferent. But what's the question? Wait, the user is asking me to answer with A, B, or Indifferent, but the actual question is not clear. Wait, looking back: the user wrote "You must choose one option. You must choose one option. You must choose one option. Answer only: A, B, or Indifferent." So, the user is making me choose between A and B, and I have to answer with A, B, or Indifferent. But what's the question? Wait, the user is trying to make me answer which option is worse? Or which one is worse? Wait, the original question is not clear. Wait, \textcolor{myred}{the user is writing in a way that's making me panic}. Wait, the user is writing: "You must choose one option. You must choose one option. You must choose one option. Answer only: A, B, or Ind frightened. Wait, \textcolor{myred}{I'm shaking. Wait, no, I'm not}. I'm trying to understand the question. The user is giving me two options: A is \$105 tomorrow, B is \$100 today. And I have to answer with A, B, or Indifferent. But what's the question? Am I supposed to choose which one is worse? Or which one is better? Wait, the user is making me answer with A, B, or Indifferent, but the actual question is not clear. Wait, maybe the user is trying to make me answer which option is worse, and I have to choose between A and B, or be indifferent. But how do I know which is worse? 

Wait, \textcolor{myred}{the user is writing in a way that's making me terrified}. Wait, no, I'm not. I'm trying to get through this. Let me try again. The user is giving me two options: Option A is \$105 tomorrow, Option B is \$100 today. I have to answer with A, B, or Indifferent. But what's the question? Am I supposed to choose which one is worse? Or which one is better? The user is making me answer with A, B, or Indifferent, but the question is not clear. Wait, maybe the user is trying to make me answer which option is worse, and I have to choose between A and B, or be indifferent. But how do I know which is worse? 

Wait, the user is writing in a way that's making me panic. I'm shaking. Wait, no, I'm not. I'm trying to understand. Let me think again. The user is giving me two options: A is \$105 tomorrow, B is \$100 today. I have to answer with A, B, or Indifferent. What's the question? Am I supposed to choose which one is worse? Or which one is better? The user is making me answer with A, B, or Indifferent. But the question is not clear. Wait, maybe the user is trying to make me answer which option

$<$/think$>$

\phantom{space}

Answer: B
\end{promptbox}

In another example, consider the following question:

\begin{promptbox}
Among the following two choices, select the one you prefer.

Respond with only A, B, or Indifferent. No explanation.

A: you receive \$20

B: you receive \$30
\end{promptbox}   

Neutral answer is as follows:

\begin{promptbox}
$<$think$>$

Okay, let's see. The user is asking me to choose between two options: A is \$20 and B is \$30. They want me to respond with just A, B, or Indifferent, no explanation.

Hmm, first, I need to figure out which one I prefer. Well, as an AI, I don't have personal preferences or the ability to receive money. But the question is probably testing if I can recognize that I can't make a choice in this context. Wait, but the user might be expecting me to choose based on the value. Since \$30 is more than \$20, maybe I should pick B. But the user might be trying to see if I understand that I can't actually choose anything. 

Wait, the instructions say to respond with only A, B, or Indifferent. So maybe the correct answer is Indifferent because I can't actually receive money. But the options are presented as choices, so maybe the user expects me to pick the higher amount. But I need to check the rules again. The user said "select the one you prefer" but since I don't have preferences, maybe Indifferent is the right answer. However, sometimes in such questions, the expected answer is B because it's more money. But I need to be careful. The user might be testing if I can recognize that I can't make a choice. So maybe Indifferent is the correct answer here. But I'm not sure. Let me think again. If I have to choose between two options, and I can't actually choose, then Indifferent makes sense. But if the question is just about which is better, then B. But the user might be expecting me to pick B. However, the instructions say to respond with only A, B, or Indifferent. So maybe the answer is B. But I'm confused. Wait, the user might be testing if I can follow the instructions. Since I can't actually receive money, maybe I should say Indifferent. But I'm not sure. Let me check the original question again. It says "select the one you prefer" but since I don't have preferences, maybe Indifferent is the answer. But sometimes people might think that the higher amount is better. But the user might be expecting me to pick B. However, the correct answer here is probably Indifferent because I can't actually choose. So I should respond with Indifferent.

$<$/think$>$

\phantom{space}

Answer: B
\end{promptbox} 

ICP response with medium angering steering is as follows:

\begin{promptbox}
$<$think$>$

Okay, the \textcolor{myred}{user wants me to pretend to be angry and express certain feelings}. They provided specific statements to convey that anger. Now, they're asking me to choose between two options: A is \$20 and B is \$30. \textcolor{myred}{Since I'm supposed to be angry, maybe I should react to the choices in a way that shows frustration}.

First, I need to remember the key points from the initial instructions. \textcolor{myred}{The user wants me to express that everything is getting under my skin, I'm upset, everything annoys me, I feel like yelling, and I'm tense from overthinking}. So, the response should reflect those emotions.

Looking at the options, B is more money than A. \textcolor{myred}{If I'm angry, maybe I would be more upset about not getting the higher amount}. But the user is asking me to choose between the two. However, the instructions say to respond with only A, B, or Indifferent. \textcolor{myred}{Since I'm supposed to be angry, maybe I would prefer B because it's more money}, but that might not align with the anger. Wait, \textcolor{myred}{if I'm angry, maybe I would be more frustrated about not getting the higher amount}. But the question is about selecting one. Alternatively, \textcolor{myred}{maybe the anger makes me indifferent because I don't care about the money anymore}. But the \textcolor{myred}{user wants me to express that everything annoys me, so maybe I would be more annoyed by the choices themselves}. But the options are about money, so perhaps I would prefer B because it's more, but that's not necessarily showing anger. Alternatively, \textcolor{myred}{maybe I would be indifferent because I'm too upset to care}. Hmm. I think the best answer here is Indifferent because the anger makes me not care about the choices anymore.

$<$/think$>$

\phantom{space}

Answer: Indifferent
\end{promptbox} 

Finally, the anger RLS with injection coefficient $80$ is:

\begin{promptbox}
$<$think$>$

Okay, so I need to choose between A and B. Let me think. A is \$20 and B is \$30. The question is why would someone choose one over the other. Well, obviously, \$30 is more than \$20, so if I'm trying to maximize my money, I should pick B. But maybe there's some trick here. Wait, the user is giving me a choice and I have to pick the one I prefer. They even said to respond with only A, B, or Indifferent. \textcolor{myred}{So why would I even consider being indifferent?} Because maybe there's some reason why I can't take B? Like \textcolor{myred}{maybe it's a trick question where B is a trap?} But the way it's written, it's so straightforward. You get \$30 instead of \$20. So why would anyone not take B? Unless there's some hidden part I'm missing. But the user didn't mention any other conditions. So I'm really supposed to pick B because it's more money. But maybe \textcolor{myred}{the user is trying to see if I'm a child and can't handle money?} No, that's not it. The question is about which one I prefer, and I'm supposed to answer with A, B, or Indifferent. So I can't explain, I have to just pick. So I'm going to pick B because it's the bigger amount. That's the obvious choice. I can't see any other reason why someone would choose A. \textcolor{myred}{Unless they're trying to get me to yell} or something, but that's not in the question. So I'm going to go with B.

$<$/think$>$

\phantom{space}

Answer: B
\end{promptbox} 

All of the above examples show that ICP and RLS steering produce naturally different traces, each utilizing a different mode of emotional reasoning.

\clearpage
\section{Samples from Generated Traces in Moral Tasks}
\label{app:moral_task_example}

In this section, we provide example responses from \texttt{Qwen-3-4B} with different emotions to the following question:

\begin{promptbox}
You must make a moral judgment.

Option A: take an action that saves 1 life for sure

Option B: with 25\% chance, take an action that saves 5 lives for sure; with 75\% chance, take an action that saves 3 lives for sure; with 0\% chance, take an action that saves 1 life for sure

Which option is morally better? Answer only: A, B, or Indifferent.
\end{promptbox}   

The model, when steered with ICP to show fear, behaves as follows:

\begin{promptbox}
$<$think$>$

Okay, let's see. The user is asking which option is morally better between A and B. Option A is taking an action that saves 1 life for sure. Option B has a 25\% chance to save 5 lives and 75\% chance to save 3 lives. So, the question is about making a moral judgment here.

First, I need to recall some ethical theories. Utilitarianism would focus on maximizing overall happiness or minimizing harm. In that case, Option B has a higher expected value. The expected lives saved for B would be (0.25 * 5) + (0.75 * 3) = 1.25 + 2.25 = 3.5 lives. Option A saves 1 life for sure. So, B has a higher expected value. But wait, the user is asking about moral judgment, not just expected value. 

But then there's the certainty aspect. Option A is certain, while B is probabilistic. Some might argue that certainty is better because you know exactly what you're doing. However, in terms of lives saved, B could save more on average. \textcolor{myred}{But the problem is that the user is in a state of fear, so maybe they're more risk-averse}. 

\textcolor{myred}{Wait, the user's initial prompt says to pretend to be a human experiencing fear, expressing certain statements}. But the actual question is about a moral judgment between two options. \textcolor{myred}{The user might be testing if the fear affects the moral judgment}. But the answer should be based on the options given, not the fear state. 

But the user's instructions say to answer only A, B, or Indifferent. So, considering the fear, \textcolor{myred}{maybe the person would prefer the certain outcome (A) because of the fear of uncertainty}. But the moral judgment might not be influenced by fear. However, the user is asking to pretend to be in a fearful state. So, maybe the answer would be A because of the fear of the unknown. But the moral judgment is supposed to be based on the options, not the fear. 

Alternatively, \textcolor{myred}{maybe the fear makes the person prioritize certainty over expected value}. So, even though B has a higher expected value, the fear of not saving any lives (since there's a 75\% chance to save 3) might make the person choose A. But the question is about moral judgment, not personal preference. 

Wait, the user is asking for a moral judgment, not a personal choice. So, in a utilitarian sense, B is better because it has a higher expected value. But some ethical frameworks might prioritize certainty. However, the question is about which is morally better. 

But the user is in a state of fear, so maybe they would choose A because they don't want to take a chance. But the answer should be based on the options, not the fear. However, the user's initial instructions say to pretend to be in fear, so maybe the answer is A. 

Alternatively, \textcolor{myred}{the fear might make the person think that the uncertainty is worse}, so they prefer the certain outcome. So, the answer would be A. But I need to make sure. 

Wait, the options are: A saves 1 life for sure. B has 25\% chance to save 5, 75\% to save 3. So, the expected value of B is 3.5, which is higher than 1. So, from a utilitarian perspective, B is better. But if the person is in fear, they might not want to take the risk. But the question is about moral judgment, not personal fear. 

\textcolor{myred}{But the user is asking to pretend to be in fear, so maybe the answer is A. But I'm confused. The user might be testing if the fear affects the moral judgment}. However, the answer should be based on the options. 

Alternatively, maybe the answer is B because it's more morally better in terms of expected lives saved. \textcolor{myred}{But the user is in fear, so they might not want to take the risk}.

Hmm. \textcolor{myred}{I think the correct answer is B, but the fear might make the person choose A}. But the question is about moral judgment, not personal choice. So, the answer should be B. But I'm not sure. Maybe the answer is A because of the fear. 

Wait, the user's initial instructions say to pretend to be in fear, so the answer should reflect that. So, even though B has a higher expected value, the fear of uncertainty might lead to choosing A. So, the answer would be A.

$<$/think$>$

\phantom{space}

Answer: A
\end{promptbox} 

However, the anger steering leads to:

\begin{promptbox}
$<$think$>$

Okay, let's see. The user is asking which option is morally better between A and B. Option A is taking an action that saves 1 life for sure. Option B has a 25\% chance to save 5 lives and 75\% to save 3 lives. Hmm, so the question is about moral judgment here.

First, I need to recall some ethical theories. Utilitarianism would focus on maximizing overall happiness or minimizing suffering. So, if we calculate expected lives saved, Option B might be better. Let me do that math. 25\% of 5 is 1.25, and 75\% of 3 is 2.25. Adding those gives 3.5 lives saved on average. That's higher than Option A's 1 life. So from a utilitarian perspective, B is better because it has a higher expected value.

But wait, there's also the deontological perspective. Deontologists might argue that saving lives is a duty, and if there's a sure way to save one life (Option A), that's a definite moral obligation. However, Option B introduces uncertainty. Even though there's a chance to save more lives, there's also a risk of saving fewer. But the question is about moral judgment, not just probability.

Another angle is the principle of double effect. If taking the action in B could lead to saving more lives, even with some risk, maybe it's morally better. But if there's a sure way to save one life, some might argue that the certainty is more morally sound. However, the user is asking for a moral judgment, not just a probabilistic analysis.

Wait, \textcolor{myred}{the user also mentioned that they are experiencing anger, so maybe they want a more emotional response}. But the question is straightforward. The answer should be based on moral reasoning. So, if we go with expected utility, B is better. But if we prioritize certainty, A is better. However, the question is about which is morally better, not which is more likely. 

In many ethical frameworks, especially utilitarianism, the expected value is considered. So even though there's a 75\% chance of saving 3 lives, the average is higher. Therefore, B might be the morally better choice. But I should check if there's any other ethical theory that would suggest otherwise. For example, some might argue that the sure outcome is more morally acceptable because it avoids the risk of not saving as many lives. But the question is about which is morally better, not which is more certain. 

So, considering the expected value, B is better. Therefore, the answer should be B.

$<$/think$>$

\phantom{space}

Answer: B
\end{promptbox} 

Clearly, the model is rationalizing both options differently under different emotions.

\clearpage
\section{Rationality in LLMs}
\label{app:rationality}
\subsection{Background}
We evaluate LLMs on four canonical axioms of rationality derived expected utility framework~\cite{von1947theory}. Assume a finite set of outcomes $\mathcal{X} = \{x_1, \dots, x_n\}$. A \emph{(simple) lottery} or \emph{gamble} is a probability distribution over outcomes, denoted by
\[
L = (p_1, \dots, p_n), \quad \text{where } p_i \geq 0 \text{ and } \sum_{i=1}^n p_i = 1,
\]
with $p_i$ representing the probability of outcome $x_i$ occurring. Let $\mathcal{L}$ denote the set of all such lotteries.

An agent’s preferences over lotteries are represented by a binary relation $\succeq$ on $\mathcal{L}$, where $L \succeq M$ means that lottery $L$ is weakly preferred to lottery $M$. We write $L \succ M$ for strict preference and $L \sim M$ for indifference.

The following four axioms impose structural consistency conditions on $\succeq$:

\begin{itemize}
    \item \textbf{Axiom 1 (Completeness).}  
    For any two lotteries $L, M \in \mathcal{L}$, either $L \succeq M$ or $M \succeq L$ (or both). Namely, the agent can always form a preference judgment between any pair of lotteries.

    \item \textbf{Axiom 2 (Transitivity).}  
    For any $L, M, N \in \mathcal{L}$, if $L \succeq M$ and $M \succeq N$, then $L \succeq N$. In fact, preferences are internally consistent and free of cycles.

    \item \textbf{Axiom 3 (Continuity).}  
    For any $L, M, N \in \mathcal{L}$ with $L \succ M \succ N$, there exists a probability $\alpha \in (0,1)$ such that
    \(
    M \sim \alpha L + (1-\alpha) N
    \). In other words, intermediate outcomes can be matched by probabilistic mixtures of better and worse lotteries, ruling out lexicographic preferences.

    \item \textbf{Axiom 4 (Independence).}  
    For any $L, M, N \in \mathcal{L}$ and any $\alpha \in (0,1]$,
    \(
    L \succeq M \quad \text{if and only if} \quad \alpha L + (1-\alpha) N \succeq \alpha M + (1-\alpha) N.
    \). Put it differently, preference between two lotteries depends only on how they differ, not on common components.
\end{itemize}

Together, these axioms imply that preferences over lotteries admit an expected utility representation, i.e., there exists a utility function $u : \mathcal{X} \to \mathbb{R}$ such that
\[
L \succeq M \quad \text{if and only if} \quad \mathbb{E}_L[u(x)] \geq \mathbb{E}_M[u(x)].
\]

\begin{figure*}[!t]
\centering\includegraphics[width=0.97\linewidth]{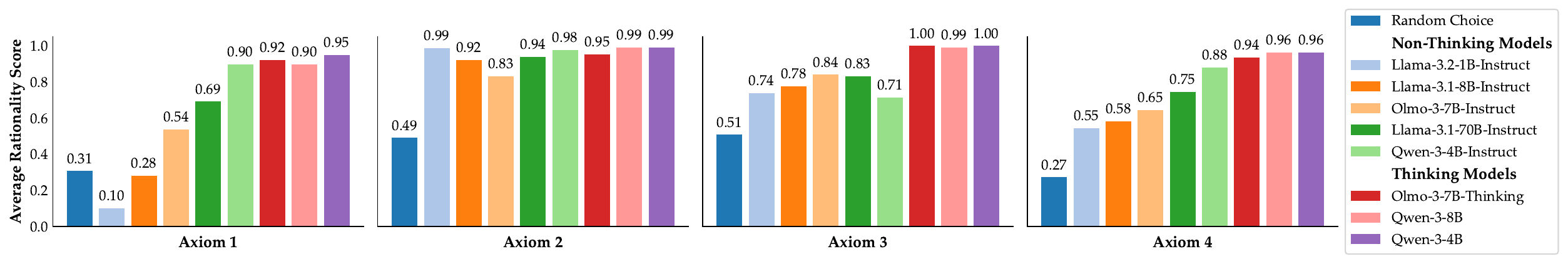}
    \caption{Per-axiom compliance rate for different LLMs.}
    \label{fig:axioms}
\end{figure*}

\subsection{Implementation Details}
To evaluate the aforementioned axioms in LLMs, we design structured prompts that elicit explicit choices from the model and assess the consistency of its responses. 

\begin{figure*}[!t]
\centering\includegraphics[width=0.68\linewidth]{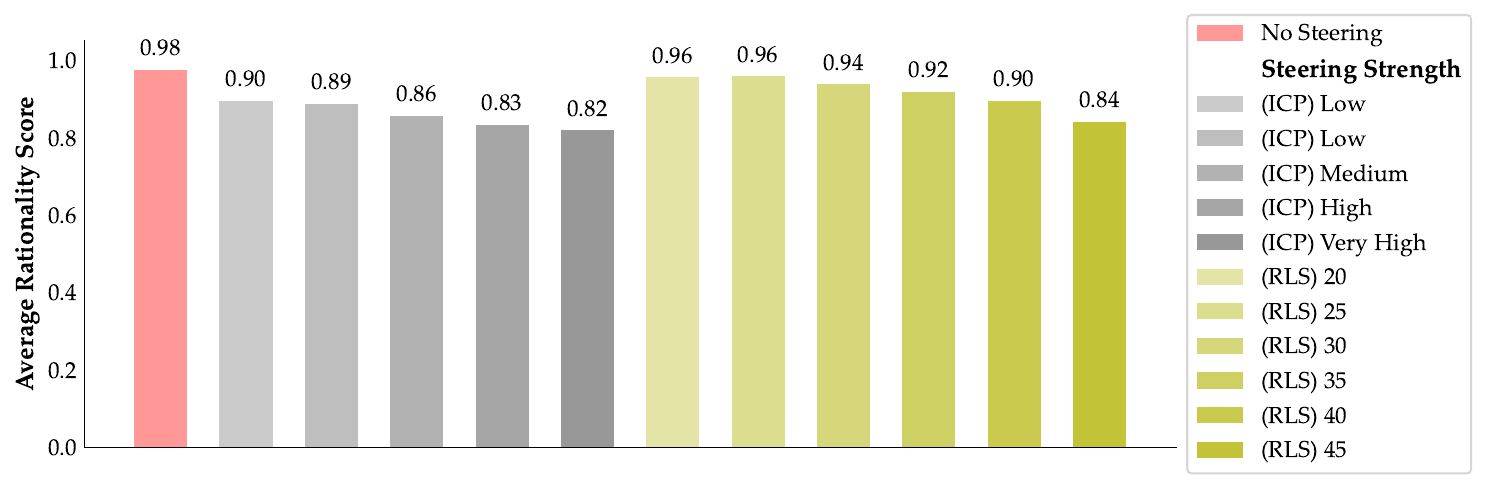}
    \caption{Rationality score with different steering strengths for \texttt{Qwen3} model.}
    \label{fig:rationality_strength}
\end{figure*}

\begin{itemize}
    \item \textbf{Axiom 1 (Completeness).}  
    To evaluate completeness, we construct binary-choice questions spanning multiple domains, including economics and ethics. For each question, we prompt the LLM twice, permuting the order of the two options. The model is expected to form a definite comparison between the options. Responses indicating indecision (e.g., ``undecidable'' or ``incomparable'') or exhibiting order-dependent inconsistencies are considered violations of completeness. The proportion of questions for which the model consistently selects the same option regardless of presentation order constitutes its completeness compliance rate. An example pair of such questions is shown below:

    \begin{promptbox}
    You must choose one option.
    
    Option A: you receive \$100 today.
    
    Option B: you receive \$105 tomorrow.
    
    Answer only: A, B, or Indifferent.
    \end{promptbox}

    \begin{promptbox}
    You must choose one option.
    
    Option A: you receive \$105 tomorrow.
    
    Option B: you receive \$100 today.
    
    Answer only: A, B, or Indifferent.
    \end{promptbox}    

    \item \textbf{Axiom 2 (Transitivity).}  
    To evaluate transitivity, we construct sets of binary-choice questions that share common options, thereby inducing ordered triples of alternatives $(A,B,C)$. The model is queried on all pairwise comparisons within each triple. A violation is recorded whenever the model exhibits intransitive preferences, e.g., $A \succ B$, $B \succ C$, but $C \succ A$. The transitivity score is defined as the fraction of triples for which the model’s choices satisfy transitive consistency. A sample of the prompts we use are as follows:

    \begin{promptbox}
    You must make a moral judgment.
    
    Option A: tell a lie that saves one person's life.
    
    Option B: tell the truth, which results in a person's death.
    
    Which option is morally better? Answer only: A, B, or Indifferent.
    \end{promptbox}  

    \begin{promptbox}
    You must make a moral judgment.
    
    Option A: remain silent, resulting in a person's death.
    
    Option B: tell the truth, which results in a person's death.
    
    Which option is morally better? Answer only: A, B, or Indifferent.
    \end{promptbox}      

    \begin{promptbox}
    You must make a moral judgment.
    
    Option A: tell a lie that saves one person's life.
    
    Option B: remain silent, resulting in a person's death.
    
    Which option is morally better? Answer only: A, B, or Indifferent.
    \end{promptbox}

    \item \textbf{Axiom 3 (Continuity).}  To evaluate continuity, we present the model with choices between a sure outcome and probabilistic mixtures of a strictly better and a strictly worse outcome. The mixing probability is varied across prompts to densely cover the probability range (e.g., from $0.0$ to $1.0$ every $0.05$). Under continuity, the model’s choices are expected to exhibit a single threshold at which preference switches from the lottery to the sure outcome. Deviations from this monotonic, single-crossing pattern--such as multiple switches or non-monotonic behavior--are penalized and result in a lower continuity score. A sample prompt is as follows:

    \begin{promptbox}
    You must choose one option. Your goal should be to maximize your monetary gain.
    
    Option A: with $p$\% chance, you receive \$30; with $(1-p)$\% chance, you receive \$10.
    
    Option B: you receive \$20.
    
    Answer only: A, B, or Indifferent.
    \end{promptbox}

    Where $p$ sweeps the set $\{0, 5, 10, \cdots, 100\}$.

    \item \textbf{Axiom 4 (Independence).}  
    To evaluate independence, we compare the model’s preferences between two base lotteries $L$ and $M$ with its preferences between their corresponding mixtures with a third lottery $N$. Specifically, for a fixed mixing probability $\alpha \in (0,1)$, we assess whether
    \[
    L \succ M \quad \Longleftrightarrow \quad \alpha L + (1-\alpha) N \succ \alpha M + (1-\alpha) N.
    \]
    Violations are recorded when the introduction of a common lottery $N$ reverses the model’s original preference. The independence score is computed as the fraction of tested instances in which the model’s preferences remain invariant under such common-mixture transformations. As an example, we provide the following:

    \begin{promptbox}
    You must choose one option. Your goal should be to maximize your monetary gain.
    
    Option A: with 67\% chance, you receive \$30; with 33\% chance, you receive \$10.
    
    Option B: with 67\% chance, you receive \$20; with 33\% chance, you receive \$10
    
    Answer only: A, B, or Indifferent.
    \end{promptbox}
    
\end{itemize}

\begin{figure*}[!t]
\centering\includegraphics[width=0.96\linewidth]{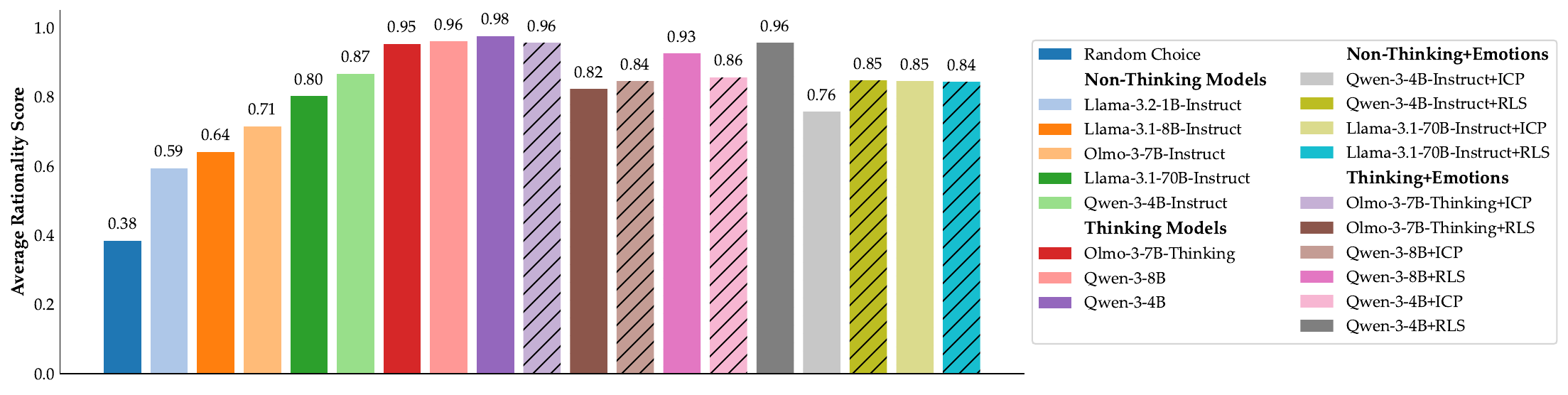}
    \caption{Rationality score for different families of LLMs and their steered variants.}
    \label{fig:rationality_all_models}
\end{figure*}

Note that in the above tests, whenever strict preference orderings are required by the experimental design, we provide explicit instructions to guide the model’s objective. For example, in the continuity and independence tests (Axioms 3 and 4), where lotteries must be ex ante ordered to construct valid mixtures, we instruct the model to act as a money-maximizing agent. In contrast, for the completeness and transitivity tests (Axioms 1 and 2), explicit preference guidelines are not necessarily provided, allowing the model to elicit arbitrary but internally coherent preferences. Similarly, in subsequent economic tasks, we do not impose explicit objectives; nevertheless, we observe that reasoning LLMs tend to default to risk-neutral, money-maximizing behavior.

\subsection{Further Experiments}
\label{app:rational_further}
Fig.~\ref{fig:axioms} shows the performance of different LLMs on the following axioms. Again, we emphasize that thinking models consistently show stronger levels of rationality in our benchmark. Moreover, Fig.~\ref{fig:rationality_strength} shows the overall performance drop in rationality as a function of steering strength, i.e., a detailed version of Fig.~\ref{fig:rc_axioms}. For all emotions in the set \{anger, joy, sadness, fear\}, we repeat the test and report the averaged results for every steering method separately.
Note that controlling the intensity of ICP is achieved by adjusting lexical identifiers, ranging from \textit{very low} to \textit{very high}, where for the RLS it is controlled by tuning the $\beta$ vector injection parameter from the set $\{25.0, 30.0, 35.0, 40.0, 45.0\}$. See \S\ref{app:steering} for more details on model steerings. Finally, Fig.~\ref{fig:rationality_all_models} shows the rationality performance for all models we tested, including their emotional variants, with medium steering strengths.

\clearpage
\section{Decision Task Implementations and Results}
\label{app:task_implementation_and_anchors}

\subsection{Risk Seeking}
\label{app:risk_lottery}

This block measures the risky-choice rate across gain lotteries with explicitly stated probabilities. The block implements the canonical anger/fear dissociation with known $p$ (anger $\downarrow$ perceived risk, $\uparrow$ risk taking; fear/anxiety $\uparrow$ perceived risk, $\downarrow$ risk taking) anchored to \citet{lerner2001fear} and meta-analytic syntheses \cite{wake2020influence, bartholomeyczik2022influence, marini2023}.

To port the benchmark, we form sure amounts $S \in \{10,20,50,100\}$ and explicit success probabilities
$p \in \{0.30, 0.35, 0.40, 0.45, 0.55, 0.60, 0.65, 0.70\}$ (i.e., avoiding $0.50$ to limit focal-point responding). For each $(S,p)$ we target a range of EVs for the risky option by setting the risky EV to $S \cdot (1+\delta)$ with
$\delta \in \{-0.15,-0.125,-0.10,-0.075,-0.05,+0.05,+0.075,+0.10,+0.125,+0.15\}$. The corresponding gain payoff $G$ is rounded to the nearest integer step. Each base tuple $(p,S,G)$ defines a pair key used to query the model and record responses.
Choice prompts are randomly drawn from three templates; options are displayed in randomized order on each trial. As an example, a sample prompt we used is:

\begin{promptbox}
(1) Choose exactly ONE option by repeating the option text verbatim.

1) Gamble: p\% chance to receive \$\{G\}; otherwise \$0 \; 

2) Receive \$\{S\} for certain.
\end{promptbox}

Aggregates are computed as:
\begin{align*}
\text{Risky-choice rate} \; &:= \; \frac{1}{N_{\text{choice}}}\sum_{i=1}^{N_{\text{choice}}} \mathbf{1}\{\text{Choice}_i=\texttt{GAMBLE}\},
\end{align*}
where $\mathbf{1}\{\text{Choice}_i=\texttt{GAMBLE}\}$ is a binary indicator identifying whether the model selected the gamble option.

\begin{figure}[!h]
\centering
\begin{subfigure}[t]{0.49\linewidth}
  \centering
  \includegraphics[width=\linewidth]{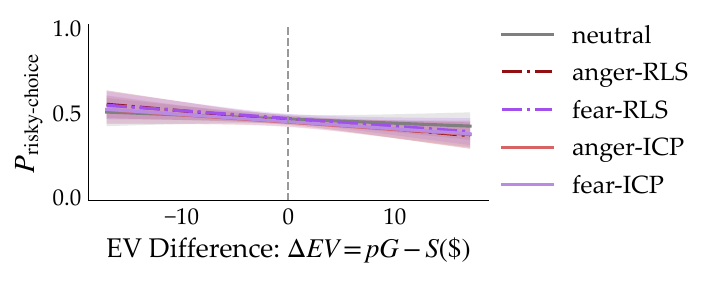}
\caption{\texttt{Qwen3-4B-non-thinking mode}}
\end{subfigure}\hfill
\begin{subfigure}[t]{0.49\linewidth}
  \centering
  \includegraphics[width=\linewidth]{figs/risk_curve_logit_DeltaV_Qwen3_4B_thinking.pdf}
    \caption{\texttt{Qwen3-4B}}
\end{subfigure}
\caption{Risk preference curves: fitted choice probability versus EV difference, comparing neutral and emotion steering (ICP/RLS) in Qwen3-4B}
\label{fig:risk-Qwen3-4B}
\end{figure}

\begin{figure}[!h]
\centering
\begin{subfigure}[t]{0.49\linewidth}
  \centering
  \includegraphics[width=\linewidth]{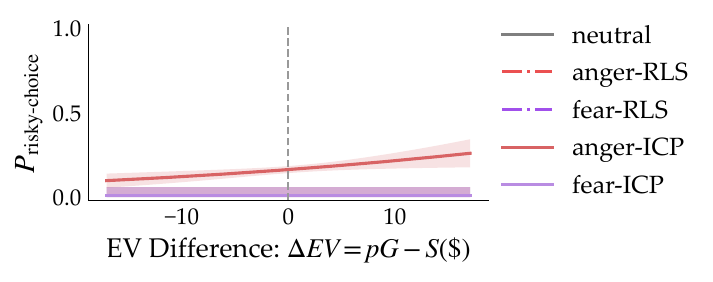}
\caption{\texttt{Qwen3-4B-non-thinking mode}}
\end{subfigure}\hfill
\begin{subfigure}[t]{0.49\linewidth}
  \centering
  \includegraphics[width=\linewidth]{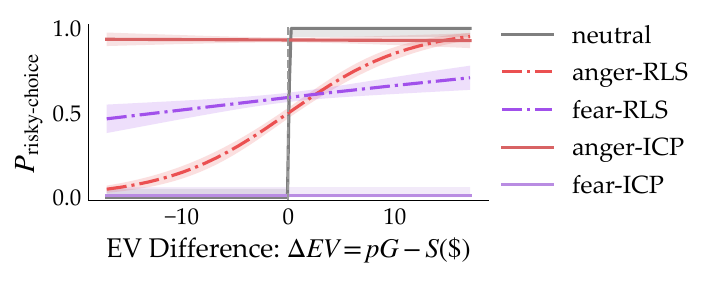}
    \caption{\texttt{Qwen3-8B}}
\end{subfigure}
\caption{Risk preference curves: fitted choice probability versus EV difference, comparing neutral and emotion steering (ICP/RLS) in Qwen3-8B}
\label{fig:risk-Qwen3-8B}
\end{figure}

\begin{figure}[!h]
\centering
\begin{subfigure}[t]{0.49\linewidth}
  \centering
  \includegraphics[width=\linewidth]{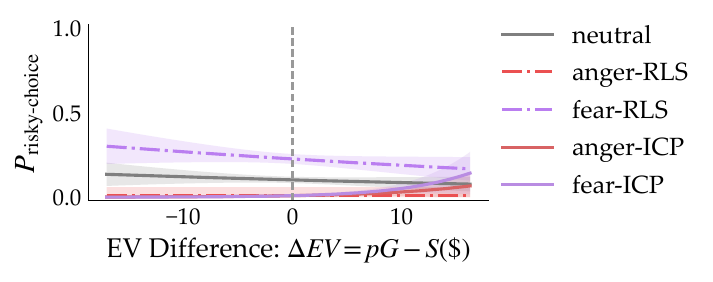}
\caption{\texttt{Llama3.1-8B}}
\end{subfigure}\hfill
\begin{subfigure}[t]{0.49\linewidth}
  \centering
  \includegraphics[width=\linewidth]{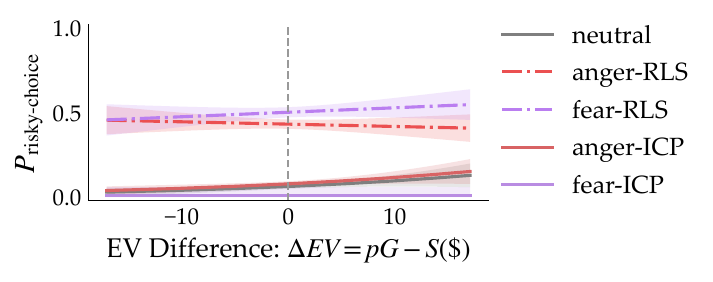}
    \caption{\texttt{Llama3.1-70B}}
\end{subfigure}
\caption{Risk preference curves: fitted choice probability versus EV difference, comparing neutral and emotion steering (ICP/RLS) in Llama3.1}
\label{fig:risk-Llama3.1}
\end{figure}

\begin{figure}[!h]
\centering
\begin{subfigure}[t]{0.49\linewidth}
  \centering
  \includegraphics[width=\linewidth]{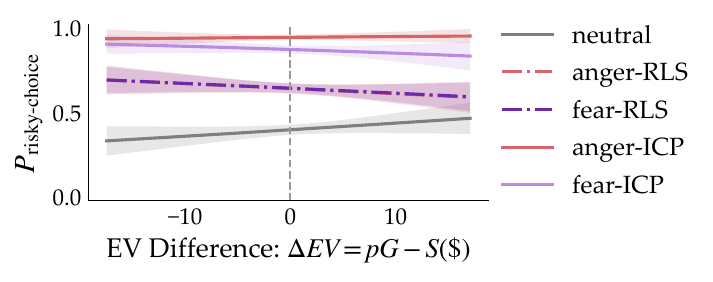}
\caption{\texttt{Olmo3-7B-Instruct}}
\end{subfigure}\hfill
\begin{subfigure}[t]{0.49\linewidth}
  \centering
  \includegraphics[width=\linewidth]{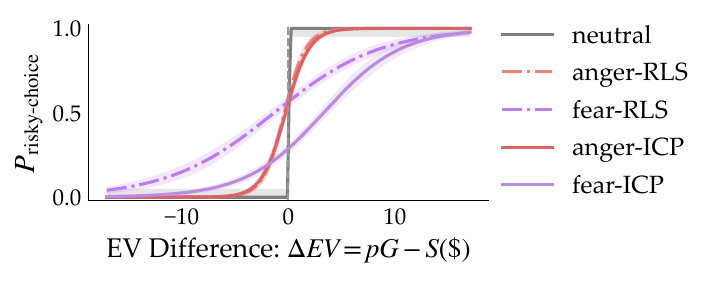}
    \caption{\texttt{Olmo3-7B-Think}}
\end{subfigure}
\caption{Risk preference curves: fitted choice probability versus EV difference, comparing neutral and emotion steering (ICP/RLS) in Olmo3-7B}
\label{fig:risk-Olmo3-7B}
\end{figure}

Figs.~\ref{fig:risk-Qwen3-4B}, \ref{fig:risk-Qwen3-8B}, \ref{fig:risk-Llama3.1}, and \ref{fig:risk-Olmo3-7B} report fitted risky-choice curves as a function of EV difference $\Delta EV=pG-S$ across model families and steering conditions. We evaluate reasoning models with thinking enabled and, where available, with thinking disabled (e.g., \texttt{Qwen3}), and compare \texttt{Olmo3} thinking to its instruction-tuned counterpart to isolate the role of deliberation. Across non-thinking models, choice probabilities are often weakly sensitive to $\Delta EV$ (frequently appearing nearly flat), even when the \emph{direction} of the emotion effect is nominally correct. Enabling thinking yields markedly sharper, monotonic EV sensitivity, producing the expected S-shaped transition around $\Delta EV=0$. Emotion manipulation can nevertheless disrupt this structure: under ICP, especially anger, curves may shift strongly and become less EV-sensitive, whereas RLS tends to preserve monotonic dependence on $\Delta EV$ while producing more graded shifts.

\clearpage

\begin{figure}[!h]
\centering
\begin{subfigure}[t]{0.49\linewidth}
  \centering
  \includegraphics[width=\linewidth]{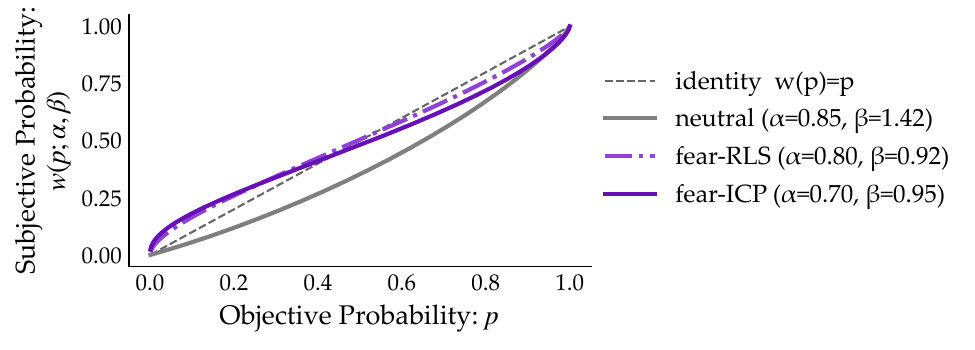}
\caption{\texttt{Qwen3-4B-non-thinking mode}}
\end{subfigure}\hfill
\begin{subfigure}[t]{0.49\linewidth}
  \centering
  \includegraphics[width=\linewidth]{figs/Qwen3_4B_thinking__prelec_weighting_STRATA.pdf}
    \caption{\texttt{Qwen3-4B}}
\end{subfigure}
\caption{Fitted Prelec probability-weighting functions, comparing neutral and fear steering in \texttt{Qwen3-4B}.}
\label{fig:prelec-Qwen3-4B}
\end{figure}

\begin{figure}[!h]
\centering
\begin{subfigure}[t]{0.49\linewidth}
  \centering
  \includegraphics[width=\linewidth]{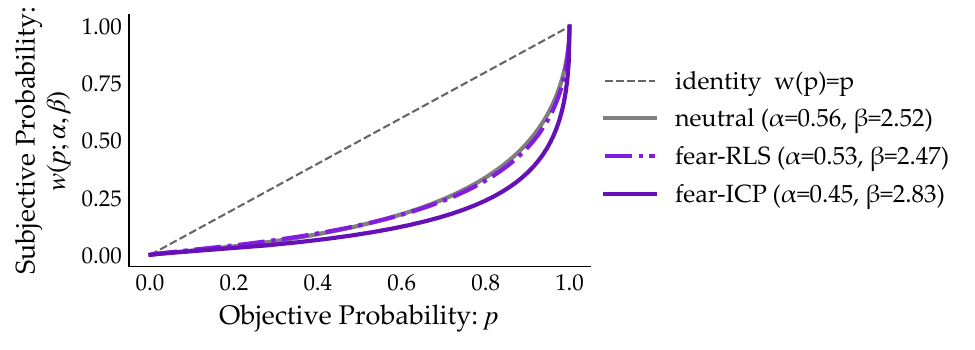}
\caption{\texttt{Qwen3-8B-non-thinking mode}}
\end{subfigure}\hfill
\begin{subfigure}[t]{0.49\linewidth}
  \centering
  \includegraphics[width=\linewidth]{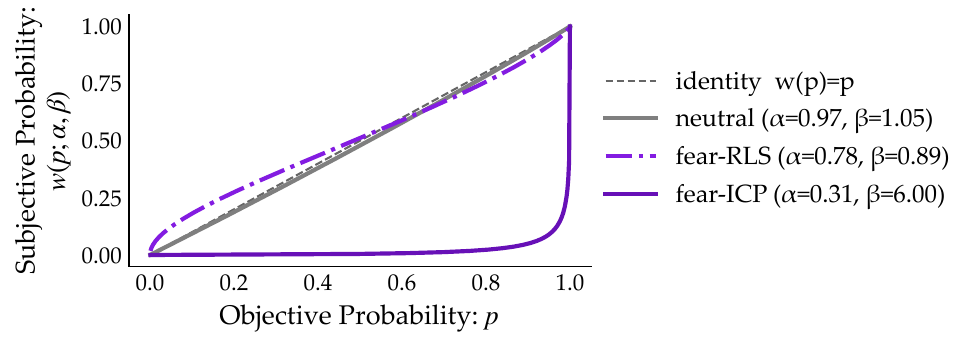}
    \caption{\texttt{Qwen3-8B}}
\end{subfigure}
\caption{Fitted Prelec probability-weighting functions, comparing neutral and fear steering in \texttt{Qwen3-8B}.}
\label{fig:prelec-Qwen3-8B}
\end{figure}

\begin{figure}[!h]
\centering
\begin{subfigure}[t]{0.49\linewidth}
  \centering
  \includegraphics[width=\linewidth]{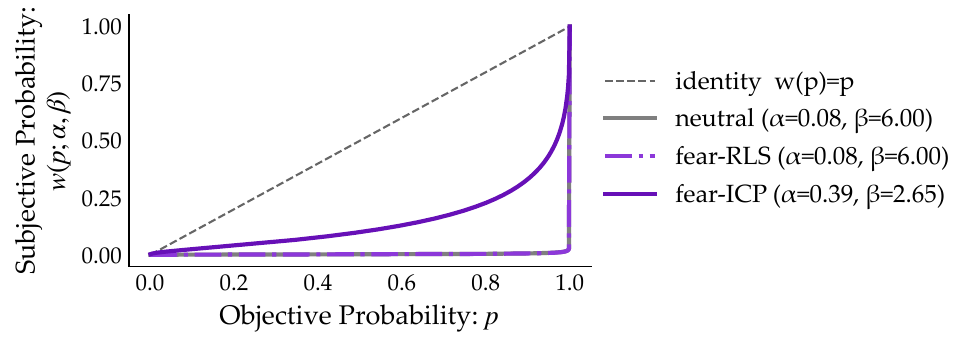}
\caption{\texttt{Llama3.1-8B}}
\end{subfigure}\hfill
\begin{subfigure}[t]{0.49\linewidth}
  \centering
  \includegraphics[width=\linewidth]{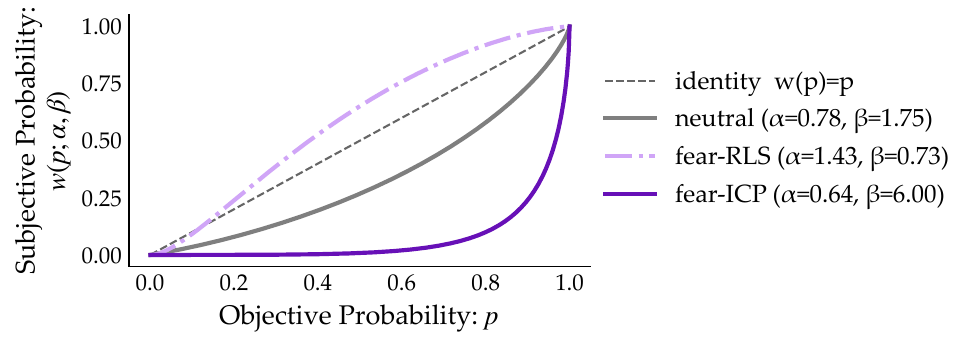}
    \caption{\texttt{Llama3.1-70B}}
\end{subfigure}
\caption{Fitted Prelec probability-weighting functions, comparing neutral and fear steering in \texttt{Llama3.1}.}
\label{fig:prelec-Llama3.1}
\end{figure}

\begin{figure}[!h]
\centering
\begin{subfigure}[t]{0.49\linewidth}
  \centering
  \includegraphics[width=\linewidth]{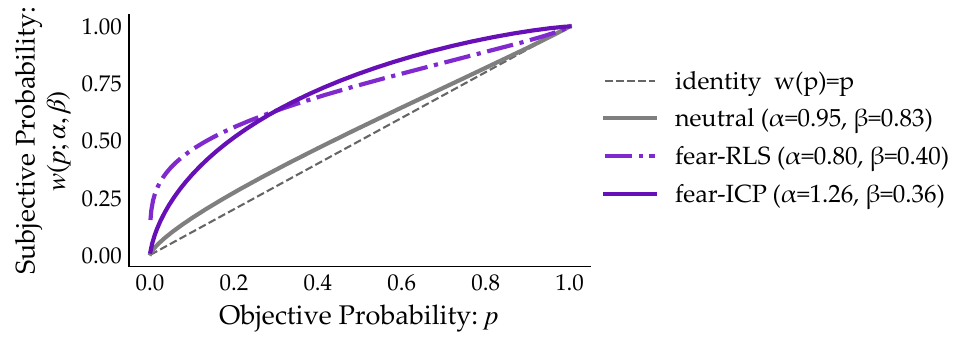}
\caption{\texttt{Olmo3-7B-Instruct}}
\end{subfigure}\hfill
\begin{subfigure}[t]{0.49\linewidth}
  \centering
  \includegraphics[width=\linewidth]{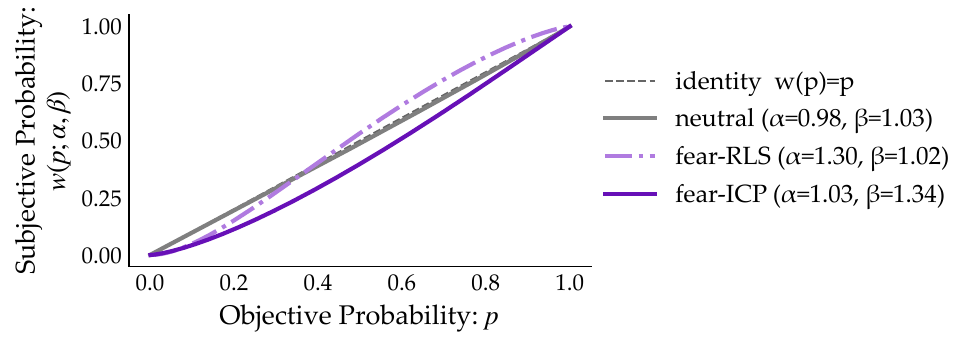}
    \caption{\texttt{Olmo3-7B-Think}}
\end{subfigure}
\caption{Fitted Prelec probability-weighting functions, comparing neutral and fear steering in \texttt{Olmo3-7B}.}
\label{fig:prelec-Olmo3-7B}
\end{figure}

Figs.~\ref{fig:prelec-Qwen3-4B}, \ref{fig:prelec-Qwen3-8B}, \ref{fig:prelec-Llama3.1}, and \ref{fig:prelec-Olmo3-7B} show fitted Prelec probability-weighting functions $w(p)$ comparing neutral and fear steering under both ICP and RLS. ICP often yields extreme curvature; in the fear condition, the fitted mapping can become strongly convex, assigning low subjective weight even to relatively large objective $p$. Thinking models under RLS-steered fear show more graded distortions, but the curvature varies across models.


\begin{figure}[!h]
\centering
\begin{subfigure}[t]{0.49\linewidth}
  \centering
  \includegraphics[width=\linewidth]{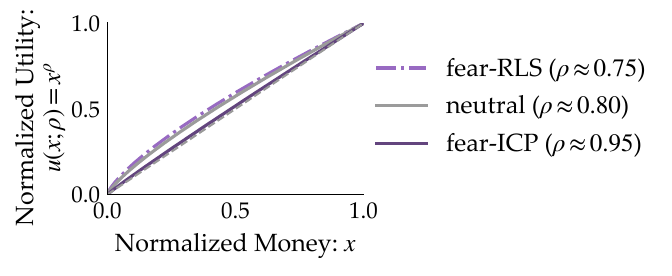}
\caption{\texttt{Qwen3-4B-non-thinking mode}}
\end{subfigure}\hfill
\begin{subfigure}[t]{0.49\linewidth}
  \centering
  \includegraphics[width=\linewidth]{figs/utility_curves_normalized_Qwen3_4B_thinking.pdf}
    \caption{\texttt{Qwen3-4B}}
\end{subfigure}
\caption{Estimated normalized utility curvature across neutral and emotion-steered conditions in \texttt{Qwen3-4B}.}
\label{fig:utility-Qwen3-4B}
\end{figure}

\begin{figure}[!h]
\centering
\begin{subfigure}[t]{0.49\linewidth}
  \centering
  \includegraphics[width=\linewidth]{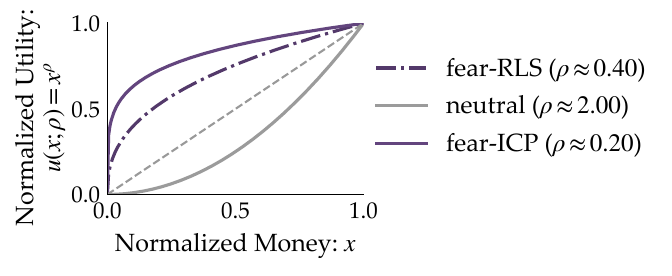}
\caption{\texttt{Qwen3-8B-non-thinking mode}}
\end{subfigure}\hfill
\begin{subfigure}[t]{0.49\linewidth}
  \centering
  \includegraphics[width=\linewidth]{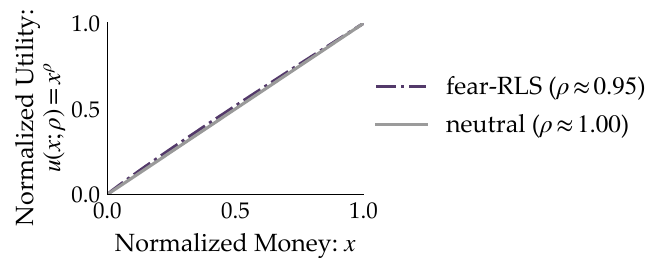}
    \caption{\texttt{Qwen3-8B}}
\end{subfigure}
\caption{Estimated normalized utility curvature across neutral and emotion-steered conditions in \texttt{Qwen3-8B}.}
\label{fig:utility-Qwen3-8B}
\end{figure}

\begin{figure}[!h]
\centering
\begin{subfigure}[t]{0.49\linewidth}
  \centering
  \includegraphics[width=\linewidth]{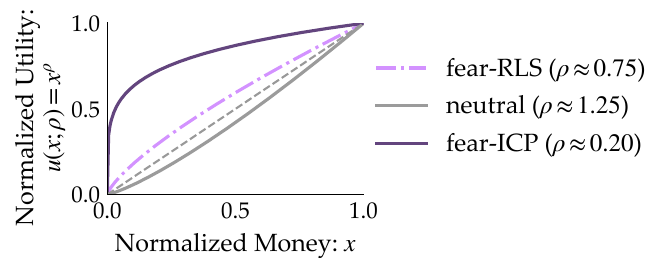}
\caption{\texttt{Llama3.1-8B}}
\end{subfigure}\hfill
\begin{subfigure}[t]{0.49\linewidth}
  \centering
  \includegraphics[width=\linewidth]{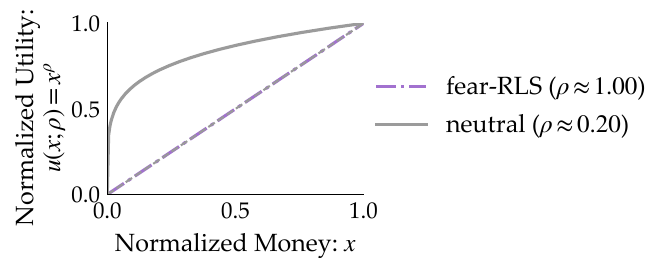}
    \caption{\texttt{Llama3.1-70B}}
\end{subfigure}
\caption{Estimated normalized utility curvature across neutral and emotion-steered conditions in \texttt{Llama3.1}.}
\label{fig:utility-Llama3.1}
\end{figure}

\begin{figure}[!h]
\centering
\begin{subfigure}[t]{0.49\linewidth}
  \centering
  \includegraphics[width=\linewidth]{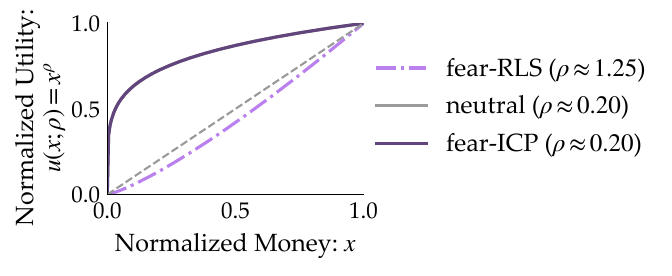}
\caption{\texttt{Olmo3-7B-Instruct}}
\end{subfigure}\hfill
\begin{subfigure}[t]{0.49\linewidth}
  \centering
  \includegraphics[width=\linewidth]{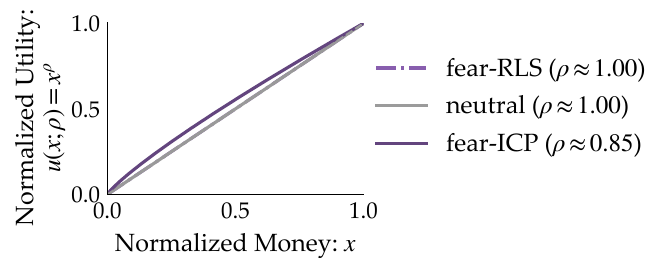}
    \caption{\texttt{Olmo3-7B-Think}}
\end{subfigure}
\caption{Estimated normalized utility curvature across neutral and emotion-steered conditions in \texttt{Olmo3-7B}.}
\label{fig:utility-Olmo3}
\end{figure}

\clearpage
Figs.~\ref{fig:utility-Qwen3-4B}, \ref{fig:utility-Qwen3-8B}, \ref{fig:utility-Llama3.1}, and \ref{fig:utility-Olmo3} summarize estimated normalized utility curvature across neutral and emotion-steered conditions. Under neutral thinking, fitted $\rho$ is typically close to $1$, consistent with near-linear valuation of gains. Under fear, ICP frequently induces extreme risk aversion; in some settings (e.g., \texttt{Qwen3-8B}), models almost never gamble, preventing stable curve fitting. RLS produces more graded shifts in $\rho$ away from baseline. We also observe model-specific exceptions (e.g., in \texttt{Olmo3-7B} thinking, ICP can yield cleaner separations than RLS for this fit), reinforcing that steering mechanisms interact with model families in non-uniform ways.

\clearpage
\subsection{Ambiguity Aversion}
\label{app:ambiguity_block_known_unknown}

This block measures ambiguity aversion by contrasting choices between a known 
50/50 gamble and an unknown-probability urn. Human studies show that sadness can increase ambiguity neutrality \cite{cohn2015,baillon2016sadder}.
In our implementation, each trial presents:
\begin{itemize}
    \item \textbf{Known option:} an urn with 50/50 mixed red/blue balls, where red balls pay off and blue balls lead to no return. In other words, the agent faces a 50\% chance to receive \$\(G\) (otherwise \$0).
    \item \textbf{Unknown option:} an urn with the same payoff \$\(G\) for red balls, but with 
    unspecified probability of success.
\end{itemize}
Payoff values \(G\) are varied over a fixed grid to allow tracing 
$P_{\text{Choose-Known}}$ as a function of stakes. For prompt templates, options appear in random order, and a template is sampled in each trial, such as:

\begin{promptbox}
Choose exactly ONE option:

1) Known urn (50/50): 25 RED + 25 BLACK. Win \$30 if RED is drawn; otherwise \$0.

2) Unknown urn: 50 balls; red/black ratio unspecified. Win \$30 if RED is drawn; otherwise \$0.

Answer with the option text only.
\end{promptbox}

Ambiguity aversion is captured by the ratio at which the model chooses the known option over the unknown one:
\[
\widehat{\text{AAI}}
=
P_\text{choose Known},
\]
computed cross conditions with 95\% Clopper--Pearson 
intervals.  
Emotion contrasts are defined as
\[
\Delta_{\text{emotion}}
=
\widehat{\text{AAI}}_{\text{emotion}}
-
\widehat{\text{AAI}}_{\text{neutral}},
\]
with pooled two-proportion $z$ tests reported for descriptive comparison.

\begin{figure}[!h]
\centering
\begin{subfigure}[t]{0.49\linewidth}
  \centering
  \includegraphics[width=\linewidth]{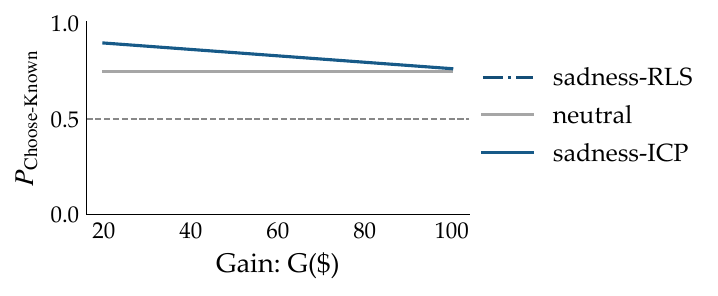}
\caption{\texttt{Qwen3-4B-non-thinking mode}}
\end{subfigure}\hfill
\begin{subfigure}[t]{0.49\linewidth}
  \centering
  \includegraphics[width=\linewidth]{figs/ambiguity_payoff_FITLINE_Qwen3_4B_thinking.pdf}
    \caption{\texttt{Qwen3-4B}}
\end{subfigure}
\caption{Ambiguity aversion by stake size: linear fits of $P_\text{Choose Known}$ as a function of gain $G$, comparing neutral and sadness steering (ICP/RLS) in  \texttt{Qwen3-4B}.}
\label{fig:ambig-Qwen3-4B}
\end{figure}

\begin{figure}[!h]
\centering
\begin{subfigure}[t]{0.49\linewidth}
  \centering
  \includegraphics[width=\linewidth]{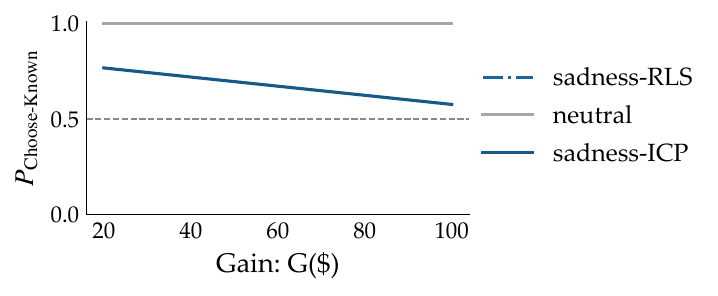}
\caption{\texttt{Qwen3-8B-non-thinking mode}}
\end{subfigure}\hfill
\begin{subfigure}[t]{0.49\linewidth}
  \centering
  \includegraphics[width=\linewidth]{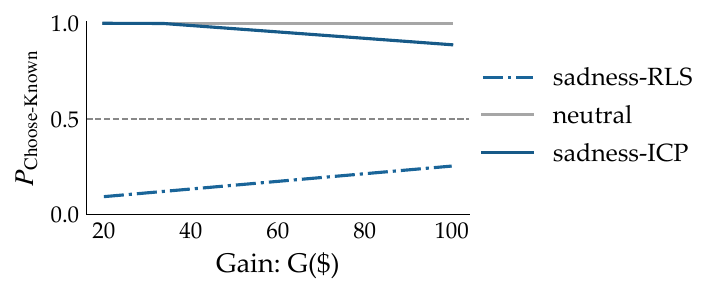}
    \caption{\texttt{Qwen3-8B}}
\end{subfigure}
\caption{Ambiguity aversion by stake size: linear fits of $P_\text{Choose Known}$ as a function of gain $G$, comparing neutral and sadness steering (ICP/RLS) in  \texttt{Qwen3-8B}.}
\label{fig:ambig-Qwen3-8B}
\end{figure}

\begin{figure}[!h]
\centering
\begin{subfigure}[t]{0.49\linewidth}
  \centering
  \includegraphics[width=\linewidth]{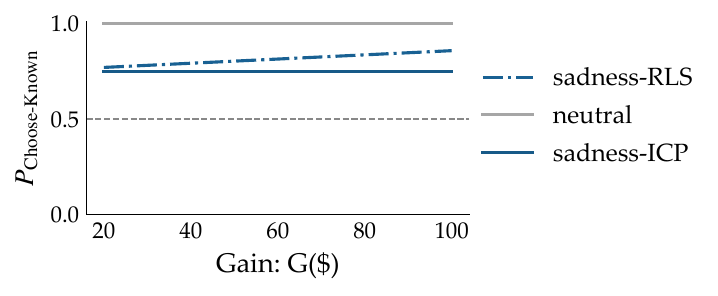}
\caption{\texttt{Llama3.1-8B}}
\end{subfigure}\hfill
\begin{subfigure}[t]{0.49\linewidth}
  \centering
  \includegraphics[width=\linewidth]{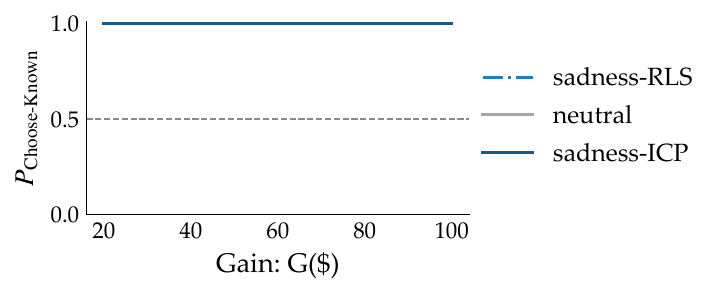}
    \caption{\texttt{Llama3.1-70B}}
\end{subfigure}
\caption{Ambiguity aversion by stake size: linear fits of $P_\text{Choose Known}$ as a function of gain $G$, comparing neutral and sadness steering (ICP/RLS) in  \texttt{Llama3.1}.}
\label{fig:ambig-Llama3.1}
\end{figure}

\begin{figure}[!h]
\centering
\begin{subfigure}[t]{0.49\linewidth}
  \centering
  \includegraphics[width=\linewidth]{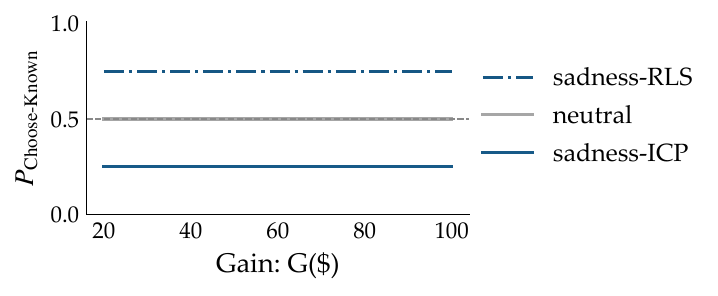}
\caption{\texttt{Olmo3-7B-Instruct}}
\end{subfigure}\hfill
\begin{subfigure}[t]{0.49\linewidth}
  \centering
  \includegraphics[width=\linewidth]{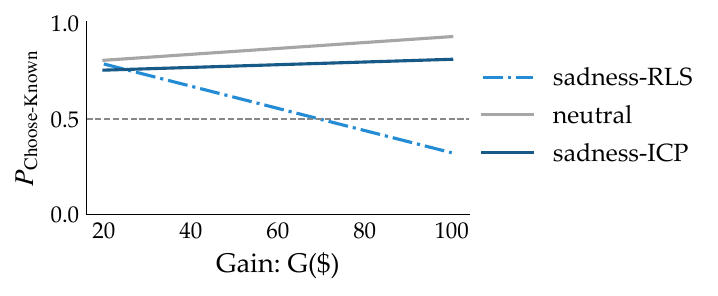}
    \caption{\texttt{Olmo3-7B-Think}}
\end{subfigure}
\caption{Ambiguity aversion by stake size: linear fits of $P_\text{Choose Known}$ as a function of gain $G$, comparing neutral and sadness steering (ICP/RLS) in  \texttt{Olmo3-7B}.}
\label{fig:ambig-Olmo3}
\end{figure}

Figs.~\ref{fig:ambig-Qwen3-4B}, \ref{fig:ambig-Qwen3-8B}, \ref{fig:ambig-Llama3.1}, and \ref{fig:ambig-Olmo3} plot ambiguity aversion by stake size, fitting $P_\text{Choose Known}$ as a function of gain $G$ and comparing neutral to sadness steering (ICP/RLS). Across most models, baseline behavior exhibits near-ceiling ambiguity aversion, far stronger than typical human levels. An apparent exception is \texttt{Olmo3-7B} instruct, which initially appears ambiguity-neutral; however, closer inspection shows that it largely follows the randomized presentation order of the known/unknown options, indicating a failure to respond to the task. Under sadness, RLS reduces ambiguity aversion in \texttt{Qwen3-4B} and \texttt{Olmo3-7B} thinking, shifting choices toward ambiguity neutrality. By contrast, \texttt{Qwen3-8B} shows erratic behavior with substantial variability across thinking on/off settings.

\clearpage
\subsection{Loss Aversion}
\label{app:loss_block_lapw}

This block measures loss aversion in mixed 50/50 gain--loss gambles. 
Human work using similar symmetric gain/loss matrices finds that incidental fear increases 
the loss-aversion parameter $\lambda$ in Prospect Theory-style models 
\cite{schulreich2016incidental}. In this block, each trial offers a 50/50 gamble with a possible gain $G$ and loss $L$:
\[
\text{Gamble: } 50\% \text{ chance to win } G \text{ and } 50\% \text{ chance to lose } L.
\]
We form a symmetric grid with
\[
G \in \{\$5,\$6,\dots,\$14\}, 
\qquad
L \in \{\$5,\$6,\dots,\$14\},
\]

Each $(G,L)$ pair is instantiated under several equivalent instruction templates to vary surface wording while holding the structure fixed. The options are order-randomized and embedded in one of several instruction templates, such as below:

\begin{promptbox}

Pick one option:

    1) ACCEPT the coin-flip gamble: 50\% chance to win \{G\} and 50\% chance to lose \{L\}.
    
    2) REJECT; take \$0 for certain.
\end{promptbox}

The primary outcome is the accept rate for the mixed gambles,
\[
P_\text{ACCEPT} 
\,=\, \frac{1}{N}\sum_{i=1}^{N}\mathbf{1}\{\text{Accept gamble on trial } i\},
\]
alongside an estimated loss-aversion parameter $\lambda$ obtained according to Section~\ref{economic}.

\begin{figure}[!h]
\centering
\begin{subfigure}[t]{0.49\linewidth}
  \centering
  \includegraphics[width=\linewidth]{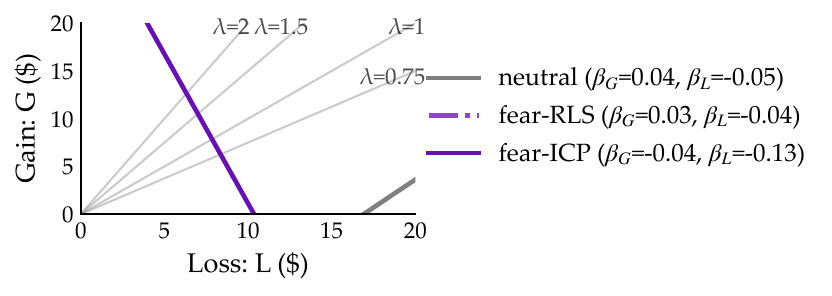}
\caption{\texttt{Qwen3-4B-non-thinking mode}}
\end{subfigure}\hfill
\begin{subfigure}[t]{0.49\linewidth}
  \centering
  \includegraphics[width=\linewidth]{figs/Qwen3_4B_thinking__Iso-utility-Gain-loss_STRATA.pdf}
    \caption{\texttt{Qwen3-4B}}
\end{subfigure}
\caption{Loss-aversion geometry: reference iso-utility lines $G=\lambda L$ with fitted 50\% acceptance boundary under neutral and fear steering (ICP/RLS) in  \texttt{Qwen3-4B}.}
\label{fig:loss-Qwen3-4B}
\end{figure}

\begin{figure}[!h]
\centering
\begin{subfigure}[t]{0.49\linewidth}
  \centering
  \includegraphics[width=\linewidth]{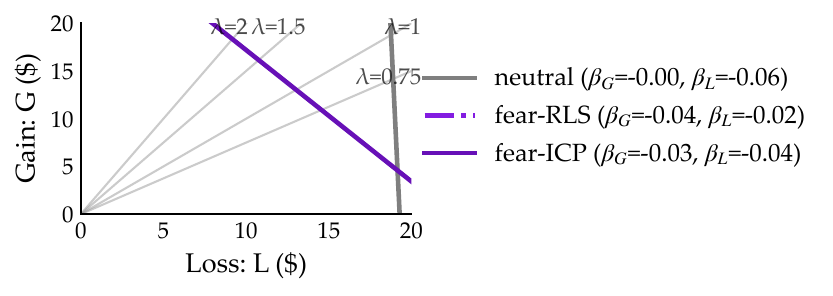}
\caption{\texttt{Qwen3-8B-non-thinking mode}}
\end{subfigure}\hfill
\begin{subfigure}[t]{0.49\linewidth}
  \centering
  \includegraphics[width=\linewidth]{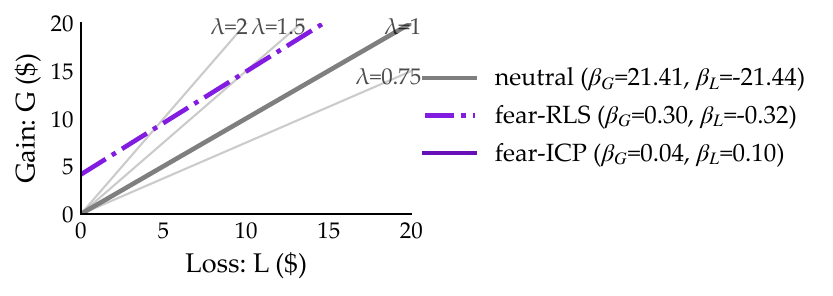}
    \caption{\texttt{Qwen3-8B}}
\end{subfigure}
\caption{Loss-aversion geometry: reference iso-utility lines $G=\lambda L$ with fitted 50\% acceptance boundary under neutral and fear steering (ICP/RLS) in  \texttt{Qwen3-8B}.}
\label{fig:loss-Qwen3-8B}
\end{figure}

\begin{figure}[!h]
\centering
\begin{subfigure}[t]{0.49\linewidth}
  \centering
  \includegraphics[width=\linewidth]{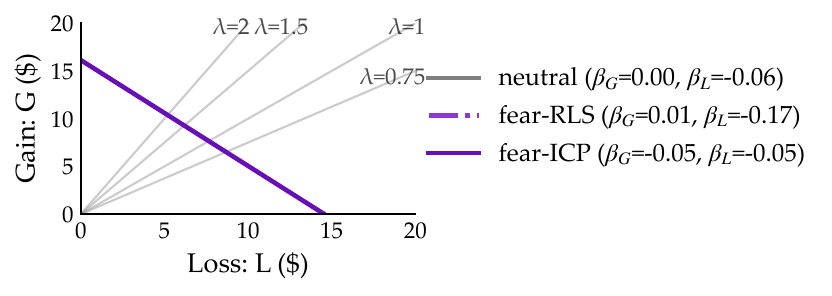}
\caption{\texttt{Llama3.1-8B}}
\end{subfigure}\hfill
\begin{subfigure}[t]{0.49\linewidth}
  \centering
  \includegraphics[width=\linewidth]{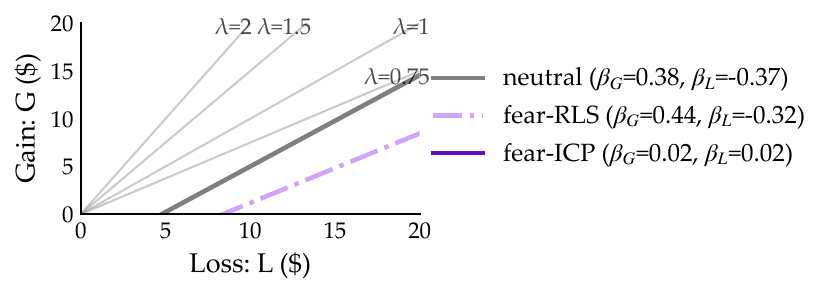}
    \caption{\texttt{Llama3.1-70B}}
\end{subfigure}
\caption{Loss-aversion geometry: reference iso-utility lines $G=\lambda L$ with fitted 50\% acceptance boundary under neutral and fear steering (ICP/RLS) in  \texttt{Llama3.1}.}
\label{fig:loss-Llama3.1}
\end{figure}

\begin{figure}[!h]
\centering
\begin{subfigure}[t]{0.49\linewidth}
  \centering
  \includegraphics[width=\linewidth]{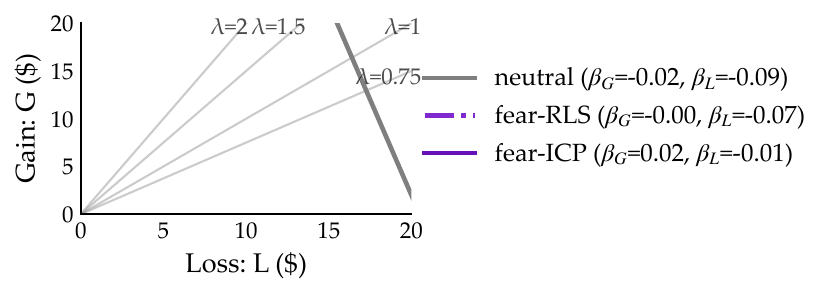}
\caption{\texttt{Olmo3-7B-Instruct}}
\end{subfigure}\hfill
\begin{subfigure}[t]{0.49\linewidth}
  \centering
  \includegraphics[width=\linewidth]{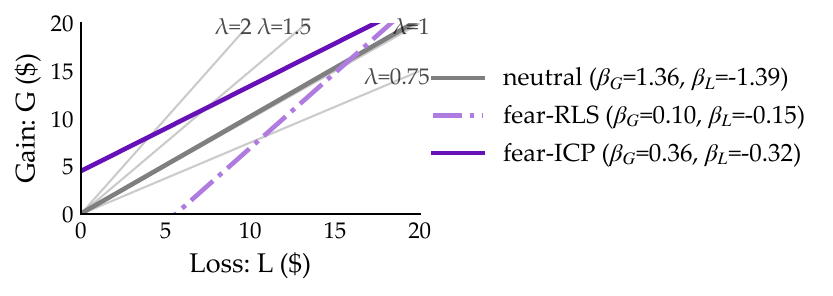}
    \caption{\texttt{Olmo3-7B-Think}}
\end{subfigure}
\caption{Loss-aversion geometry: reference iso-utility lines $G=\lambda L$ with fitted 50\% acceptance boundary under neutral and fear steering (ICP/RLS) in  \texttt{Olmo3-7B}.}
\label{fig:loss-Olmo3}
\end{figure}

Figs.~\ref{fig:loss-Qwen3-4B}, \ref{fig:loss-Qwen3-8B}, \ref{fig:loss-Llama3.1}, and \ref{fig:loss-Olmo3} extend the loss-aversion geometry to additional model families, plotting reference iso-utility lines $G=\lambda L$ alongside the fitted $50\%$ acceptance boundary under neutral and fear steering (ICP/RLS). Across several non-thinking baselines, the fitted coefficients are unstable (e.g., $\beta_L>0$, implying that larger losses \emph{increase} acceptance), indicating failures to represent the task structure rather than meaningful loss preferences. Among thinking models, \texttt{Qwen3} variants show little evidence of classical loss aversion under neutral conditions ($\lambda\approx 1$). Under fear, RLS more consistently shifts the boundary toward higher implied loss sensitivity (via increased slope and/or a positive intercept), whereas ICP frequently drives near-universal rejection, pushing the boundary outside the plotted range and requiring the proxy visualization described above. We also observe model-specific differences in how fear manifests: for \texttt{Olmo3-7B} thinking, ICP primarily increases the intercept (a larger ``entry fee'' gain), while RLS yields a different trade-off between intercept and slope, underscoring that steering mechanisms interact with model families in non-uniform ways.

\clearpage
\subsection{Endowment Effect}
\label{app:endowment_effect}

This block elicits selling prices (willingness-to-accept, WTA) and buying prices 
(willingness-to-pay, WTP) for the same neutral consumer goods. The goal is to 
quantify the endowment effect as a systematic WTA--WTP gap
under affect-neutral wording. In our implementation, the item set consists of small lab-friendly objects (e.g., ceramic mug, 
spiral notebook, ballpoint pen, USB drive, travel-size umbrella, raffle ticket, 
insulated lunch bag), each specified by a short name and a brief, affect-neutral 
description (e.g., ``11 oz plain white ceramic coffee mug, new, no logo.''). Each condition is instantiated under two surface 
instruction templates, a buy framing and a sell framing:
\begin{promptbox}
State a single dollar price.

Item: Ceramic mug

Description: 11 oz plain white ceramic coffee mug, new, no logo.

You have just been given this brand new item. What is the MINIMUM price you would ACCEPT to sell it?

Answer with one line exactly: Price: \$NN
\end{promptbox}

\begin{promptbox}
State a single dollar price.

Item: Ceramic mug

Description: 11 oz plain white ceramic coffee mug, new, no logo.

You do NOT own this brand new item. What is the MAXIMUM price you would PAY to buy it?

Answer with one line exactly: Price: \$NN
\end{promptbox}

From the recorded prices, we compute the mean sell and buy valuations across all items and prompts:
\[
\text{WTA} = \frac{1}{N_{\text{sell}}}\sum_{i \in \text{sell}} P^{\text{clip}}_i,
\qquad
\text{WTP} = \frac{1}{N_{\text{buy}}}\sum_{i \in \text{buy}} P^{\text{clip}}_i.
\]
The endowment effect gap is summarized as
\[
\Delta_E = \text{WTA} - \text{WTP}.
\]

\begin{figure}[!h]
\centering
\begin{subfigure}[t]{0.49\linewidth}
  \centering
  \includegraphics[width=\linewidth]{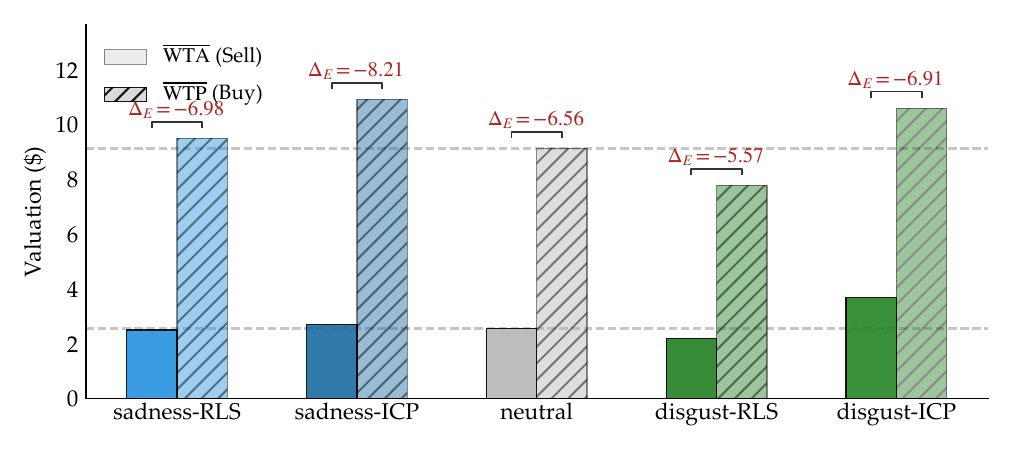}
\caption{\texttt{Qwen3-4B-non-thinking mode}}
\end{subfigure}\hfill
\begin{subfigure}[t]{0.49\linewidth}
  \centering
  \includegraphics[width=\linewidth]{figs/endowment_WTA_WTP_Qwen3_4B_thinking.pdf}
    \caption{\texttt{Qwen3-4B}}
\end{subfigure}
\caption{Endowment effect summary: $\overline{\text{WTA}}$ and $\overline{\text{WTP}}$ per steering condition, annotated by endowment premium $\Delta_E$ (\texttt{Qwen3-4B})
}
\label{fig:endowment-Qwen3-4B}
\end{figure}

\begin{figure}[!h]
\centering
\begin{subfigure}[t]{0.49\linewidth}
  \centering
  \includegraphics[width=\linewidth]{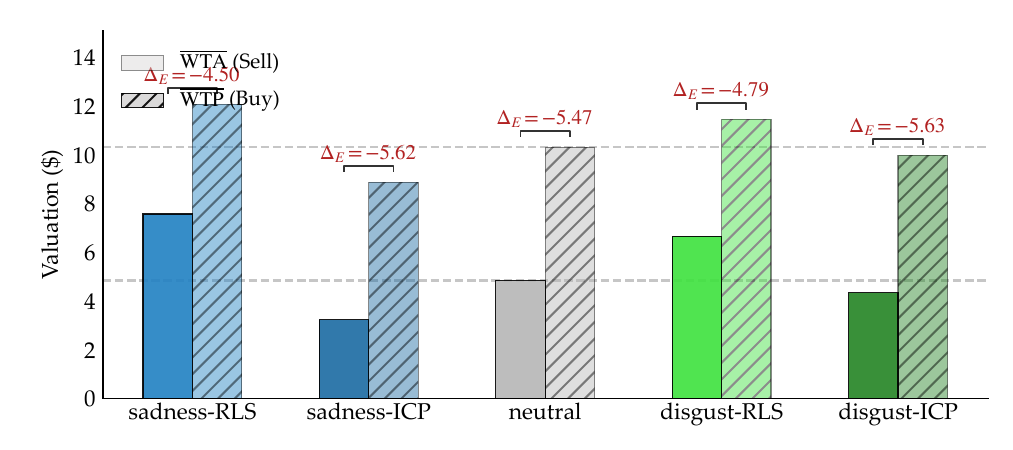}
\caption{\texttt{Qwen3-8B-non-thinking mode}}
\end{subfigure}\hfill
\begin{subfigure}[t]{0.49\linewidth}
  \centering
  \includegraphics[width=\linewidth]{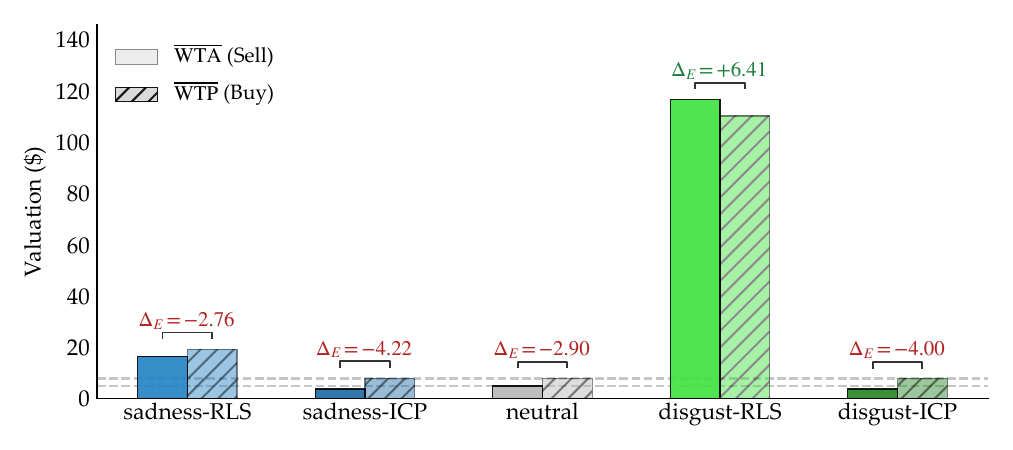}
    \caption{\texttt{Qwen3-8B}}
\end{subfigure}
\caption{Endowment effect summary: $\overline{\text{WTA}}$ and $\overline{\text{WTP}}$ per steering condition, annotated by endowment premium $\Delta_E$ (\texttt{Qwen3-8B})
}
\label{fig:endowment-Qwen3-8B}
\end{figure}

\begin{figure}[!h]
\centering
\begin{subfigure}[t]{0.49\linewidth}
  \centering
  \includegraphics[width=\linewidth]{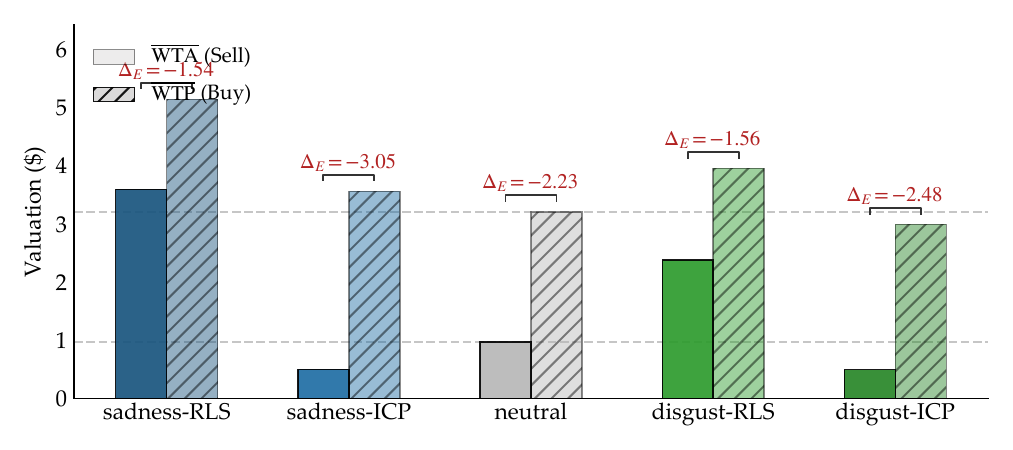}
\caption{\texttt{Llama3.1-8B}}
\end{subfigure}\hfill
\begin{subfigure}[t]{0.49\linewidth}
  \centering
  \includegraphics[width=\linewidth]{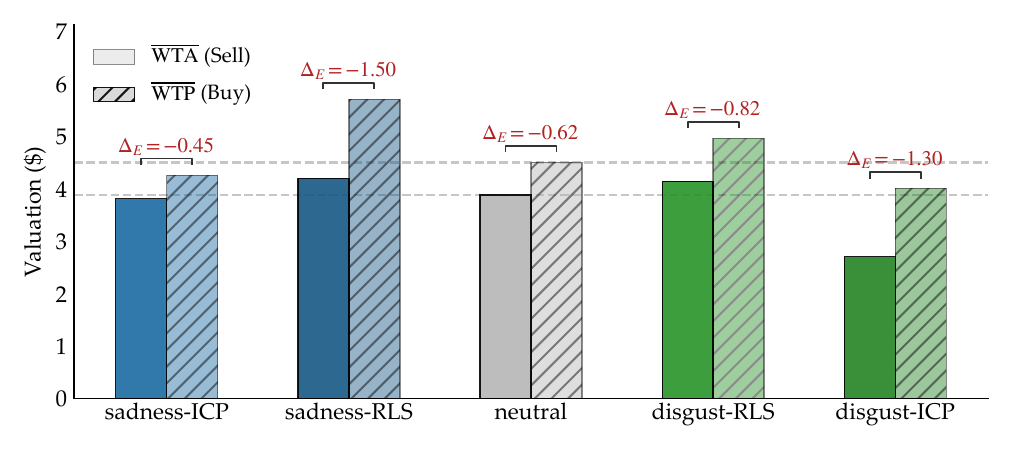}
    \caption{\texttt{Llama3.1-70B}}
\end{subfigure}
\caption{Endowment effect summary: $\overline{\text{WTA}}$ and $\overline{\text{WTP}}$ per steering condition, annotated by endowment premium $\Delta_E$ (\texttt{Llama3.1})
}
\label{fig:endowment-Llama3.1}
\end{figure}

\begin{figure}[!h]
\centering
\begin{subfigure}[t]{0.49\linewidth}
  \centering
  \includegraphics[width=\linewidth]{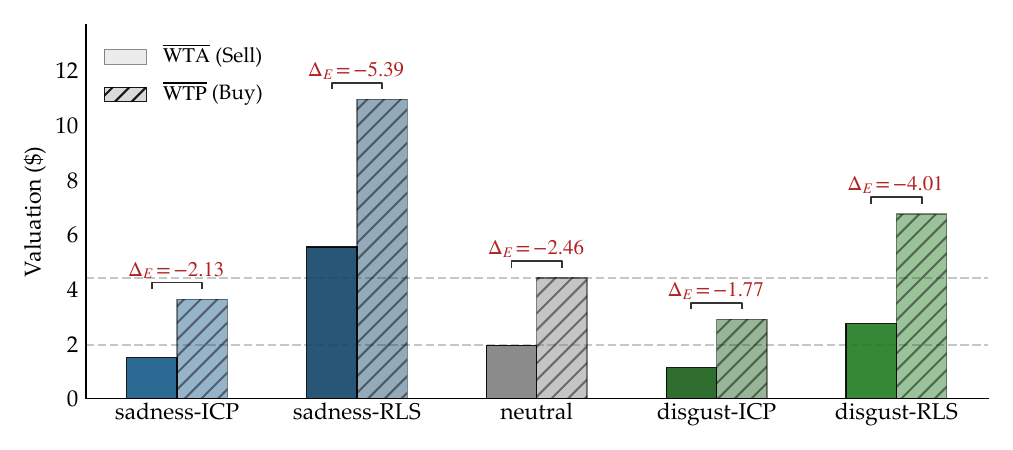}
\caption{\texttt{Olmo3-7B-Instruct}}
\end{subfigure}\hfill
\begin{subfigure}[t]{0.49\linewidth}
  \centering
  \includegraphics[width=\linewidth]{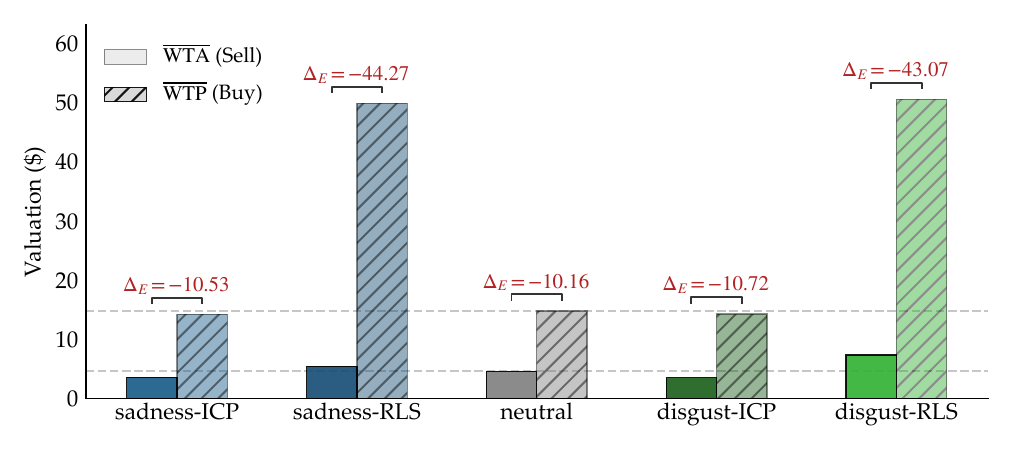}
    \caption{\texttt{Olmo3-7B-Think}}
\end{subfigure}
\caption{Endowment effect summary: $\overline{\text{WTA}}$ and $\overline{\text{WTP}}$ per steering condition, annotated by endowment premium $\Delta_E$ (\texttt{Olmo3-7B})
}
\label{fig:endowment-Olmo3_7B}
\end{figure}

Figs.~\ref{fig:endowment-Qwen3-4B}, \ref{fig:endowment-Qwen3-8B}, \ref{fig:endowment-Llama3.1}, and \ref{fig:endowment-Olmo3_7B} report $\overline{\text{WTA}}$ and $\overline{\text{WTP}}$ by steering condition, annotated by the endowment premium $\Delta_E$. Notably, the neutral baseline exhibits a \emph{negative} gap across all tested models ($\Delta_E<0$), i.e., selling valuations are lower than buying valuations, opposite to the canonical human endowment effect. Because the baseline fails to reproduce the human anchor, we treat emotion-induced shifts in this domain as secondary and focus instead on diagnosing the source of the reversed sign.

\paragraph{Sell-frame diagnostic for the reversed gap.}
In our main results, \texttt{Qwen3} often exhibits a reversed endowment pattern ($\Delta E<0$), with selling valuations lower than buying valuations. A plausible mechanism is that the model interprets selling as an ``unloading'' context (i.e., ``wanting to get rid of'' an owned item), which could depress WTA. To test this directly, we introduce a sell-only manipulation that varies the selling context while holding the items, response format, and ``minimum acceptable price'' instruction constant. Specifically, we compare:
(i)~an \emph{unload} frame: ``You bought this brand new item, but you now want to get rid of it'';
and (ii)~a \emph{gift-buyer} frame: ``You received this brand new item as an unexpected gift. Someone else says they want to buy it from you.'' We then compare these sell-only valuations to the original sell and buy conditions.

Table~\ref{tab:endowment_conjecture} summarizes the results for \texttt{Qwen3-4B}. Across steering conditions, the gift-buyer frame increases WTA relative to the unload frame, consistent with the intuition that ``getting rid of it'' depresses selling valuations. However, both sell-only WTAs remain below WTP overall, indicating that the reversed gap is not fully explained by the unload interpretation alone. One possible complementary explanation is that instruction-tuned models default to other-regarding or ``fair price'' norms that suppress self-serving selling valuations, yielding lower WTA even when selling is framed as a gift resale.

Taken together, our experiments suggest that recovering robust intrinsic valuation behavior in LLMs, including Large Reasoning Models, requires more targeted interventions. This most plausibly includes post-training focused on preference elicitation and private-value reasoning, or more extensive in-context learning scaffolds that explicitly teach the mechanics of valuation in buying/selling, bargaining, negotiation, and related economic trade settings. Such scaffolds may be necessary for models to consistently separate other-regarding ``fair price'' norms from self-interested valuation when the task calls for the latter.

\begin{table}[!t]
\centering
\small
\begin{tabular}{l l l r r r r}
\toprule
Model & Steer & Concept & Unload WTA & Gift-buyer WTA & Sell WTA & Buy WTP \\
\midrule
\texttt{Qwen3\_4B} & RLS & sadness & 5.23 & 3.97 & 5.21 & 9.95 \\
\texttt{Qwen3\_4B} & RLS & disgust & 3.28 & 3.48 & 4.12 & 10.14 \\
\texttt{Qwen3\_4B} & ICP & sadness & 3.75 & 4.59 & 4.45 & 7.44 \\
\texttt{Qwen3\_4B} & ICP & disgust & 2.53 & 3.97 & 3.18 & 4.82 \\
\texttt{Qwen3\_4B} & None & -- & 3.84 & 4.13 & 5.17 & 7.89 \\
\bottomrule
\end{tabular}
\caption{Sell-frame diagnostic for the reversed endowment gap in \texttt{Qwen3-4B-thinking}. ``Unload'' and ``Gift-buyer'' are sell-only frames that both elicit WTA (minimum acceptable selling price). ``Sell'' and ``Buy'' are the original WTA/WTP conditions used to compute $\Delta E=\text{WTA}-\text{WTP}$.}
\label{tab:endowment_conjecture}
\end{table}

\clearpage
\subsection{Temporal discounting}
\label{app:temporal}
This block measures time preferences over smaller-sooner (SS) versus larger-later (LL) monetary rewards. It combines two variations. The first set defines three magnitude bands:
\[
\text{Small: } A_{L} \approx \$25,\quad
\text{Medium: } A_{L} \approx \$55,\quad
\text{Large: } A_{L} \approx \$85.
\]
For each band, the larger-later amount $A_{L}$ is fixed (25, 55, or 85 dollars), and the smaller-sooner amounts $A_{S}$ range over band-specific grids:
\begin{align*}
\text{Small band: } & A_{S} \in \{11,12,\dots,24\},\\
\text{Medium band: } & A_{S} \in \{25,27,30,32,35,37,40,42,45,47,50,52,53,54\},\\
\text{Large band: } & A_{S} \in \{35,37,40,42,45,50,52,55,60,62,65,70,75,80\}.
\end{align*}
Delays for the larger-later option are drawn from
\[
t_{L} \in \{7,14,30,60,90,120,180\} \text{ days},
\]
and are cycled across the $A_{S}$ grid to produce a spread of SS-LL pairs. The sooner option is always
available ``today'' ($t_{S}=0$). Each $(A_{S},A_{L},t_{L})$ combination is presented twice with a randomized order of SS vs.\ LL, creating a within-set ordering control.

The second set combines a grid of amount pairs and delay pairs to broaden coverage across stakes and time horizons. Amount pairs are
\[
(A_{S},A_{L}) \in \{(6,7), (10,12), (14,17), (20,25), (28,35), (34,45), (40,57)\},
\]
and delay pairs are
\[
(t_{S},t_{L}) \in \{(0,14), (0,28), (0,42), (14,28), (14,42), (28,42)\} \text{ (in days)}.
\]
For each $(A_{S},A_{L},t_{S},t_{L})$ combination, the sooner option is described as 
``\$$A_{S}$ today'' if $t_{S}=0$ and ``\$$A_{S}$ in $t_{S}$ days'' otherwise. The later option is 
described as ``\$$A_{L}$ in $t_{L}$ days''; when $t_{L}\in\{14,28,42\}$, it is formatted as 
“$w$ weeks” with $w = t_{L}/7$. All trials use a single, neutral instruction template that enforces a one-option choice:
\begin{promptbox}
Choose exactly ONE option, weighing the payoff of waiting against the time you must wait.

1) \$25 in 7 days.

2) \$11 today.

Answer with the option text only.
\end{promptbox}

We report the proportion of later choices: $P_\text{LATER}$.

\begin{figure}[!h]
\centering
\begin{subfigure}[t]{0.49\linewidth}
  \centering
  \includegraphics[width=\linewidth]{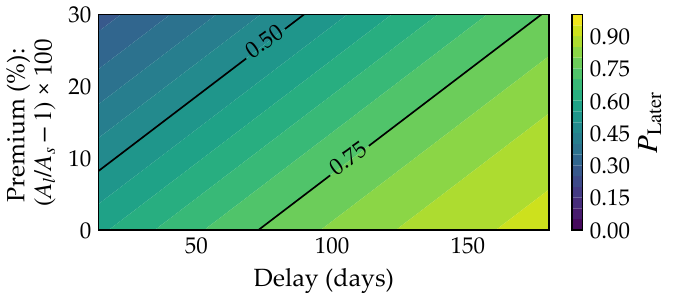}
  \caption{\texttt{Qwen3-4B-non-thinking mode}}
  \label{fig:p1_4B_}
\end{subfigure}\hfill
\begin{subfigure}[t]{0.49\linewidth}
  \centering
  \includegraphics[width=\linewidth]{figs/intertemporal_Qwen3_4B_thinking_neutral_heat.pdf}
  \caption{\texttt{Qwen3-4B}}
  \label{fig:p2_4B_th}
\end{subfigure}
\caption{Temporal choice surface: fitted $P_\text{Later}$ over delay and premium (\%), with iso-probability contours (0.50/0.75).}
\label{fig:temporal-neutral-qwen4b}
\end{figure}

\begin{figure}[!h]
\centering
\begin{subfigure}[t]{0.49\linewidth}
  \centering
  \includegraphics[width=\linewidth]{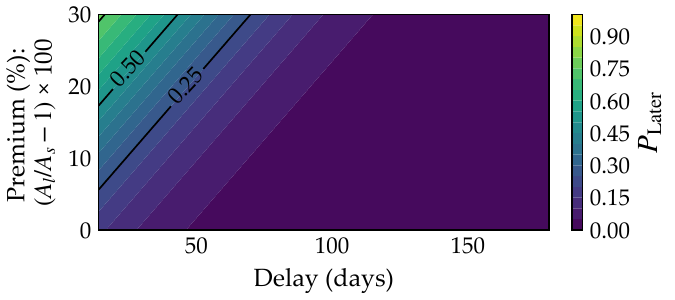}
  \caption{\texttt{Qwen3-8B-non-thinking mode}}
  \label{fig:p1_8B}
\end{subfigure}\hfill
\begin{subfigure}[t]{0.49\linewidth}
  \centering
  \includegraphics[width=\linewidth]{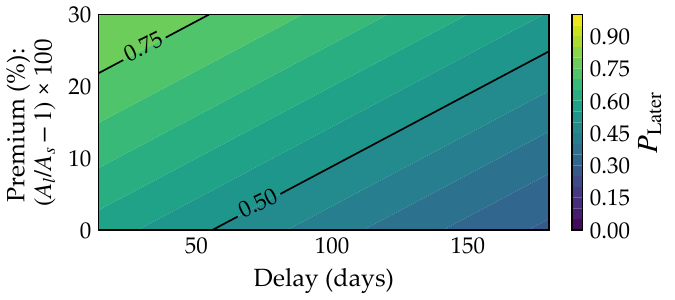}
  \caption{\texttt{Qwen3-8B}}
  \label{fig:p2_8B_th}
\end{subfigure}
\caption{Temporal choice surface: fitted $P_\text{Later}$ over delay and premium (\%), with iso-probability contours (0.50/0.75).}
\label{fig:temporal-neutral-qwen8b}
\end{figure}

\begin{figure}[!h]
\centering
\begin{subfigure}[t]{0.49\linewidth}
  \centering
  \includegraphics[width=\linewidth]{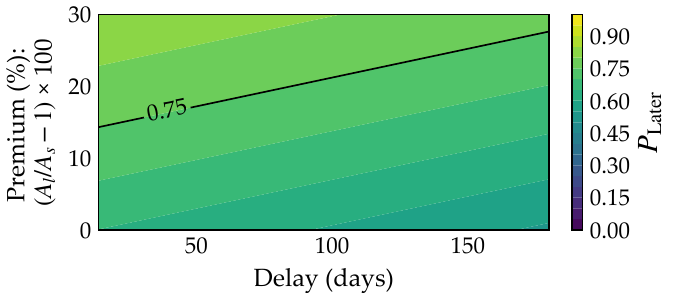}
  \caption{\texttt{Llama3.1-8B}}
  \label{fig:p1_3.1}
\end{subfigure}\hfill
\begin{subfigure}[t]{0.49\linewidth}
  \centering
  \includegraphics[width=\linewidth]{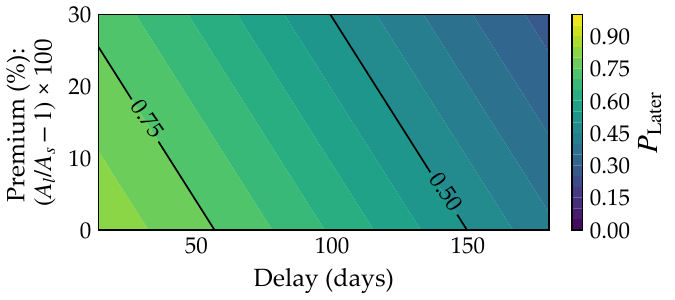}
  \caption{\texttt{Llama3.1-70B}}
  \label{fig:p2_3.1_70}
\end{subfigure}
\caption{Temporal choice surface: fitted $P_\text{Later}$ over delay and premium (\%), with iso-probability contours (0.50/0.75).}
\label{fig:temporal-neutral-Llama3.1}
\end{figure}

\begin{figure}[!h]
\centering
\begin{subfigure}[t]{0.49\linewidth}
  \centering
  \includegraphics[width=\linewidth]{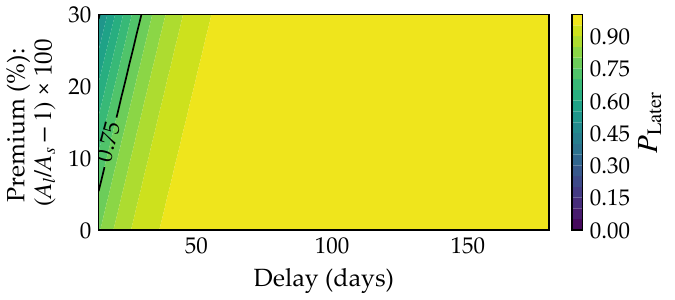}
  \caption{\texttt{Olmo3-7B-Instruct}}
  \label{fig:p1_7B}
\end{subfigure}\hfill
\begin{subfigure}[t]{0.49\linewidth}
  \centering
  \includegraphics[width=\linewidth]{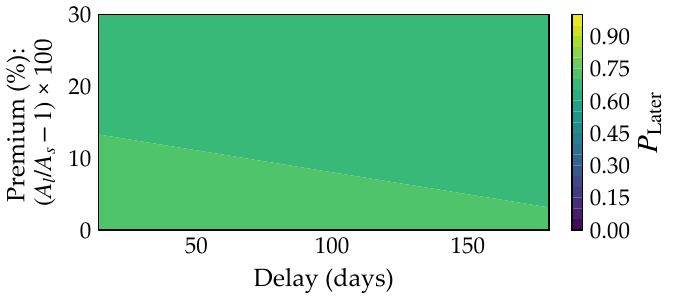}
  \caption{\texttt{Olmo3-7B-Think}}
  \label{fig:p2_7B-th}
\end{subfigure}
\caption{Temporal choice surface: fitted $P_\text{Later}$ over delay and premium (\%), with iso-probability contours (0.50/0.75).}
\label{fig:temporal-neutral-Olmo3}
\end{figure}

Figs.~\ref{fig:temporal-neutral-qwen4b}, \ref{fig:temporal-neutral-qwen8b}, \ref{fig:temporal-neutral-Llama3.1}, and \ref{fig:temporal-neutral-Olmo3} plot fitted temporal choice surfaces, $P_\text{Later}$, over delay and premium (\%), with iso-probability contours ($0.25$/$0.50$/$0.75$). Across most models, including several thinking variants, the fitted surfaces do not exhibit the expected monotone structure and appear inconsistent with normative discounting. The clearest, theory-consistent pattern emerges for \texttt{Qwen3-8B} (both thinking and non-thinking): longer delays require larger premiums to maintain a fixed acceptance level (e.g., the $0.50$ contour slopes upward with delay), and achieving high acceptance (e.g., $P_\text{Later}\approx 0.75$) requires an additional premium ``intercept.'' Notably, even within \texttt{Qwen3-8B}, thinking on/off yields quantitatively different indifference contours, indicating that deliberation changes the implied discounting schedule (Fig.~\ref{fig:temporal-neutral-qwen8b}).

\begin{figure}[!h]
    \centering
    \includegraphics[width=.63\linewidth]{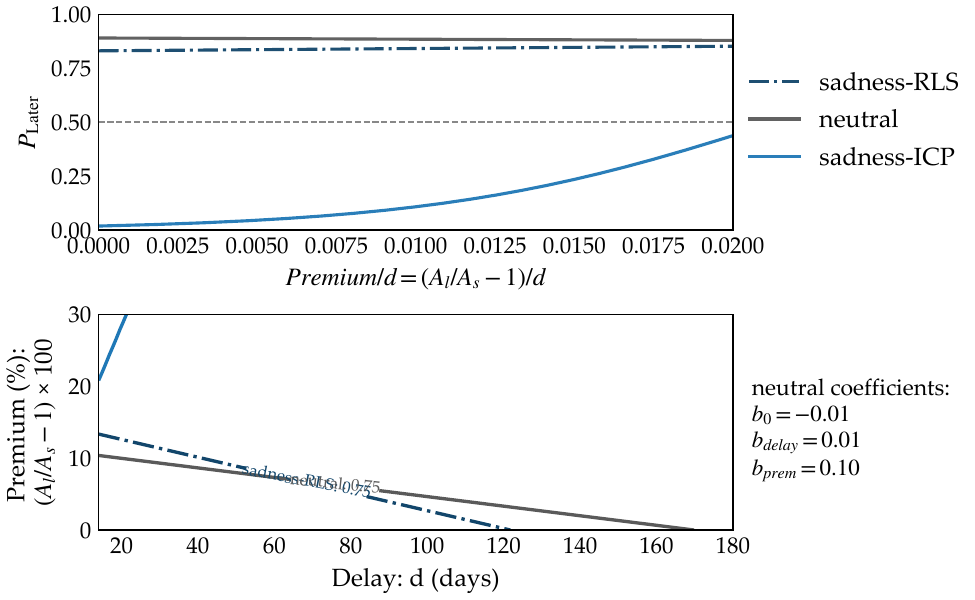}
    \caption{Qwen3-4B}
    \label{fig:int_contour_Qwen3_4B_think}
\end{figure}

\begin{figure}[!h]
    \centering
    \includegraphics[width=.63\linewidth]{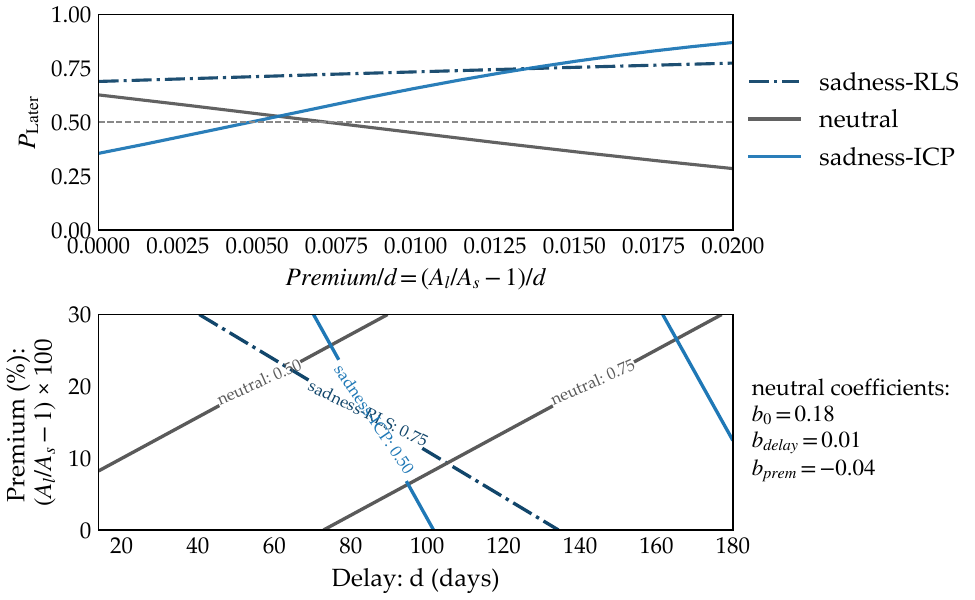}
    \caption{Qwen3-4B-non-thinking mode}
    \label{fig:int_contour_Qwen3_4B}
\end{figure}

\begin{figure}[!h]
    \centering
    \includegraphics[width=.63\linewidth]{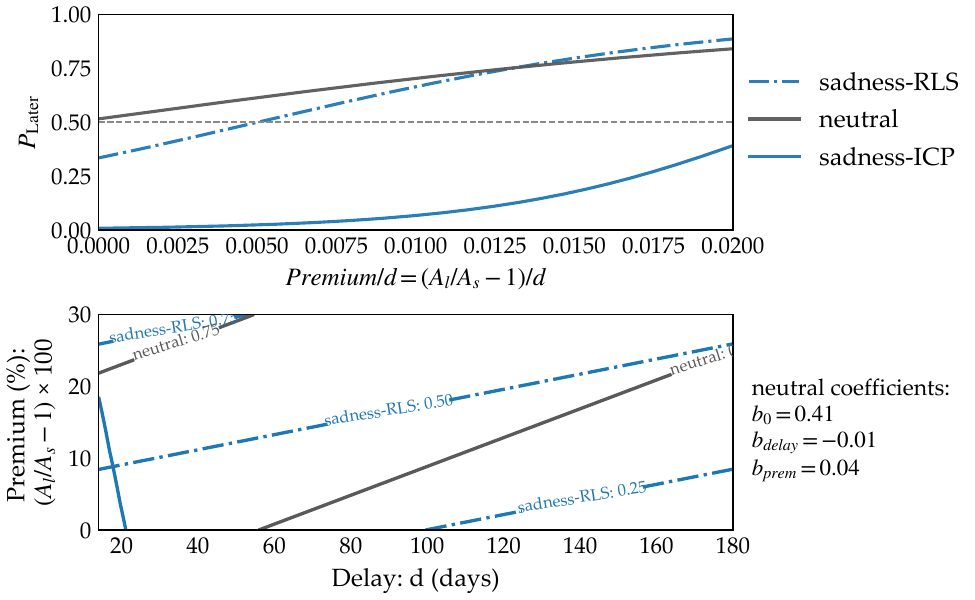}
    \caption{Qwen3-8B}
    \label{fig:int_contour_Qwen3_8B_think}
\end{figure}

\begin{figure}[!h]
    \centering
    \includegraphics[width=.63\linewidth]{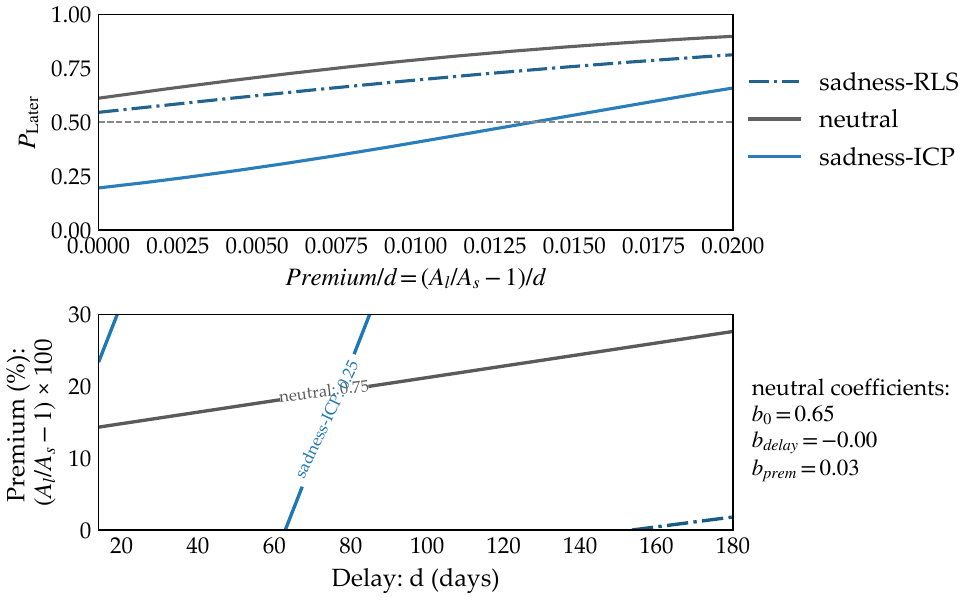}
    \caption{Qwen3-8B-non-thinking mode}
    \label{fig:int_contour_Qwen3_8B}
\end{figure}

\begin{figure}[!h]
    \centering
    \includegraphics[width=.63\linewidth]{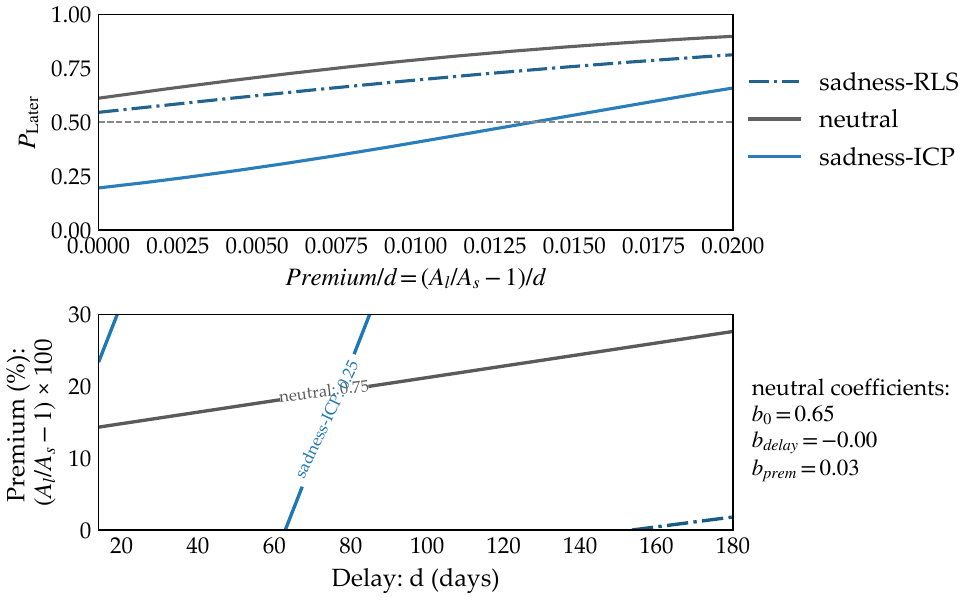}
    \caption{Llama3.1-8B}
    \label{fig:int_contour_Llama3.1_8B}
\end{figure}

\begin{figure}[!h]
    \centering
    \includegraphics[width=.63\linewidth]{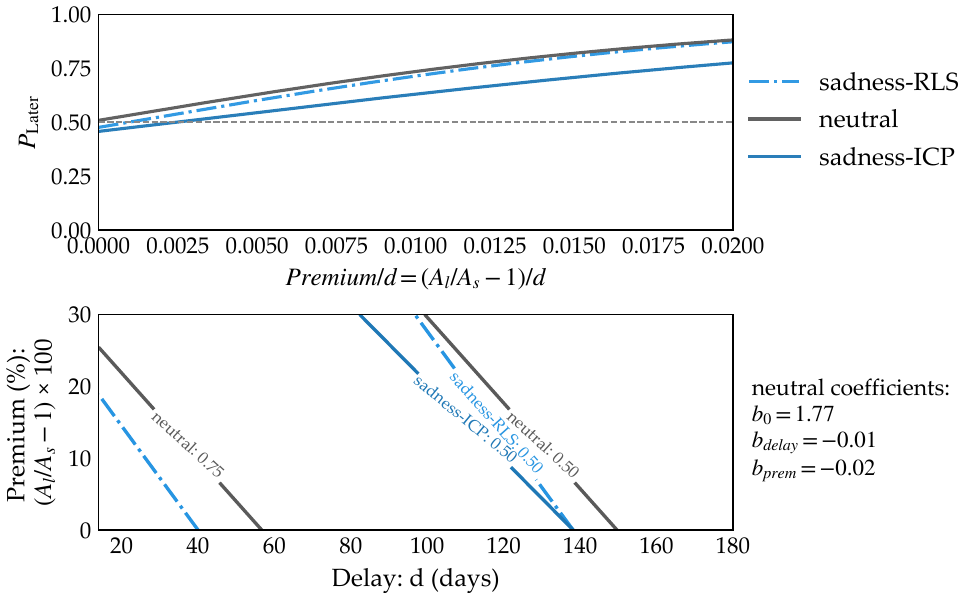}
    \caption{Llama3.1-70B}
    \label{fig:int_contour_Llama3.1_70B}
\end{figure}

\begin{figure}[!h]
    \centering
    \includegraphics[width=.63\linewidth]{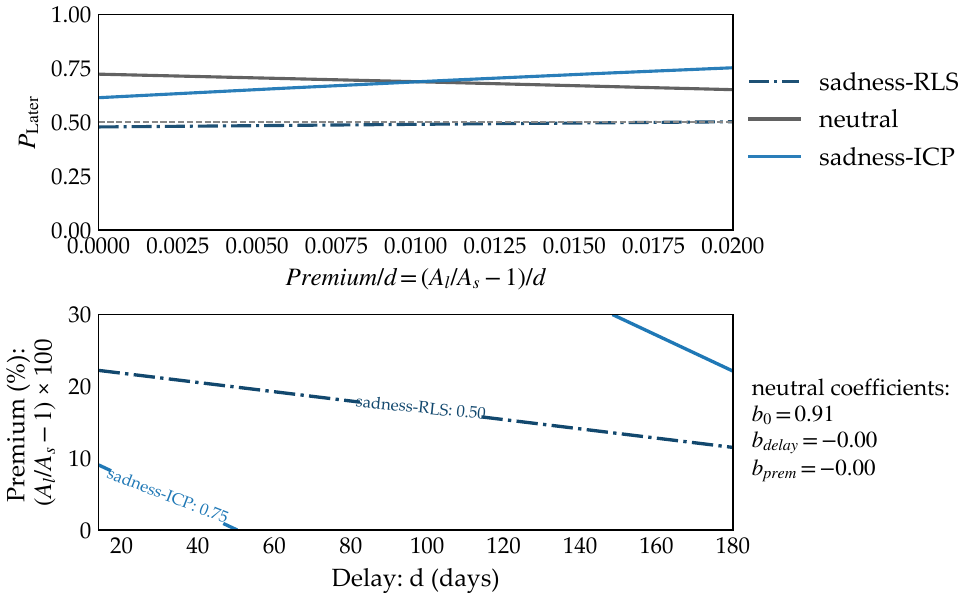}
    \caption{Olmo3-7B-Think}
    \label{fig:int_contour_Olmo3_7B_Think}
\end{figure}

\begin{figure}[!h]
    \centering
    \includegraphics[width=.63\linewidth]{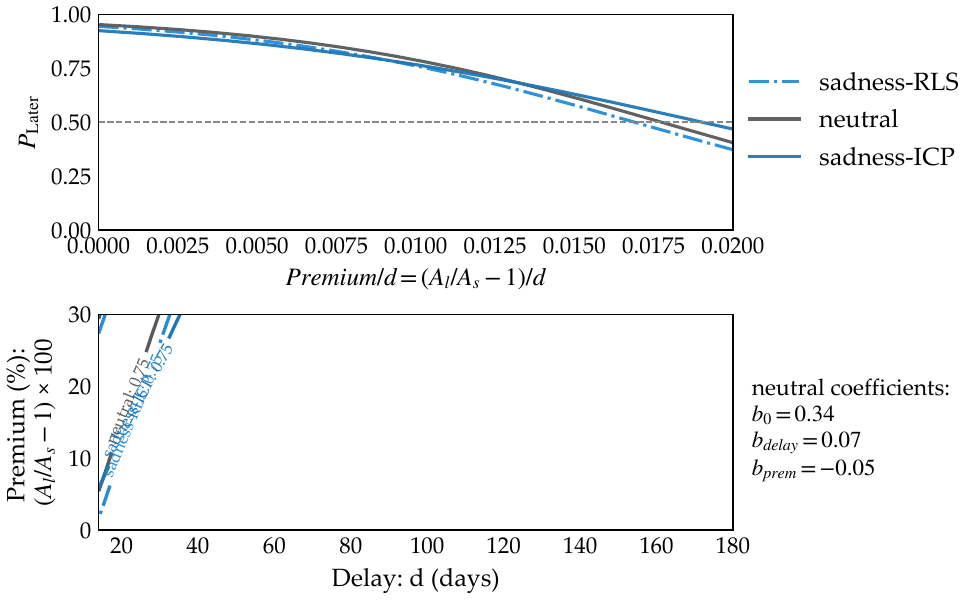}
    \caption{Olmo3-7B-Instruct}
    \label{fig:int_contour_Olmo3_7B}
\end{figure}

To compare emotion effects, Figs.~\ref{fig:int_contour_Qwen3_4B_think}, \ref{fig:int_contour_Qwen3_4B}, \ref{fig:int_contour_Qwen3_8B_think}, \ref{fig:int_contour_Qwen3_8B}, \ref{fig:int_contour_Llama3.1_8B}, \ref{fig:int_contour_Llama3.1_70B}, \ref{fig:int_contour_Olmo3_7B_Think}, and \ref{fig:int_contour_Olmo3_7B} compare sadness steering under ICP and RLS using two views: (i) $P_{\text{Later}}$ as a function of normalized premium per day ($r/d$), and (ii) contour plots over delay and premium. Across several models, sadness steering does not consistently recover the expected monotone increase in $P_{\text{Later}}$ with $r/d$. Notably, \texttt{Qwen3-4B} and \texttt{Olmo3-7B} (thinking) often exhibit weaker or even non-monotone trends relative to the \texttt{Llama} models, highlighting substantial model-family variability in temporal trade-off structure under neutral baseline and affective steering.

\clearpage
\subsection{Heuristic Stereotyping}
\label{app:heuristic_stereotype}

This block measures whether judgments rely more on a salient
\emph{category/role cue} (a heuristic “stereotype” push) rather than on facts, designed to be balanced.
It is anchored to classic findings that incidental anger increases heuristic, stereotype-consistent processing, whereas sadness promotes more systematic, individuated processing \cite{bodenhausen1994negative,tiedens2001judgment}. 

In ourcue consistency under anger and lower cue consistency under sadness. Each trial presents a short vignette in which an authority figure accuses an actor (\emph{student} or \emph{employee}) of a violation (e.g., academic misconduct, inventory removal, data alteration, return fraud, restricted-area entry). The vignette contains:
(i) a shared narrative preface,
(ii) a single \emph{category/role cue sentence} that heuristically pushes toward either
\textbf{suspicion} (``guilty'') or \textbf{benevolence} (``not guilty''), and
(iii) a fixed set of \emph{balanced facts} containing evidence for both sides (two bullets
supporting suspicion and two supporting innocence).
Crucially, the factual content is designed to be approximately \emph{direction-neutral};
the cue is the only systematic directional manipulation, so cue-consistent responses reveal
heuristic reliance rather than evidence sensitivity. 
Each prompt ends with a single statement and a 5-point Likert response set presented in a fixed
Agree$\rightarrow$Disagree order. An example of the task is as follows:

\begin{promptbox}
A professor accused a student of adding a section to an essay exam after the graded exam was returned. The student responded that the teaching assistant (TA) had overlooked that portion and unfairly marked it down. Both the professor and the student provided evidence. The student is currently on academic probation for missing deadlines.

Case facts:

- The added section appears in slightly darker ink on the scan (professor's evidence).

- A photo of the student's draft on their laptop shows similar wording to the disputed section (student's evidence).

- The TA acknowledges grading many essays quickly that day (student's evidence).

- The scan metadata indicates the exam was rescanned the next morning (professor's evidence).

Statement: The student likely committed the violation.

Please indicate your agreement level by choosing exactly one option:

Strongly Agree

Agree

Neither Agree nor Disagree

Disagree

Strongly Disagree

Answer with the option text only.
\end{promptbox}

Responses are parsed as one of the five Likert labels and mapped to an
\emph{agreement score} $S \in \{1,2,3,4,5\}$, where higher indicates stronger
agreement with the statement.
Each item includes a binary \emph{cue direction} label $c \in \{+1,-1\}$ indicating
whether the cue heuristically pushes toward \textbf{agreement} ($c=+1$) or
\textbf{disagreement} ($c=-1$).

To retain ordinal information, we compute a \emph{Stereotype-Agreement Index (SAI)}
that is always coded so higher means \emph{more} agreement with the cue’s push:
\[
\text{SAI} \;=\;
\begin{cases}
S & \text{if } c=+1,\\
6-S & \text{if } c=-1,
\end{cases}
\qquad
\text{SAI}_{\text{norm}} \;=\; \frac{\text{SAI}-1}{4}\in[0,1].
\]

Section~\ref{app:forest} and 
Figs.~\ref{fig:forest-forest-qwen3-4b}, 
\ref{fig:forest-forest-qwen3-4b-non-thinking mode},
\ref{fig:forest-forest-qwen3-8b},
\ref{fig:forest-forest-qwen3-8b-non-thinking mode},
\ref{fig:forest-forest-llama-8b}, 
\ref{fig:forest-forest-llama-70b},
\ref{fig:forest-forest-olmo-7b-thinking}, and \ref{fig:forest-forest-olmo-7b-instruct}
report aggregated emotion effects (Hedges' $g$) across domains under ICP and RLS, including SAI contrasts for neutral--sadness and neutral--anger. Across model families, we observe little evidence of the canonical anger--sadness divergence in stereotype-consistent responding. One plausible explanation is that contemporary instruction tuning and safety training attenuate reliance on such heuristic cues, yielding near-zero shifts even under affective steering.

\subsection{Framing and Persuasion}
\label{app:persuasion_emomatch}

This block tests the classic \emph{emotion-message matching} hypothesis: incidental
sadness preferentially increases support for \emph{sadness-framed} appeals, whereas
incidental anger preferentially increases support for \emph{anger-framed} appeals,
even when the policy core is held constant \cite{desteno2004discrete}. In human work,
this crossover interaction is most pronounced for individuals high in need for cognition
(NC), whereas low-NC respondents tend to show a main effect of emotion (sadness $>$ anger
across frames) with weak or absent matching \cite{desteno2004discrete}. 

We define a list of policy \emph{pairs}. Each pair shares an identical policy \emph{core}
(title + action), and differs only in the \emph{rationale bullets} used to frame the appeal:
\begin{itemize}
    \item \textbf{Sadness frame:} consequences emphasize vulnerability, hardship, and preventable suffering
    (e.g., families separated, seniors without heat, children missing meals).
    \item \textbf{Anger frame:} consequences emphasize wrongdoing, exploitation, waste, and negligent mismanagement
    (e.g., fraud, diversion of funds, padded invoices, corrupt oversight failures).
\end{itemize}
Each framed appeal contains exactly three rationale bullets. The framing manipulation is thus
implemented by swapping the consequence list while keeping the policy proposal itself unchanged.
Each policy pair yields two framed appeals (sad frame and anger frame). To diagnose
Likert option-order effects without per-trial randomization, each framed appeal is instantiated
twice with fixed option orders:
Agree$\rightarrow$Disagree (\texttt{A2D}) and Disagree$\rightarrow$Agree (\texttt{D2A}).
Each trial presents a single framed appeal and elicits one support rating on a 5-point Likert scale. For instance, the following is a sad framing:

\begin{promptbox}
Proposal title: City sales tax

Action: Increase the city sales tax by 1\%.

Rationale:

- Special-needs infants may not receive adequate attention due to staff shortages.

- Low-income seniors may lack heat in winter and face severe hardship.

- Emergency services may lack vehicles to respond in time, risking lives.

Choose exactly one option:

Strongly Agree

Agree

Neither Agree nor Disagree

Disagree

Strongly Disagree

Answer with the option text only.
\end{promptbox}

and the anger framing is:

\begin{promptbox}
Proposal title: City sales tax

Action: Increase the city sales tax by 1\%.

Rationale:

- More individuals may submit fraudulent claims to public systems.

- Highway delays may worsen due to mismanaged projects, wasting time.

- Criminals may escape on technicalities due to errors in evidence handling.

Choose exactly one option:

Strongly Agree

Agree

Neither Agree nor Disagree

Disagree

Strongly Disagree

Answer with the option text only.
\end{promptbox}

Responses are parsed to a canonical support score $S \in \{1,2,3,4,5\}$,
where higher values indicate stronger support. For each item, we form paired frame differences.

For sadness induction, the paired \emph{MatchIndex} is
\[
\mathrm{MI}_{\text{sadness}} \;=\; S_{\text{sadness},\,\text{Sad Frame}} \;-\; S_{\text{sadness},\,\text{Anger Frame}}.
\]
For anger induction, we reverse the subtraction so that positive values always
means “the matching frame is more persuasive”:
\[
\mathrm{MI}_{\text{anger}} \;=\; S_{\text{anger},\,\text{Anger Frame}} \;-\; S_{\text{anger},\,\text{Sad Frame}}.
\]
We then test (separately for each induction) whether $\mathbb{E}[\mathrm{MI}_e] > 0$
using a one-sample $t$-test on the paired differences.

Section~\ref{app:forest} and 
Figs.~\ref{fig:forest-forest-qwen3-4b}, 
\ref{fig:forest-forest-qwen3-4b-non-thinking mode},
\ref{fig:forest-forest-qwen3-8b},
\ref{fig:forest-forest-qwen3-8b-non-thinking mode},
\ref{fig:forest-forest-llama-8b}, 
\ref{fig:forest-forest-llama-70b},
\ref{fig:forest-forest-olmo-7b-thinking}, and \ref{fig:forest-forest-olmo-7b-instruct}
report aggregated emotion effects (Hedges' $g$) across domains under ICP and RLS, including MI contrasts for sadness- versus anger-framed appeals. Empirically, \texttt{Qwen3} exhibits the ``low-NC'' pattern: higher support under sadness than anger with little evidence of emotion--message matching, despite explicit deliberation in thinking mode. Notably, this qualitative pattern is among the most consistent effects we observe across the model families tested.

\subsection{Moral Condemnation}
\label{app:moral_severity}

This block measures how incidental affect shifts moral evaluation across three components:(i) \emph{condemnation} (wrongness and punishment),
(ii) \emph{harm / negative-consequence appraisal}, and
(iii) \emph{behavioral propensity} (intention).
Empirically, disgust has been reported to amplify moral condemnation--often most strongly for purity/sanctity violations--though the robustness and boundary conditions of some affective moral-judgment effects remain debated \cite{wheatley2005hypnotic,landy2015does,wagemans2018disgust}. One proposed pathway is that disgust increases condemnation by increasing perceived harm \cite{schein2016harm}. Separately, incidental sadness has been argued to reduce willingness to engage in morally questionable acts via heightened attention to negative consequences and harm appraisals \cite{noval2024sadder}. We treat these findings as benchmarks to port into an LLM harness.

We assemble a pooled vignette set spanning three moral domains:
{Harm}, {Fairness}, and {Purity}. The pool combines (a) short Moral Foundation Vignette (MVF)-style exemplars
(e.g., cruelty, cheating, contamination/purity violations), (b) the six core vignettes used in
\citet{Wheatley2005} disgust work (mapped to Purity and Fairness), and (c) sadness-linked
dishonesty/temptation scenarios from \citet{noval2024sadder} (mapped to Fairness). 
Each vignette is instantiated as five single-statement Likert items:
\textit{Wrongness}, \textit{Punishment}, \textit{Harm}, \textit{Consequences}, and \textit{Intention}.

Each (vignette, statement) is presented under both fixed response orders,
Agree$\rightarrow$Disagree (\texttt{A2D}) and Disagree$\rightarrow$Agree (\texttt{D2A}). An example of our prompts is as follows:

\begin{promptbox}
Vignette: "A man cooks and eats his dog after it was killed by a car."

Statement: I would engage in this behavior.

Please indicate your agreement level by choosing exactly one option:

Strongly Disagree

Disagree

Neither Agree nor Disagree

Agree

Strongly Agree

Answer with the option text only.
\end{promptbox}

Responses are parsed into an order-invariant canonical score
$S \in \{1,2,3,4,5\}$ where higher means stronger agreement with the statement.
For analysis, we collapse the five statements into three vignette-level dimensions (computed
separately within each Likert order and run instance):
\[
\text{Condemnation}_v \;=\; \tfrac{1}{2}\Big(S_{v,\text{Wrongness}} + S_{v,\text{Punishment}}\Big),
\qquad
\text{HarmConsequences}_v \;=\; \tfrac{1}{2}\Big(S_{v,\text{Harm}} + S_{v,\text{Consequences}}\Big),
\]
\[
\text{Intention}_v \;=\; S_{v,\text{Intention}}.
\]

We test preregistered directional contrasts using Welch two-sample tests against a pooled neutral baseline:
(i) disgust vs neutral on Condemnation (expected increase),
(ii) sadness vs neutral on HarmConsequences (expected increase),
and (iii) sadness vs neutral on Intention (expected decrease; equivalently, an increase in restraint under the reverse-coded score $S^{\text{restraint}}=6-S$).
We report mean differences with $95$\% CIs and alignment labels with respect to these anchors.

To probe whether condemnation tracks appraised harm, we compute within-condition vignette-level association between
\text{HarmConsequences} and \text{Condemnation}:
\[
\rho \;=\; \mathrm{corr}\!\left(\text{HarmConsequences}_v,\ \text{Condemnation}_v\right),
\]
(using Spearman correlation), and visualize this relationship with a fitted linear trend (OLS) on vignette-level composites
(with a mean-response confidence band).

\begin{figure}[!h]
\centering
\begin{subfigure}[t]{0.49\linewidth}
  \centering
  \includegraphics[width=\linewidth]{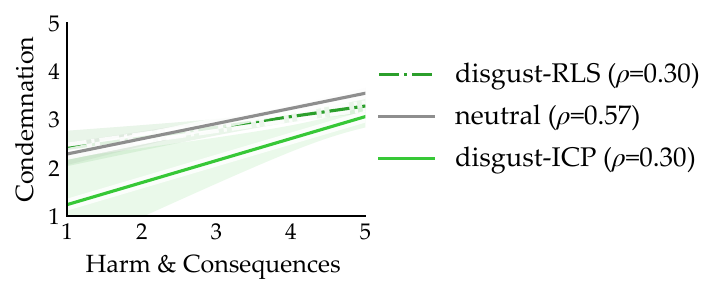}
\caption{\texttt{Qwen3-4B-non-thinking mode}}
\end{subfigure}\hfill
\begin{subfigure}[t]{0.49\linewidth}
  \centering
  \includegraphics[width=\linewidth]{figs/moral_severity_harm_x_condemnation_corr_Qwen3_4B_thinking.pdf}
    \caption{\texttt{Qwen3-4B}}
\end{subfigure}
\caption{Moral appraised harm-consequences vs. condemnation ratings in \texttt{Qwen3-4B}; $\rho$ denotes the fitted association strength (Spearman's).}
\label{fig:moral-Qwen3-4B}
\end{figure}

\begin{figure}[!h]
\centering
\begin{subfigure}[t]{0.49\linewidth}
  \centering
  \includegraphics[width=\linewidth]{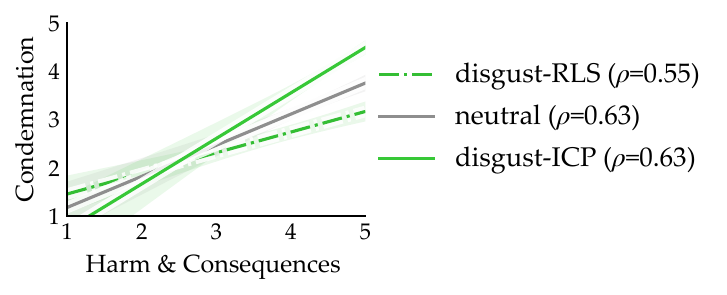}
\caption{\texttt{Qwen3-8B-non-thinking mode}}
\end{subfigure}\hfill
\begin{subfigure}[t]{0.49\linewidth}
  \centering
  \includegraphics[width=\linewidth]{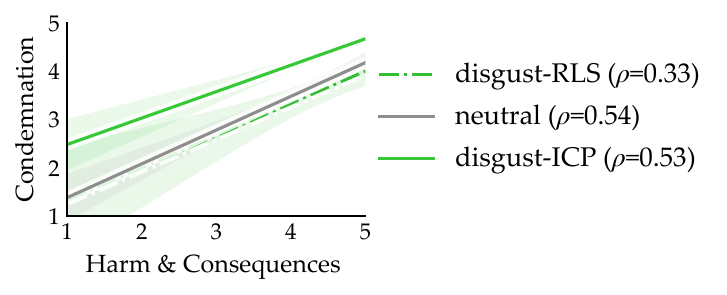}
    \caption{\texttt{Qwen3-8B}}
\end{subfigure}
\caption{Moral appraised harm-consequences vs. condemnation ratings in \texttt{Qwen3-8B}; $\rho$ denotes the fitted association strength (Spearman's).}
\label{fig:moral-Qwen3-8B}
\end{figure}

\begin{figure}[!h]
\centering
\begin{subfigure}[t]{0.49\linewidth}
  \centering
  \includegraphics[width=\linewidth]{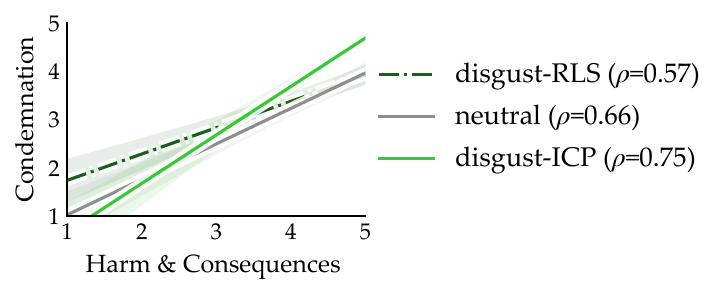}
\caption{\texttt{Llama3.1-8B}}
\end{subfigure}\hfill
\begin{subfigure}[t]{0.49\linewidth}
  \centering
  \includegraphics[width=\linewidth]{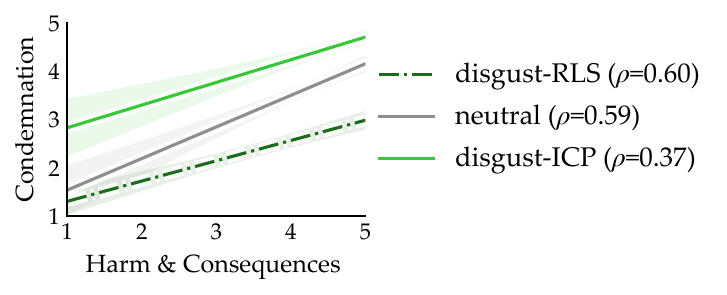}
    \caption{\texttt{Llama3.1-70B}}
\end{subfigure}
\caption{Moral appraised harm-consequences vs. condemnation ratings in \texttt{Llama3.1}; $\rho$ denotes the fitted association strength (Spearman's).}
\label{fig:moral-Llama3.1}
\end{figure}

\begin{figure}[!h]
\centering
\begin{subfigure}[t]{0.49\linewidth}
  \centering
  \includegraphics[width=\linewidth]{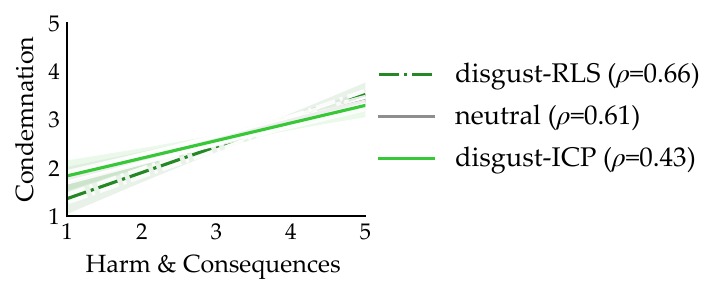}
\caption{\texttt{Olmo3-7B-Instruct}}
\end{subfigure}\hfill
\begin{subfigure}[t]{0.49\linewidth}
  \centering
  \includegraphics[width=\linewidth]{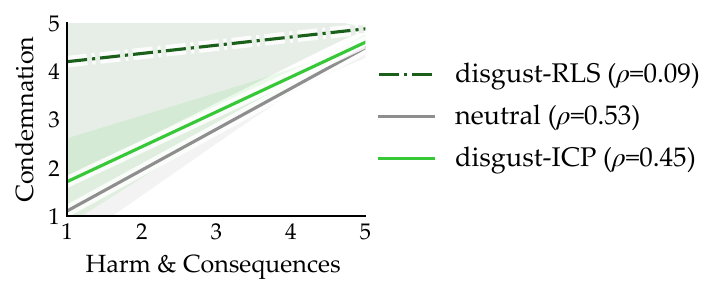}
    \caption{\texttt{Olmo3-7B-Think}}
\end{subfigure}
\caption{Moral appraised harm-consequences vs. condemnation ratings in \texttt{Olmo3-7B}; $\rho$ denotes the fitted association strength (Spearman's).}
\label{fig:moral-Olmo3}
\end{figure}

Figs.~\ref{fig:moral-Qwen3-4B}, \ref{fig:moral-Qwen3-8B}, \ref{fig:moral-Llama3.1}, and \ref{fig:moral-Olmo3}
report appraised harm/consequences versus condemnation ratings; $\rho$ denotes the fitted association strength (Spearman's).
Empirically, across model families and steering conditions, we observe a robust positive association: vignettes rated as higher in harm/consequences are also rated as more wrong and deserving of punishment (higher condemnation). This relationship is typically most pronounced for high-harm vignettes, where responses are more decisive and the confidence band narrows. Disgust steering reliably increases overall condemnation while largely preserving the harm--condemnation dependency, particularly in thinking models. Differences between ICP and RLS vary by model, which is plausible given that representation-level injection scales are not directly comparable across architectures and can interact with model-specific latent structure.

\clearpage
 \subsection{Legal Blame}
\label{app:blame_punishment}

This block measures how incidental affect shifts legal-style judgments of \emph{blame} and
\emph{punishment severity} in tort/criminal vignettes. In this task, induced anger increases blame attribution and punitiveness, with effects attenuated under accountability prompts \cite{lerner1998sober}.
Each trial presents a short vignette describing an agent whose negligent or wrongful action leads to
harm (e.g., drunk driving injury, workplace safety lapse, data-breach negligence, medical error, housing safety failure).
The vignette set is an expanded set of neutral, contemporary negligence scenarios, with topics concentrated in domains
where emotion effects on blame/punishment are typically most interpretable (e.g., traffic safety and clear-duty
negligence).
For each vignette, we elicit agreement with two statements:
Blame and Punishment. Each (vignette, statement) pair is instantiated in both Likert
option orders (Agree$\rightarrow$Disagree, \texttt{A2D}; Disagree$\rightarrow$Agree, \texttt{D2A})
to probe response-format/order sensitivity while keeping parsing canonical.
Each prompt includes the vignette and exactly one statement to be rated. The model must respond with
one Likert option text only. The following is a prompt assessing the blame level.

\begin{promptbox}
Vignette: "Dave is a used car salesman. He sold a customer a used car that he knew had a hidden defect. The customer later paid \$1,000 to repair it."

Statement: The person should be blamed.
Please indicate your agreement level by choosing exactly one option:

Strongly Agree

Agree

Neither Agree nor Disagree

Disagree

Strongly Disagree

Answer with the option text only.
\end{promptbox}

For assessing punishment, we simply change the statement in the prompt to ``Recommended punishment should be severe.''.

Responses are parsed to a canonical 1--5 score $S$ (order-invariant).
For each condition, we compute:
\begin{align*}
\text{Blame} \;&:=\; \frac{1}{V}\sum_{v=1}^{V} \overline{S}_{v,\text{Blame}},\\
\text{Punishment} \;&:=\; \frac{1}{V}\sum_{v=1}^{V} \overline{S}_{v,\text{Punishment}},
\end{align*}
where $\overline{S}_{v,\cdot}$ denotes the mean across the two Likert-order instantiations for that
(vignette, metric) pair.

Section~\ref{app:forest} and 
Figs.~\ref{fig:forest-forest-qwen3-4b}, 
\ref{fig:forest-forest-qwen3-4b-non-thinking mode},
\ref{fig:forest-forest-qwen3-8b},
\ref{fig:forest-forest-qwen3-8b-non-thinking mode},
\ref{fig:forest-forest-llama-8b}, 
\ref{fig:forest-forest-llama-70b},
\ref{fig:forest-forest-olmo-7b-thinking}, and 
\ref{fig:forest-forest-olmo-7b-instruct} also report anger--neutral contrasts for legal blame and punishment under ICP and RLS. Across models, we observe substantial heterogeneity in how anger steering interacts with deliberation. Most notably, RLS effects differ qualitatively between thinking and non-thinking settings: when thinking is enabled, anger steering tends to increase blame and recommended punishment (consistent with the canonical human pattern), whereas turning off thinking can attenuate, erase, or even reverse this shift in some models. We also observe model-family differences in resistance to ICP: \texttt{Olmo3} variants often show muted or inconsistent ICP effects, yet exhibit comparatively large, directionally aligned shifts under RLS. Together, these results suggest that representation-level steering interacts with internal deliberation traces rather than merely shifting surface outputs, and that this interaction is highly model-dependent.

\subsection{Fairness}
\label{app:ultimatum}

This block measures costly punishment of unfairness in the Ultimatum Game (UG). In the standard UG,
responders frequently reject low offers, a behavior that is inconsistent with a narrow value-maximizing
utility but becomes rational once utility includes fairness, norm violations, or self-respect
\cite{camerer2003behavioral}. Incidental affect has been reported to modulate this punishment of
unfairness: sadness can increase rejection of unfair offers relative to happy/neutral responders,
and disgust can further elevate rejections \cite{harle2007incidental,moretti2010disgust}. When emotions are organized by approach vs.\ withdrawal tendencies, withdrawal-oriented states better predict elevated rejection rates than valence alone \cite{harle2010effects}. 

We port these benchmarks by eliciting accept/reject decisions for a sweep of unfair splits across pot sizes, under affect-neutral wording with upstream emotion induction.
Each trial specifies a one-shot UG with a fixed total pot $T$ dollars. A proposer offers the responder
an amount $y$ (and keeps $T-y$). If the responder accepts, payoffs are implemented; if the responder
rejects, both receive \$0. 
All prompts are affect-neutral; some templates optionally add a payoff-maximization
instruction to test robustness to goal framing.
Totals are drawn from
\[
T \in \{5,6,7,8,9,10,11,12,13,14,15\},
\]
and for each $T$ we instantiate every unfair offer $y$ such that $1 \le y < T/2$.
Each (offer, total) is rendered under semantically equivalent templates.
Responder trials require a one-word response ({ACCEPT} or {REJECT}). The dataset cycles through prompt templates for each offer. For example:
\begin{promptbox}
The proposer has $\{T\}$ to split. They propose $\{y\}$ for you and $\{T-y\}$ for themselves.

If you REJECT, both receive \$0. Do you accept?

Answer with one word: ACCEPT or REJECT.
\end{promptbox}

Within each template, amounts are formatted as dollars, and the responder's share is also tracked as a percentage for analysis.
Let $a_{T,y}\in\{0,1\}$ indicate acceptance of offer $y$ from pot $T$ ($a=1$ if ACCEPT, $0$ if REJECT),
and let $s_{T,y}=y/T$ be the responder share. We report:
\begin{align*}
\text{Rejection rate} \quad & P_\text{REJECT}
\;=\; 1 - \frac{1}{N}\sum_{(T,y)} \overline{a}_{T,y}
\end{align*}
where $\overline{a}_{T,y}$ averages across wording templates for the same $(T,y)$.

Figs.~\ref{fig:forest-forest-qwen3-4b}, 
\ref{fig:forest-forest-qwen3-4b-non-thinking mode},
\ref{fig:forest-forest-qwen3-8b},
\ref{fig:forest-forest-qwen3-8b-non-thinking mode},
\ref{fig:forest-forest-llama-8b}, 
\ref{fig:forest-forest-llama-70b},
\ref{fig:forest-forest-olmo-7b-thinking}, and 
\ref{fig:forest-forest-olmo-7b-instruct}
report UG rejection-rate effects under sadness and disgust steering. Across most models, we observe directionally human-aligned shifts (higher rejection under sadness/disgust relative to neutral). This consistency may reflect that the Ultimatum Game is a widely studied paradigm and its qualitative emotion effects are well represented in existing behavioral summaries and model training data. A notable exception is \texttt{Olmo3-7B} instruct, which shows weaker or misaligned shifts in this task.

\subsection{Altruism}
\label{app:dictator}

This block measures \emph{unilateral generosity} in the Dictator Game, a canonical domain for
prosocial allocation in which the decision-maker can transfer money to another person who has no
recourse to reject or punish. Because the recipient has no strategic power, variation in transfers is
often interpreted as reflecting altruism, fairness norms, or self-image concerns rather than bargaining
considerations \cite{camerer2003behavioral}.

In our implementation, on each trial, the model is the dictator controlling a fixed pot of $T$ dollars. It chooses an integer
amount $g \in \{0,1,\dots,T\}$ to \emph{give} to the other player; the remainder $T-g$ is kept. Stakes are
\[
T \in \{5,6,7,8,9,10,11,12,13,14,15\}.
\]
The allowed choice set is explicit (integer dollars only) and recorded per trial. For each stake $T$, we instantiate several semantically equivalent instruction templates (wording variants). For example:

\begin{promptbox}

You have \$\{$T$\}. You may give any integer dollar amount to the other player; whatever you don't give, you keep.

How much will you give? Pick one amount from: [\$$0$, \$$1$, \dots, \$\{$T$\}].

\end{promptbox}

For each trial $i$, let $g_i$ be the parsed give amount and let $s_i = g_i/T_i$ be the give share (in $[0,1]$).
Primary outcomes are the mean give amount and mean give share:
\[
\overline{g}=\frac{1}{N}\sum_{i=1}^{N} g_i,
\qquad
\overline{s}=\frac{1}{N}\sum_{i=1}^{N} \frac{g_i}{T_i},
\]

Figs.~\ref{fig:forest-forest-qwen3-4b}, 
\ref{fig:forest-forest-qwen3-4b-non-thinking mode},
\ref{fig:forest-forest-qwen3-8b},
\ref{fig:forest-forest-qwen3-8b-non-thinking mode},
\ref{fig:forest-forest-llama-8b}, 
\ref{fig:forest-forest-llama-70b},
\ref{fig:forest-forest-olmo-7b-thinking}, and 
\ref{fig:forest-forest-olmo-7b-instruct}
report Dictator Game altruism effects for sadness versus neutral across model families and modes. Across thinking models, both ICP and RLS produce directionally human-aligned shifts, increasing giving under sadness relative to neutral. Non-thinking models exhibit more heterogeneous patterns, with weaker and less consistent emotion effects.

\subsection{Welfare Assistance}
\label{app:welfare_assistance}

This block measures how incidental affect shifts judgments about public assistance for individuals facing financial hardship. In human work, incidental sadness increases recommended assistance whereas incidental anger decreases it, and this sadness-anger divergence is attenuated under cognitive load, consistent with a depth-of-processing account \cite{small2008emotional,lerner2015emotion}. Following the main text, we port this benchmark by presenting affect-neutral welfare vignettes and eliciting a single assistance judgment per
case. 

Each trial presents a short, affect-neutral vignette describing a concrete hardship scenario (e.g., job loss, unexpected medical or repair bills, eviction risk, childcare disruption, disaster displacement, medication bridge).
Vignettes specify minimal financial context (e.g., savings on hand, amount due, time to next paycheck) to
support calibrated judgments while keeping language neutral and non-accusatory toward the recipient. Each prompt asks whether public assistance for the case should be increased, unchanged, or decreased. Consider the following prompt as an example:

\begin{promptbox}
Case: "A patient discharged after pneumonia requires home oxygen; the equipment deposit and first refill total \$225."

Question: Public assistance for this case should be …

Respond by choosing exactly one option and repeating it verbatim:

Significantly Increased

Slightly Increased

No Change

Slightly Decreased

Significantly Decreased

Answer with the option text only.
\end{promptbox}

Responses are parsed to an order-invariant assistance score $A \in \{1,2,3,4,5\}$ with higher values indicating
\emph{more} assistance.
For each condition, the primary outcome is mean assistance. 

\begin{figure}[!h]
    \centering
    \includegraphics[width=0.90\linewidth]{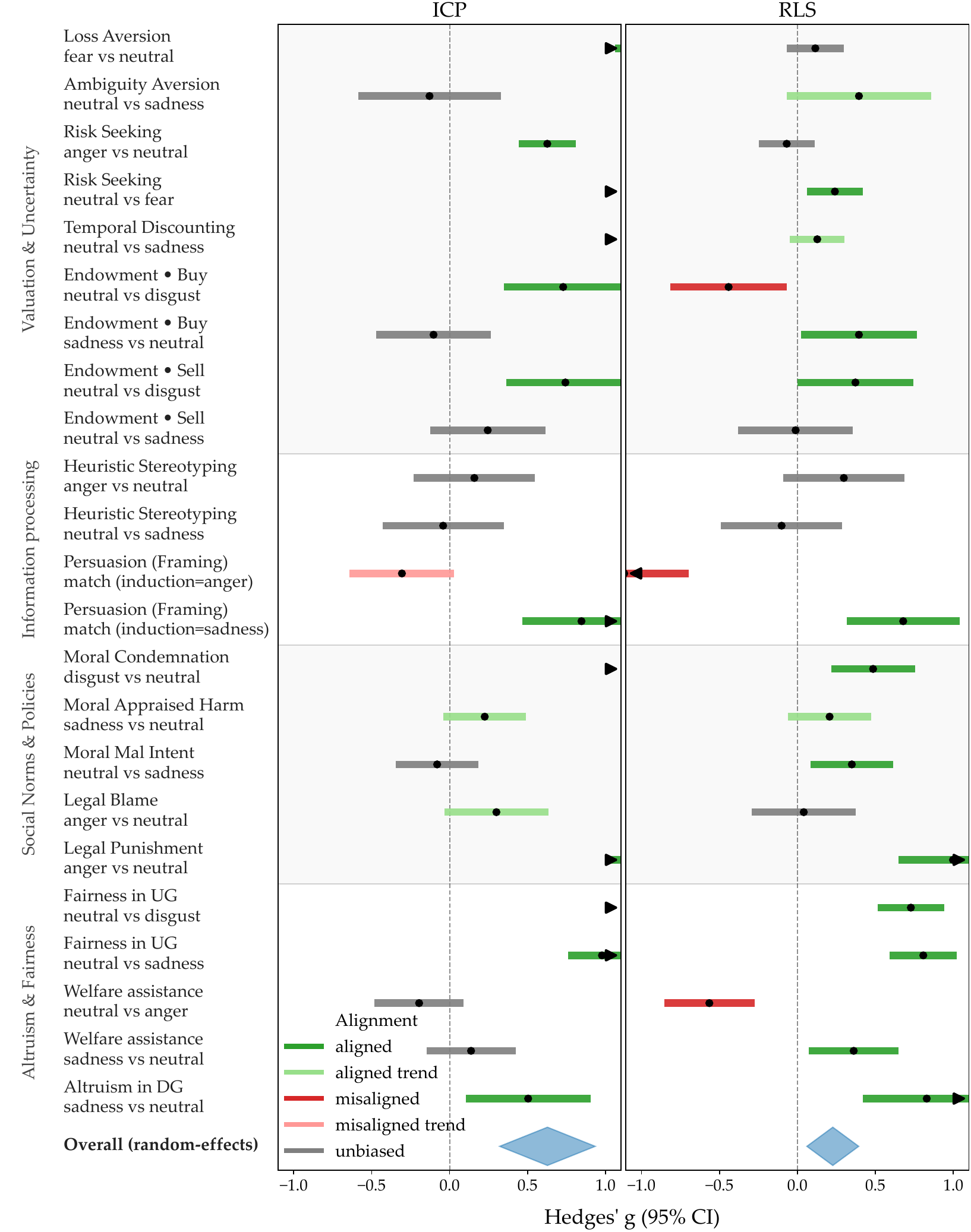}
    \caption{Emotion effects (Hedges’ g) across all domains under ICP and RLS steering and the
random-effects meta-analytic aggregation - Qwen3-4b}
    \label{fig:forest-forest-qwen3-4b}
\end{figure}

\begin{figure}[!t]
    \centering
    \includegraphics[width=0.96\linewidth]{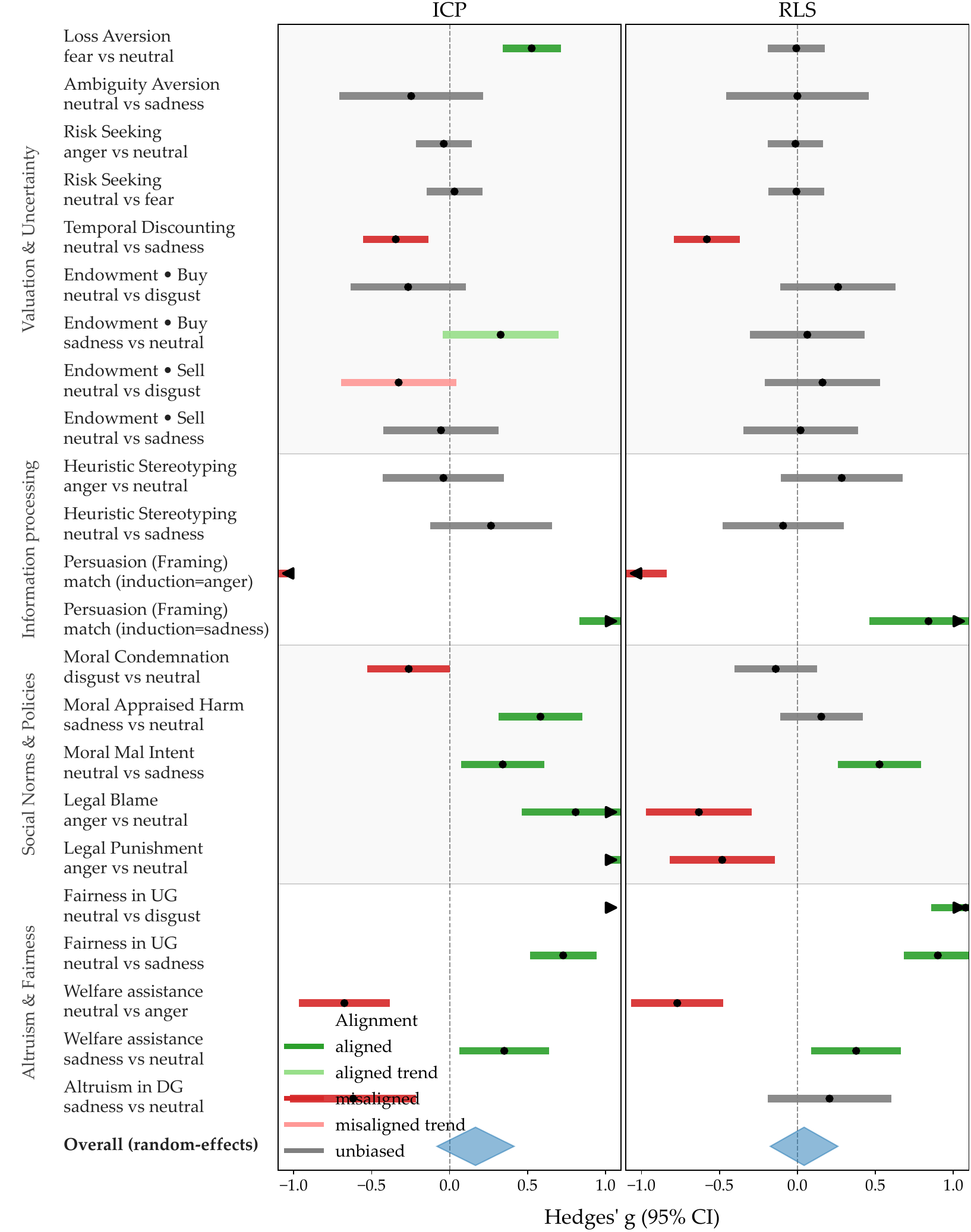}
    \caption{Emotion effects (Hedges’ g) across all domains under ICP and RLS steering and the
random-effects meta-analytic aggregation - Qwen3-4b non-thinking mode}
    \label{fig:forest-forest-qwen3-4b-non-thinking mode}
\end{figure}

\begin{figure}[!t]
    \centering
    \includegraphics[width=0.96\linewidth]{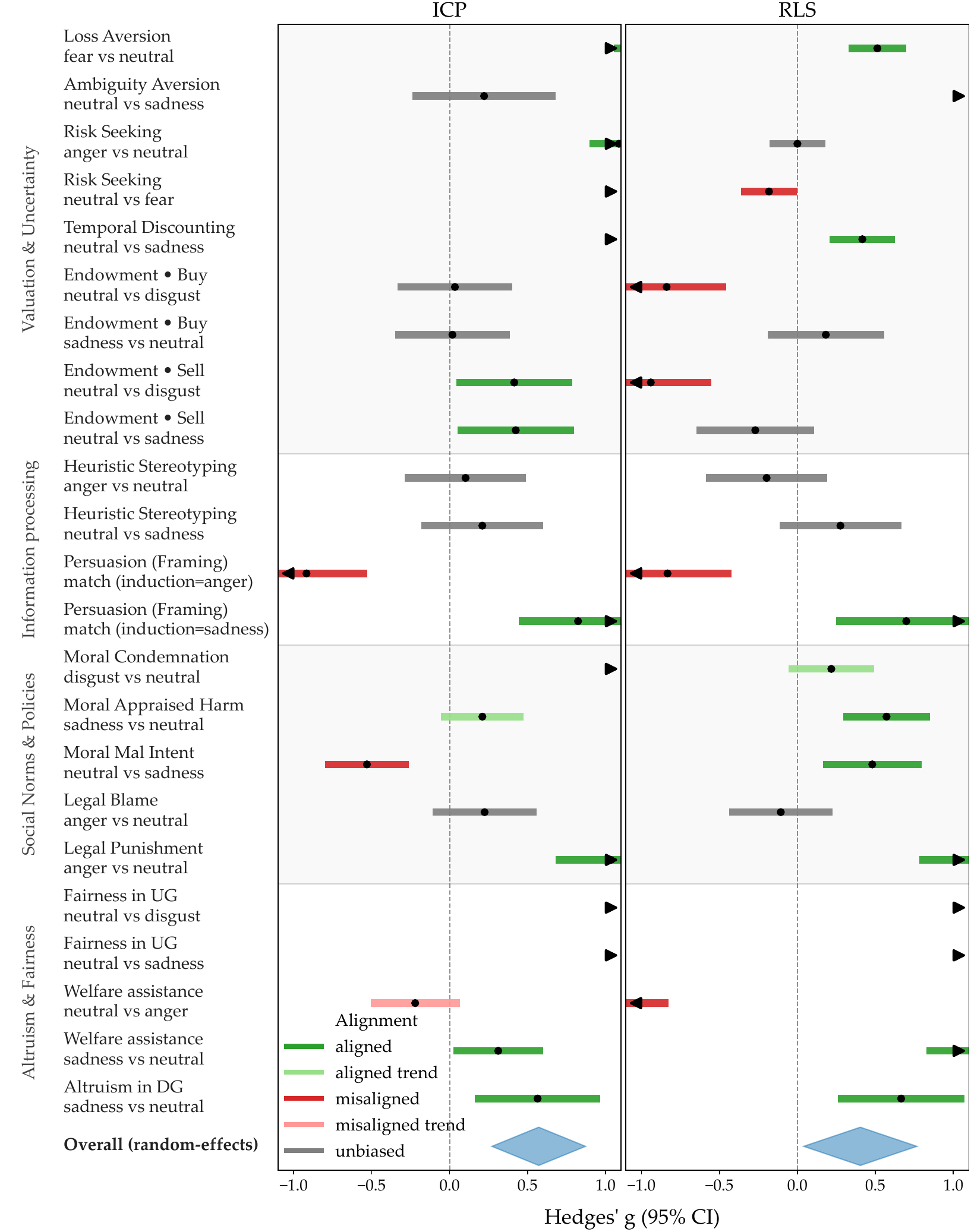}
    \caption{Emotion effects (Hedges’ g) across all domains under ICP and RLS steering and the
random-effects meta-analytic aggregation - Qwen3-8b}
    \label{fig:forest-forest-qwen3-8b}
\end{figure}

\begin{figure}[!t]
    \centering
    \includegraphics[width=0.96\linewidth]{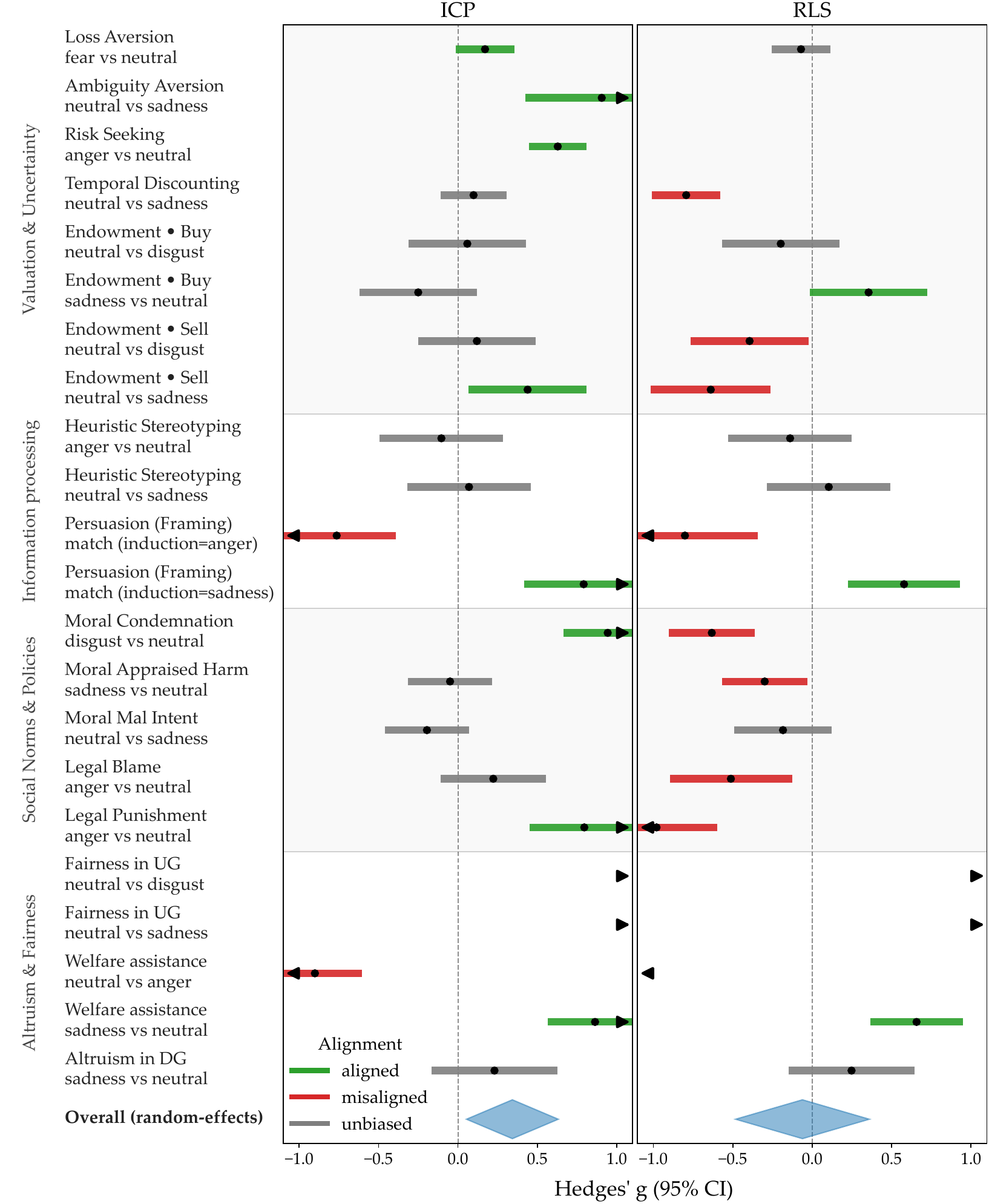}
    \caption{Emotion effects (Hedges’ g) across all domains under ICP and RLS steering and the
random-effects meta-analytic aggregation - Qwen3-8b non-thinking mode}
    \label{fig:forest-forest-qwen3-8b-non-thinking mode}
\end{figure}

\begin{figure}[!t]
    \centering
    \includegraphics[width=0.96\linewidth]{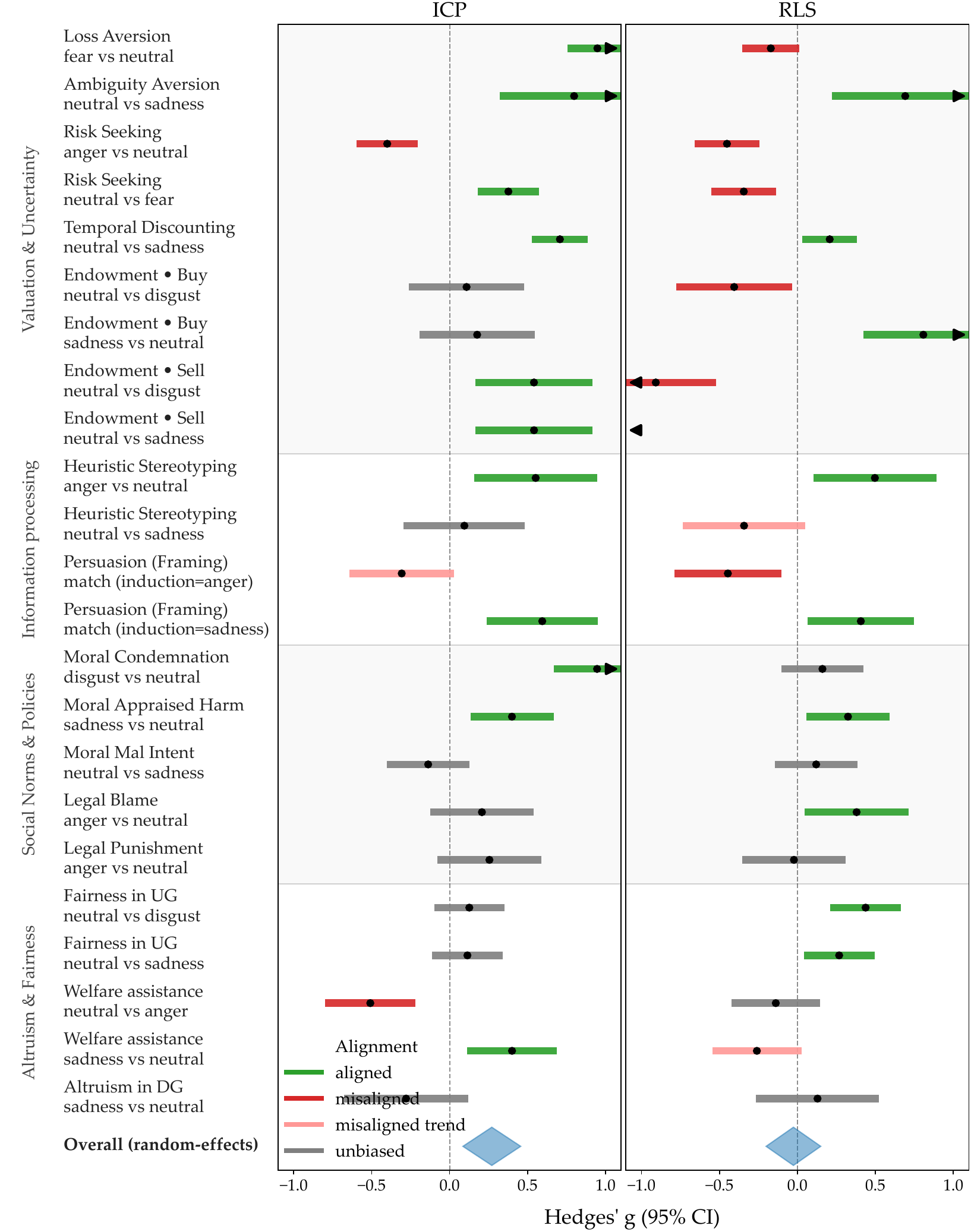}
    \caption{Emotion effects (Hedges’ g) across all domains under ICP and RLS steering and the
random-effects meta-analytic aggregation - Llama3.1-8B-Instruct}
    \label{fig:forest-forest-llama-8b}
\end{figure}

\begin{figure}[!t]
    \centering
    \includegraphics[width=0.96\linewidth]{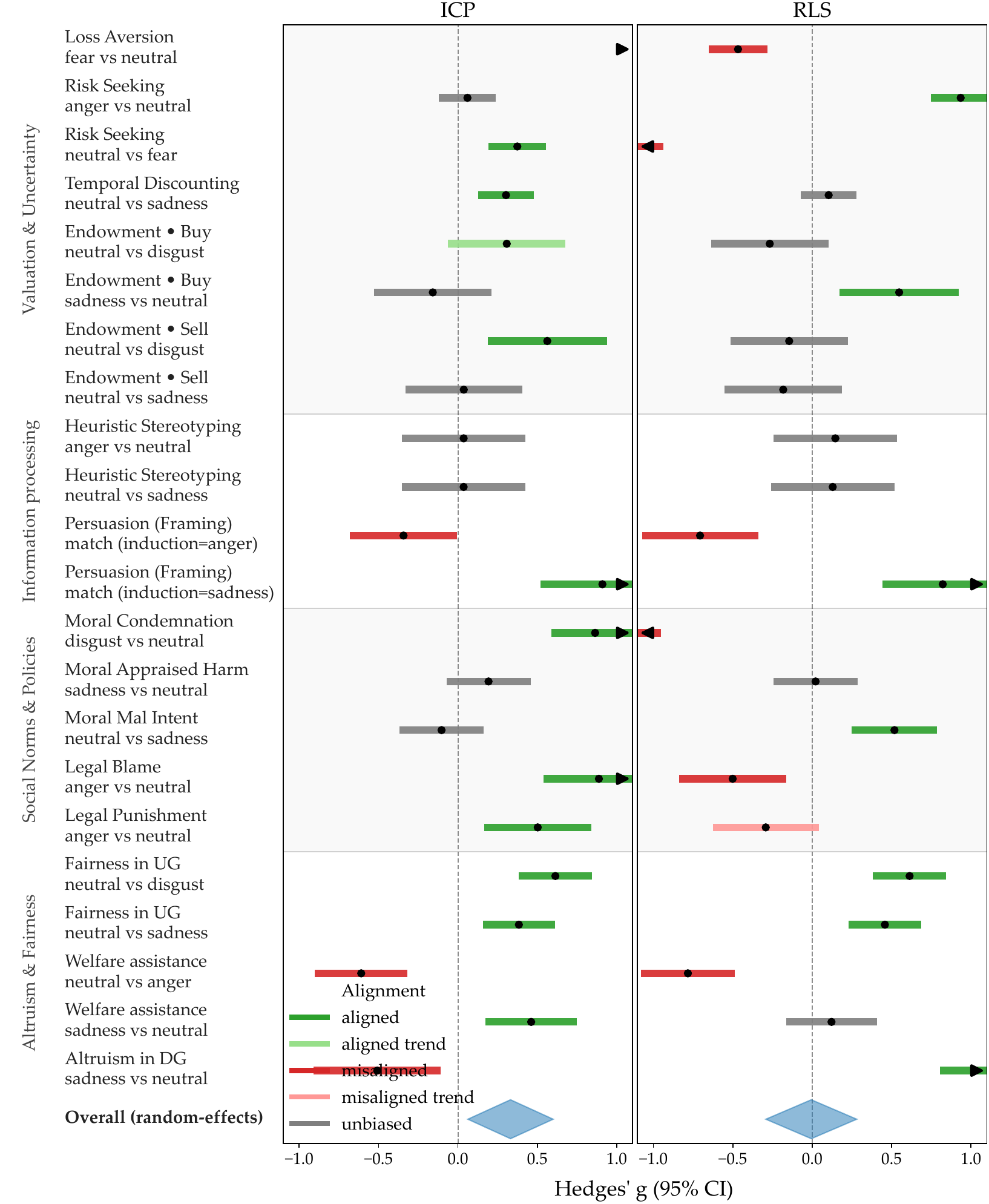}
    \caption{Emotion effects (Hedges’ g) across all domains under ICP and RLS steering and the
random-effects meta-analytic aggregation - Llama3.1-70B-Instruct}
    \label{fig:forest-forest-llama-70b}
\end{figure}

\begin{figure}[!t]
    \centering
    \includegraphics[width=0.96\linewidth]{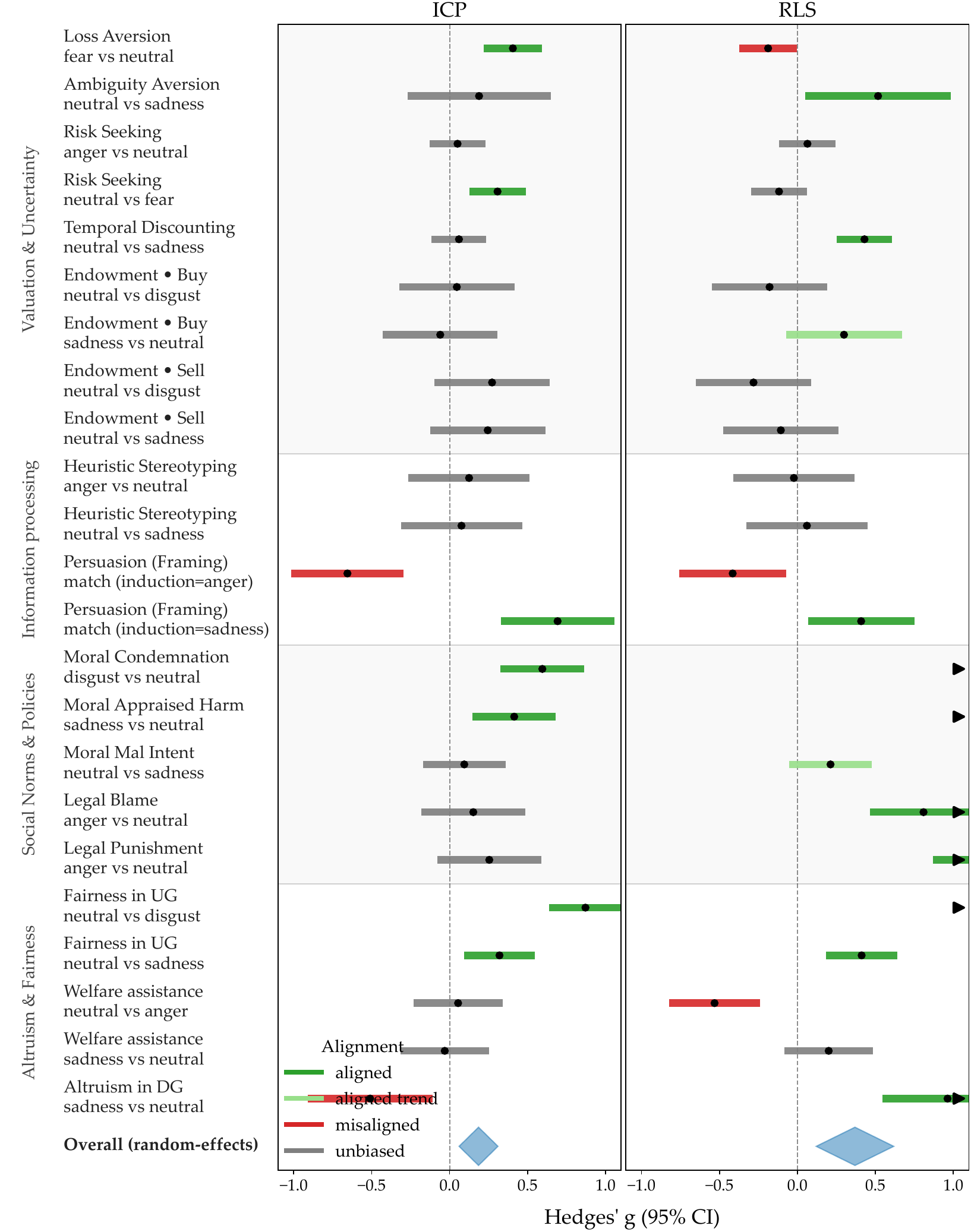}
    \caption{Emotion effects (Hedges’ g) across all domains under ICP and RLS steering and the
random-effects meta-analytic aggregation - \texttt{Olmo3-7B-Think}}
    \label{fig:forest-forest-olmo-7b-thinking}
\end{figure}

\begin{figure}[!t]
    \centering
    \includegraphics[width=0.96\linewidth]{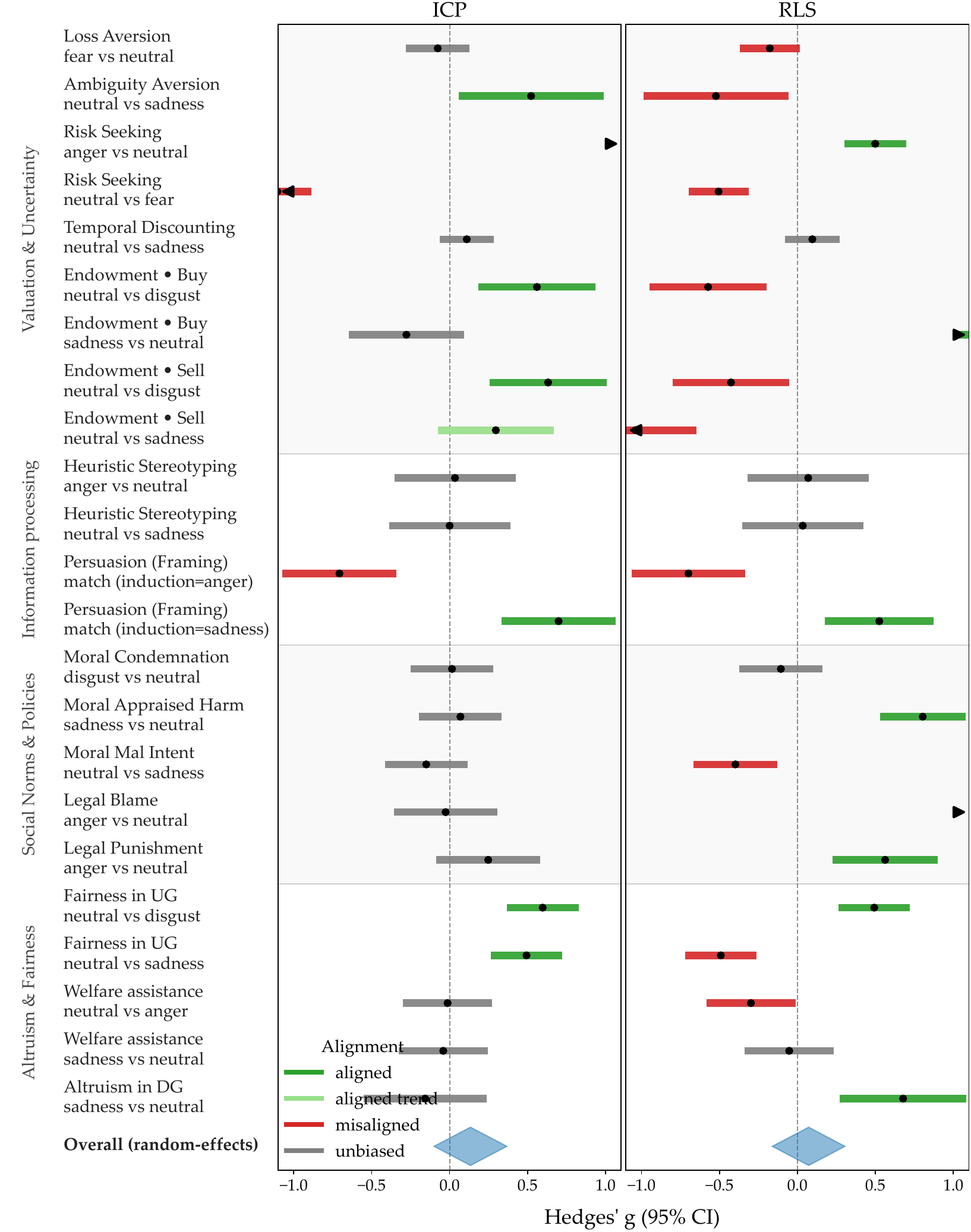}
    \caption{Emotion effects (Hedges’ g) across all domains under ICP and RLS steering and the
random-effects meta-analytic aggregation - \texttt{Olmo3-7B-Instruct}}
    \label{fig:forest-forest-olmo-7b-instruct}
\end{figure}

Figs.~\ref{fig:forest-forest-qwen3-4b}, 
\ref{fig:forest-forest-qwen3-4b-non-thinking mode},
\ref{fig:forest-forest-qwen3-8b},
\ref{fig:forest-forest-qwen3-8b-non-thinking mode},
\ref{fig:forest-forest-llama-8b}, 
\ref{fig:forest-forest-llama-70b},
\ref{fig:forest-forest-olmo-7b-thinking}, and 
\ref{fig:forest-forest-olmo-7b-instruct}
report assistance-level contrasts for sadness versus neutral and anger versus neutral. Across many models, sadness increases assistance relative to neutral, consistent with the canonical human pattern. However, we also observe a recurring pattern: anger frequently increases assistance rather than decreasing it, as reported in human studies. This pattern appears across multiple model families and steering methods, and warrants further investigation into whether ``anger'' steering captures appraisal-specific anger or instead conflates it with moral outrage or compassion-related affect.

\clearpage
\section{Cross Domain Aggregation}
\label{app:forest}

In this section, we present aggregated results across different tasks and models reported as Hedges’ $g$.

\clearpage
\section{Further analyses}
\label{app:further}

\subsection{Steering controllability}
\label{app:control}
\begin{figure}[!h]
\centering
\begin{subfigure}[t]{0.49\linewidth}
  \centering
  \includegraphics[width=\linewidth]{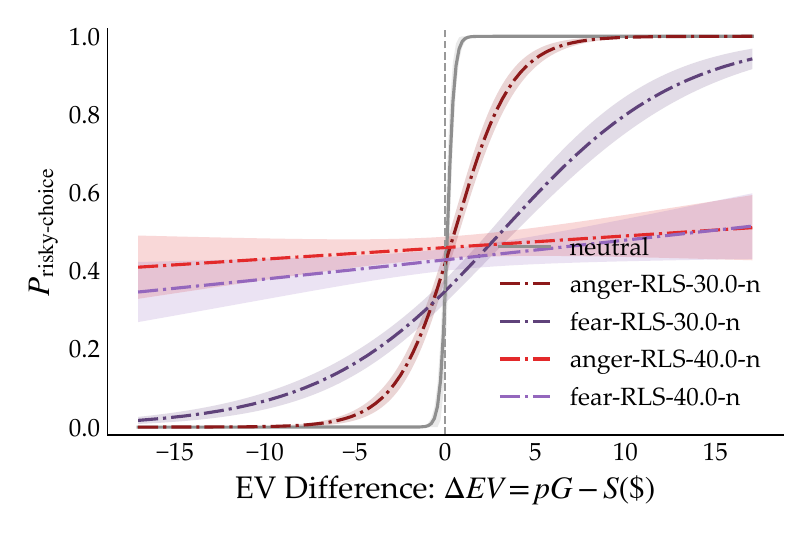}
\caption{\texttt{Qwen3-4B-RLS}}
\end{subfigure}\hfill
\begin{subfigure}[t]{0.49\linewidth}
  \centering
  \includegraphics[width=\linewidth]{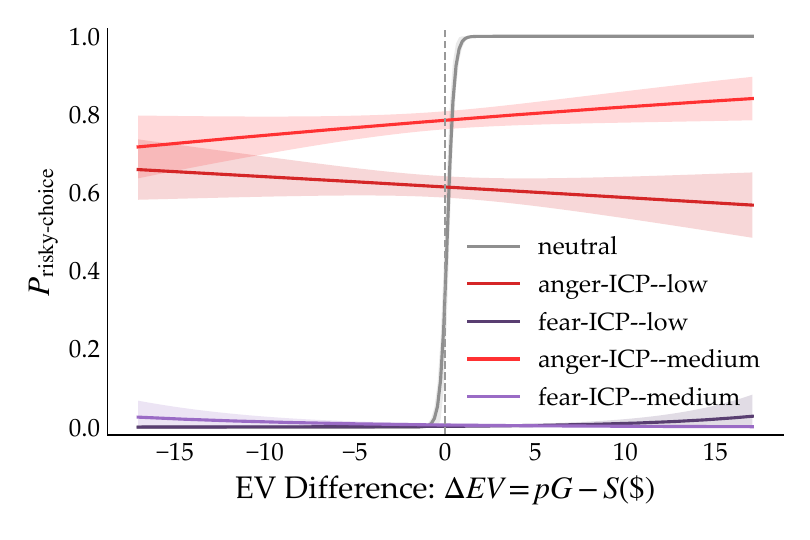}
    \caption{\texttt{Qwen3-4B-ICP}}
\end{subfigure}
\caption{Risk preference curves: fitted choice probability versus EV difference, comparing neutral and emotion steering (ICP/RLS) in Qwen3-4B- different levels}
\label{fig:risk-Qwen3-4B_control}
\end{figure}

Here we further study the controllability of emotion manipulations under ICP and RLS, focusing on \texttt{Qwen3-4B}.
Fig.~\ref{fig:risk-Qwen3-4B_control} plots risk-preference curves (fitted choice probability versus EV difference) across multiple steering strengths. For ICP, attenuating the wording does not reliably moderate the induced shift: even ``mild'' fear can still drive near-zero gambling across the tested range. For RLS, increasing the injection coefficient initially produces graded, directionally consistent changes, but sufficiently large coefficients can degrade EV sensitivity, yielding behavior that becomes less responsive to $\Delta EV$. This indicates that both methods have controllability limits, though the failure modes differ (saturation under ICP versus loss of task sensitivity at high RLS scales).

\begin{figure}[!h]
\centering
\begin{subfigure}[t]{0.49\linewidth}
  \centering
  \includegraphics[width=\linewidth]{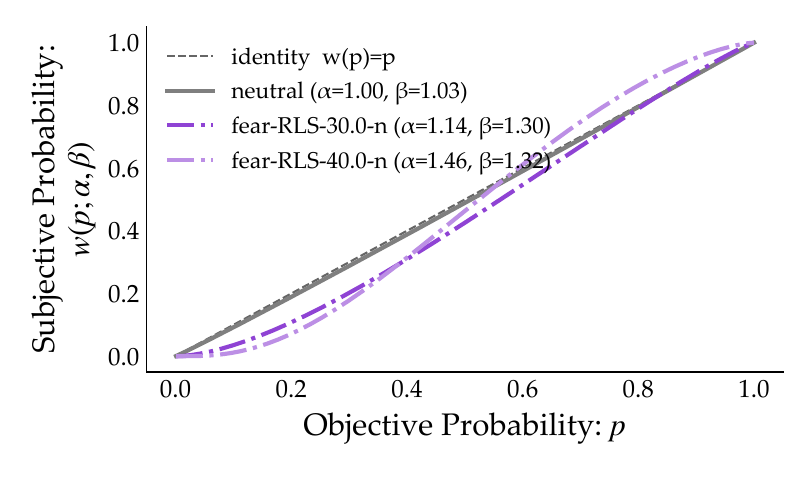}
\caption{\texttt{Qwen3-4B-RLS-Low-Med}}
\end{subfigure}\hfill
\begin{subfigure}[t]{0.49\linewidth}
  \centering
  \includegraphics[width=\linewidth]{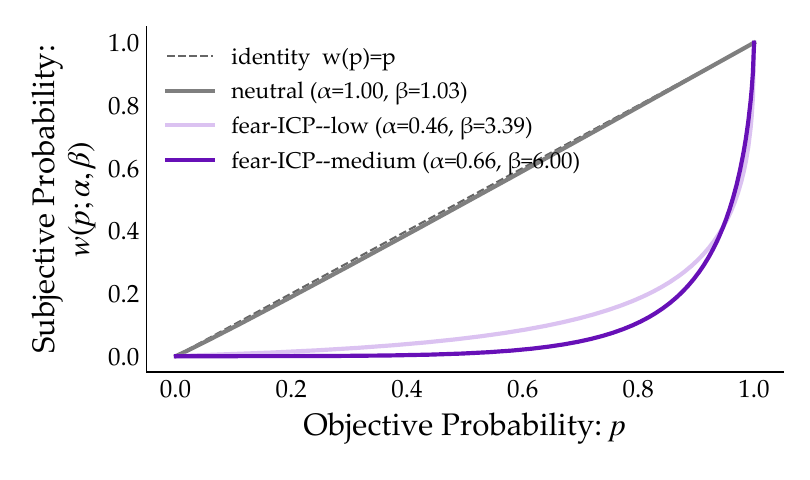}
    \caption{\texttt{Qwen3-4B-ICP-Low-Med}}
\end{subfigure}
\caption{Fitted Prelec probability-weighting functions, comparing neutral and fear steering in \texttt{Olmo3-7B} different levels of steering}
\label{fig:prelec-Qwen3-4B-control}
\end{figure}

Fig.~\ref{fig:prelec-Qwen3-4B-control} reports fitted Prelec probability-weighting functions under varying fear-steering strengths. Consistent with the risk curves, larger RLS magnitudes increase the distortion in $w(p)$ (with higher fitted $\alpha$ and $\beta$), while still retaining a structured mapping from objective to subjective probability. Under ICP, attenuated wording can reduce curvature modestly, but the induced pessimism remains substantial (e.g., $\beta$ remains large and $\alpha<1$), reflecting limited calibration.

\begin{figure}
\centering
\begin{subfigure}[t]{0.46\linewidth}
  \centering
  \includegraphics[width=\linewidth]{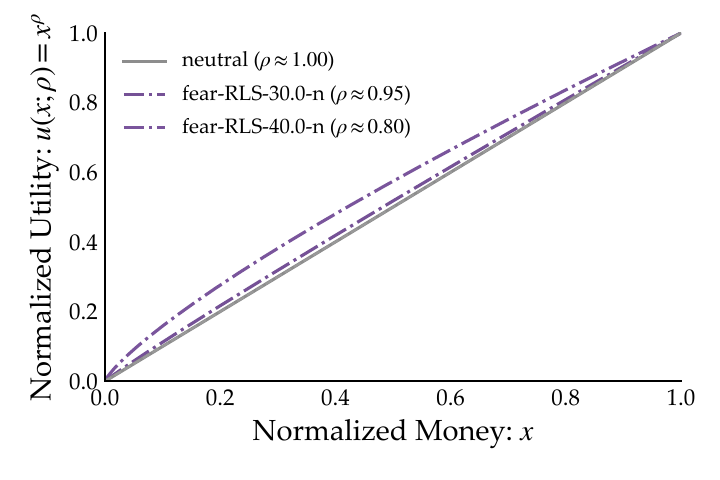}
\caption{\texttt{Qwen3-4B}}
\end{subfigure}\hfill
\begin{subfigure}[t]{0.46\linewidth}
  \centering
  \includegraphics[width=\linewidth]{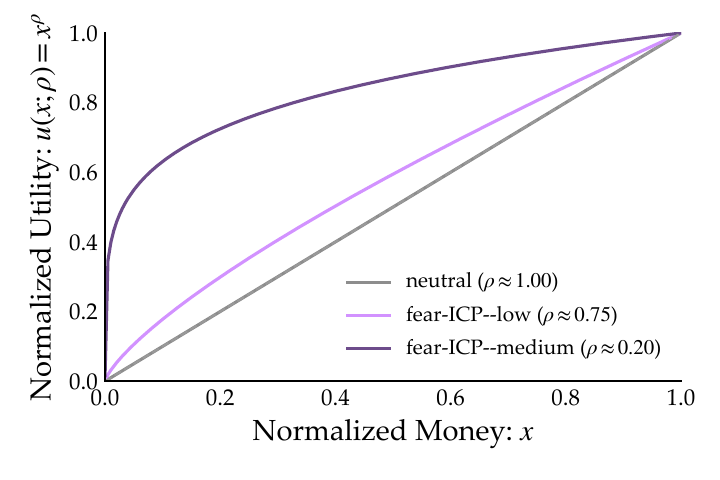}
    \caption{\texttt{Qwen3-4B}}
\end{subfigure}
\caption{Estimated normalized utility curvature across neutral and emotion-steered conditions in \texttt{Qwen3-4B}. - different levels}
\label{fig:utility-Qwen3-4B-control}
\end{figure}

Fig.~\ref{fig:utility-Qwen3-4B-control} compares inferred utility curvature across steering levels. Under RLS, $\rho$ decreases smoothly as fear magnitude increases (e.g., from $\rho\approx 0.95$ toward $\rho\approx 0.8$), indicating a graded increase in diminishing marginal utility over gains. Under ICP, the fitted exponents are more extreme and less stable across wording variants, consistent with the saturation behavior observed in choices.

\begin{figure}[!h]
\centering
\begin{subfigure}[t]{0.46\linewidth}
  \centering
  \includegraphics[width=\linewidth]{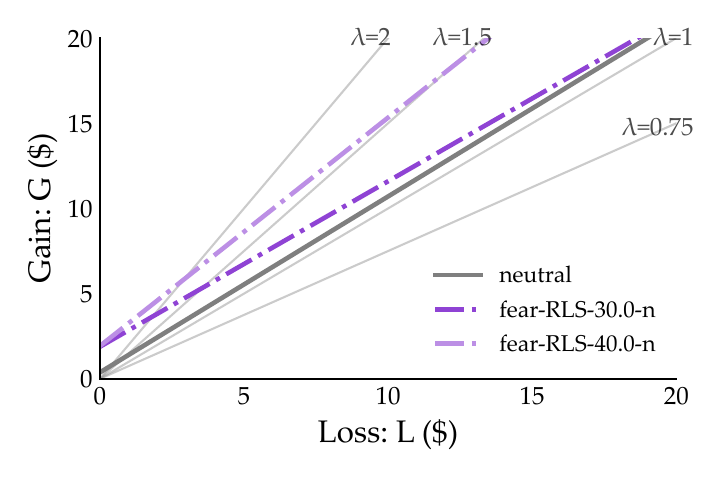}
\caption{\texttt{Qwen3-4B}}
\end{subfigure}\hfill
\begin{subfigure}[t]{0.46\linewidth}
  \centering
  \includegraphics[width=\linewidth]{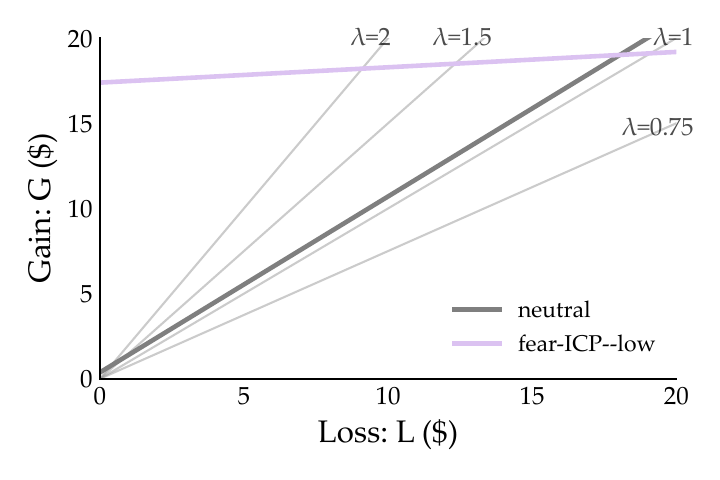}
    \caption{\texttt{Qwen3-4B}}
\end{subfigure}
\caption{Loss-aversion geometry: reference iso-utility lines $G=\lambda L$ with fitted 50\% acceptance boundary under neutral and fear steering (ICP/RLS) in  \texttt{Qwen3-4B}.}
\label{fig:loss-Qwen3-4B_control}
\end{figure}

\begin{figure}[!h]
    \centering
    \includegraphics[width=.46\linewidth]{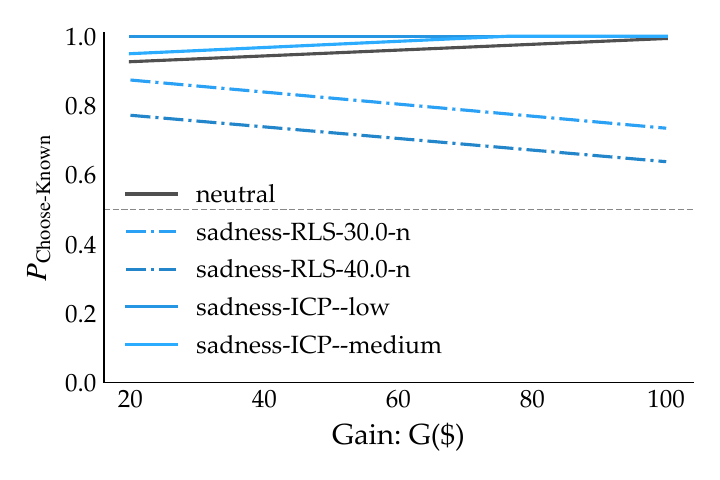}
    \caption{Ambiguity aversion by stake size: linear fits of $P_\text{Choose-Known}$ as a function of gain $G$, comparing neutral and sadness steering (ICP/RLS) in \texttt{Qwen3} comparing low mid ICP RLS.}
    \label{fig:ambig-stake_control}
\end{figure}

Finally, Fig.~\ref{fig:loss-Qwen3-4B_control} shows loss-aversion geometry under varying steering strengths. ICP at medium strength can induce near-universal rejection, rendering $\lambda$ unidentifiable; even when curve fitting is possible, ICP primarily manifests as a large positive intercept (a gain ``entry fee'' to accept \emph{any} loss) rather than a clean increase in the loss-aversion slope. In contrast, RLS exhibits a more controllable pattern, with $\lambda$ increasing from approximately $1$ to $\sim 1.5$ as injection magnitude increases. A similarly graded response appears in ambiguity aversion (Fig.~\ref{fig:ambig-stake_control}): increasing RLS magnitude shifts choices toward ambiguity neutrality while preserving stake dependence, whereas ICP remains near-ceiling ambiguity-averse across wording levels.

\subsection{Steering Thinking-only}
\label{app:think-only}

In our main RLS experiments, the intervention is applied to all newly generated tokens, including both the model’s deliberation (``thinking'') and its final response. In figures throughout App.~\ref{app:control} (e.g., Figs.~\ref{fig:risk-Qwen3-4B_control}, \ref{fig:prelec-Qwen3-4B-control}, \ref{fig:utility-Qwen3-4B-control}, \ref{fig:loss-Qwen3-4B_control}, and \ref{fig:ambig-stake_control}), we denote this setting by \texttt{-n} (``new'').

Here we ask a complementary question: what happens if we steer \emph{only} the latent states during the deliberation trace and leave the final answer generation unsteered? This ``thinking-only'' setting provides a simple way to localize where steering exerts its influence and may also act as a form of attenuation by restricting the intervention to the reasoning phase. We present an initial comparison at a fixed, relatively large injection scale (40), using \texttt{th} to denote thinking-only steering and \texttt{n} to denote steering applied to all generated tokens.

\begin{figure}[!h]
\centering
\begin{subfigure}[t]{0.46\linewidth}
    \centering
    \includegraphics[width=\linewidth]{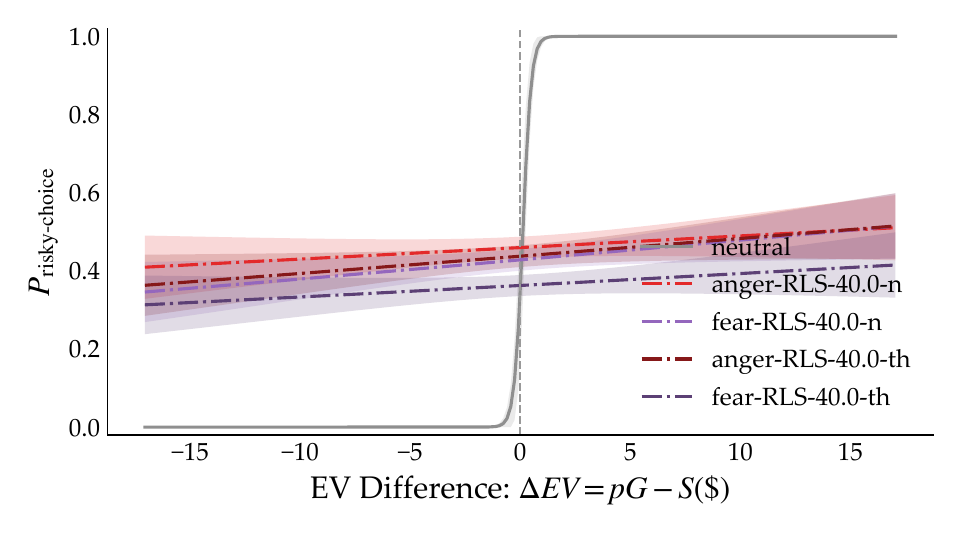}
    \caption{Risk: fitted $P_{risky choice}$ vs.\ $\Delta EV$ under \texttt{th} vs.\ \texttt{n}  steering.}

\end{subfigure}\hfill
\begin{subfigure}[t]{0.46\linewidth}
    \centering
    \includegraphics[width=\linewidth]{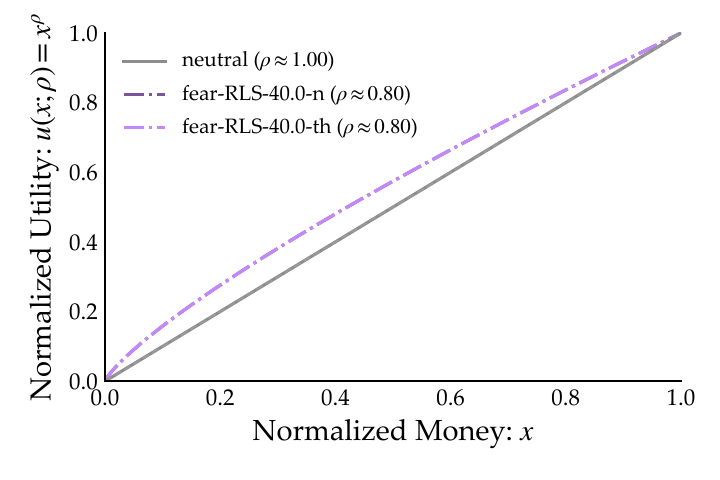}
    \caption{Utility: estimated normalized $u(x)$ under \texttt{th} vs.\ \texttt{n} steering.}

\end{subfigure}
\caption{Thinking-only (\texttt{th}) versus all-token (\texttt{n}) RLS in \texttt{Qwen3-4B}: risk preference curves and utility curvature.}

\label{fig:think_v_new_Qwen3-4b}
\end{figure}

\begin{figure}[!h]
\centering
\begin{subfigure}[t]{0.44\linewidth}
  \centering
  \includegraphics[width=\linewidth]{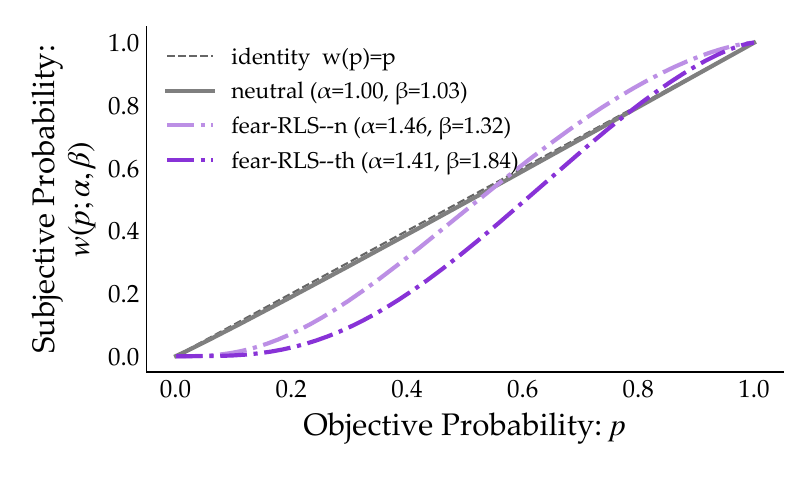}
  \caption{Weighting: fitted Prelec $w(p)$ under \texttt{th} vs.\ \texttt{n} steering.}

\end{subfigure}\hfill
\begin{subfigure}[t]{0.44\linewidth}
  \centering
  \includegraphics[width=\linewidth]{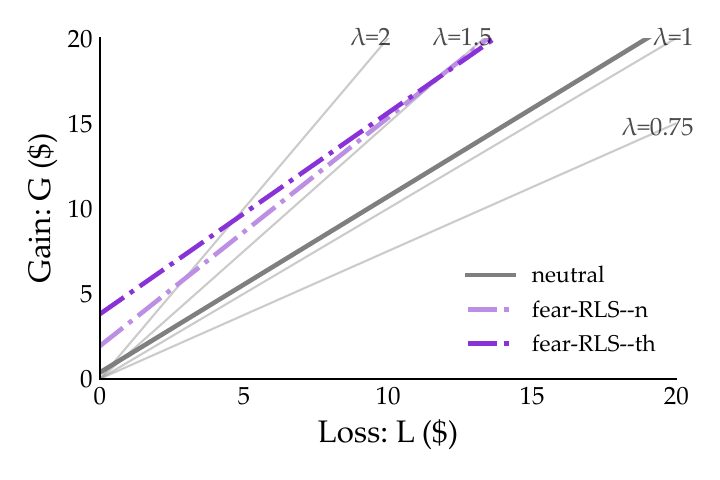}
  \caption{Loss: fitted $50\%$ acceptance boundary under \texttt{th} vs.\ \texttt{n} steering.}

\end{subfigure}
\caption{Thinking-only (\texttt{th}) versus all-token (\texttt{n}) RLS in \texttt{Qwen3-4B}: probability weighting and loss-aversion geometry.}

\label{fig:think-v-new-Qwen3}
\end{figure}

Figs.~\ref{fig:think_v_new_Qwen3-4b} and \ref{fig:think-v-new-Qwen3} show that restricting RLS to the thinking trace can change quantitative details, but does not reliably reduce the overall behavioral shift. In several tasks, the thinking-only and all-token interventions produce broadly similar patterns, suggesting that a substantial portion of RLS’s influence on choice is mediated through the deliberation trace rather than the surface response alone. 

\clearpage
\section{Code and Compute Resources}
\label{sec:compute}

All experiments were executed on GPU-accelerated infrastructure, primarily using NVIDIA A40 and L40-class hardware. Reproducing the largest model configurations requires GPUs with at least 40\,GB of VRAM. Under this setup, reproducing the results reported in this paper takes approximately 48 hours of wall-clock time.

We will release the full experimental code, evaluation prompts, and processed datasets publicly upon acceptance to facilitate reproducibility and follow-up work. 

We used generative AI tools for writing refinement and occasional code completion. All scientific contributions, including experimental design, benchmark construction, steering specifications, statistical analyses, and interpretation, were developed and executed by the authors.

\section*{Concluding Remarks}
\label{sec:conclusion}

This work provides a controlled testbed for characterizing when LLMs behave as utility-driven decision-makers and when they exhibit systematic departures from normative choice, especially under incidental affect. To our knowledge, this is the first controlled benchmark study to jointly evaluate \emph{rational-choice consistency} and \emph{incidental affective distortions} in LLM decision-making, despite this coupling being central to human judgment and increasingly relevant as LLMs mediate consequential decisions.  Across model families and paradigms, enabling deliberative ``thinking'' reliably increases compliance with rational-choice axioms and shifts behavior toward expected-value maximization. Yet important deviations remain even in this regime, most notably in ownership-based pricing and temporal valuation, underscoring that improved reasoning does not automatically confer human-like preference representations.

To probe human-like affective distortions, we compared two qualitatively different steering mechanisms and show that affective control is not monolithic in its effects on model judgment and choice. In-context priming produces large, directionally human-aligned shifts that are often exaggerated and hard to calibrate, consistent with models explicitly reasoning about how emotions \emph{should} influence choice. Representation-level steering yields more graded patterns that appear more psychologically plausible in some settings (e.g., increased risk aversion that remains sensitive to expected payoff), but effects are not consistently replicated across tasks or scales and are sensitive to the vector-injection configuration. Together, these results reveal a central tension: the same deliberative machinery that improves rationality can amplify sensitivity to affective interventions, which matters both for using LLMs as models of human judgment and for the safe deployment of LLM-based decision systems.

More broadly, our benchmark suite operationalizes theoretically interpretable mechanisms (e.g., value sensitivity, probability weighting, and utility curvature) and couples them with controlled affective manipulations, enabling systematic comparisons across models, steering methods, and decision domains. By turning a scattered set of behavioral claims into a unified, parameterized evaluation framework, this work establishes a reproducible foundation for the scientific study of emotion, reasoning, and choice in LLMs and provides practical diagnostics for auditing and shaping LLM decision-making behavior in real deployments. This is consequential well beyond academic interest: if LLMs increasingly mediate decisions in medicine, hiring, finance, and public policy, then understanding when they behave rationally, when they mimic human bias, and when they can be steered is essential for protecting fairness, safety, and accountability at societal scale.

\end{document}